\documentclass[runningheads]{llncs}
\usepackage{graphicx}
\usepackage{tikz}
\usepackage{comment}
\usepackage{amsmath,amssymb} 
\usepackage{stfloats}
\usepackage{graphicx}
\usepackage{booktabs}
\usepackage{array}
\usepackage{multirow}
\usepackage{color}
\usepackage{colortbl}
\usepackage{framed}
\usepackage{bm}
\usepackage{bbm}
\usepackage{xspace}
\usepackage{enumitem}
\usepackage{pifont}
\usepackage{xcolor}
\usepackage{scalerel,graphicx,xparse}
\usepackage{mwe}
\usepackage[pagebackref,breaklinks,hidelinks,hyperfootnotes=false]{hyperref}

\definecolor{Gray}{gray}{0.9}

\usepackage[accsupp]{axessibility}  

\newcommand{\tit}[1]{\smallbreak\noindent\textbf{#1.}}

\newcommand{\tinytit}[1]{\noindent\textbf{#1.}}

\newcommand{\cmark}{\ding{51}}%
\newcommand{\xmark}{\ding{55}}%

\newcommand{\ul}[1]{\underline{#1}}

\def \ie {\emph{i.e.}}
\def \eg {\emph{e.g.}}
\def \etal {\emph{et al.}}

\newcommand\blfootnote[1]{%
  \begingroup
  \renewcommand\thefootnote{}\footnote{#1}%
  \addtocounter{footnote}{-1}%
  \endgroup
}

\begin{document}
\pagestyle{headings}
\mainmatter
\def\ECCVSubNumber{5660}  

\title{Dress Code:\\High-Resolution Multi-Category Virtual Try-On} 
\titlerunning{Dress Code: High-Resolution Multi-Category Virtual Try-On}

\author{
Davide Morelli\inst{1} \and
Matteo Fincato\inst{1} \and
Marcella Cornia\inst{1} \and
Federico Landi\inst{1}$^{,*}$ \and
\\
Fabio Cesari\inst{2} \and
Rita Cucchiara\inst{1}
}
\authorrunning{D. Morelli~\etal}

\institute{
University of Modena and Reggio Emilia, Italy\\
\email{\{name.surname\}@unimore.it} \and
YOOX NET-A-PORTER, Italy\\
\email{\{name.surname\}@ynap.com}
}

\maketitle

\begin{abstract}
Image-based virtual try-on strives to transfer the appearance of a clothing item onto the image of a target person. Prior work focuses mainly on upper-body clothes (\eg~t-shirts, shirts, and tops) and neglects full-body or lower-body items. This shortcoming arises from a main factor: current publicly available datasets for image-based virtual try-on do not account for this variety, thus limiting progress in the field. To address this deficiency, we introduce Dress Code, which contains images of multi-category clothes. Dress Code is more than $3\times$ larger than publicly available datasets for image-based virtual try-on and features high-resolution paired images ($1024 \times 768$) with front-view, full-body reference models. To generate HD try-on images with high visual quality and rich in details, we propose to learn fine-grained discriminating features. Specifically, we leverage a semantic-aware discriminator that makes predictions at pixel-level instead of image- or patch-level. Extensive experimental evaluation demonstrates that the proposed approach surpasses the baselines and state-of-the-art competitors in terms of visual quality and quantitative results. The Dress Code dataset is publicly available at \url{https://github.com/aimagelab/dress-code}.
\keywords{Dress Code Dataset, Virtual Try-On, Image Synthesis.}
\blfootnote{$^*$Now at Huawei Technologies, Amsterdam Research Center, the Netherlands.}
\end{abstract}

\section{Introduction} \vspace{-0.1cm}
Clothes, fashion, and style play a fundamental role in our daily life and allow people to communicate and express themselves freely and directly. With the advent of e-commerce, the variety and availability of online garments have become increasingly overwhelming for the customer. Consequently, user-oriented applications such as virtual try-on, in both 2D~\cite{choi2021viton,han2019clothflow,han2018viton,yang2020towards} and 3D~\cite{mir2020learning,santesteban2021self,zhao2021m3d,zhu2020deep} settings, are increasingly important for online shopping, helping fashion companies to tailor the e-commerce experience and maximize customer satisfaction. Image-based virtual try-on aims at synthesizing an image of a reference person wearing a given try-on garment. In this task, while virtually changing clothing, the person's intrinsic information such as body shape and pose should not be modified. Also, the try-on garment is expected to properly fit the person's body while maintaining its original texture. All these elements make virtual try-on a very active and challenging research topic.

\begin{figure}[t]
\centering
\includegraphics[width=0.98\linewidth]{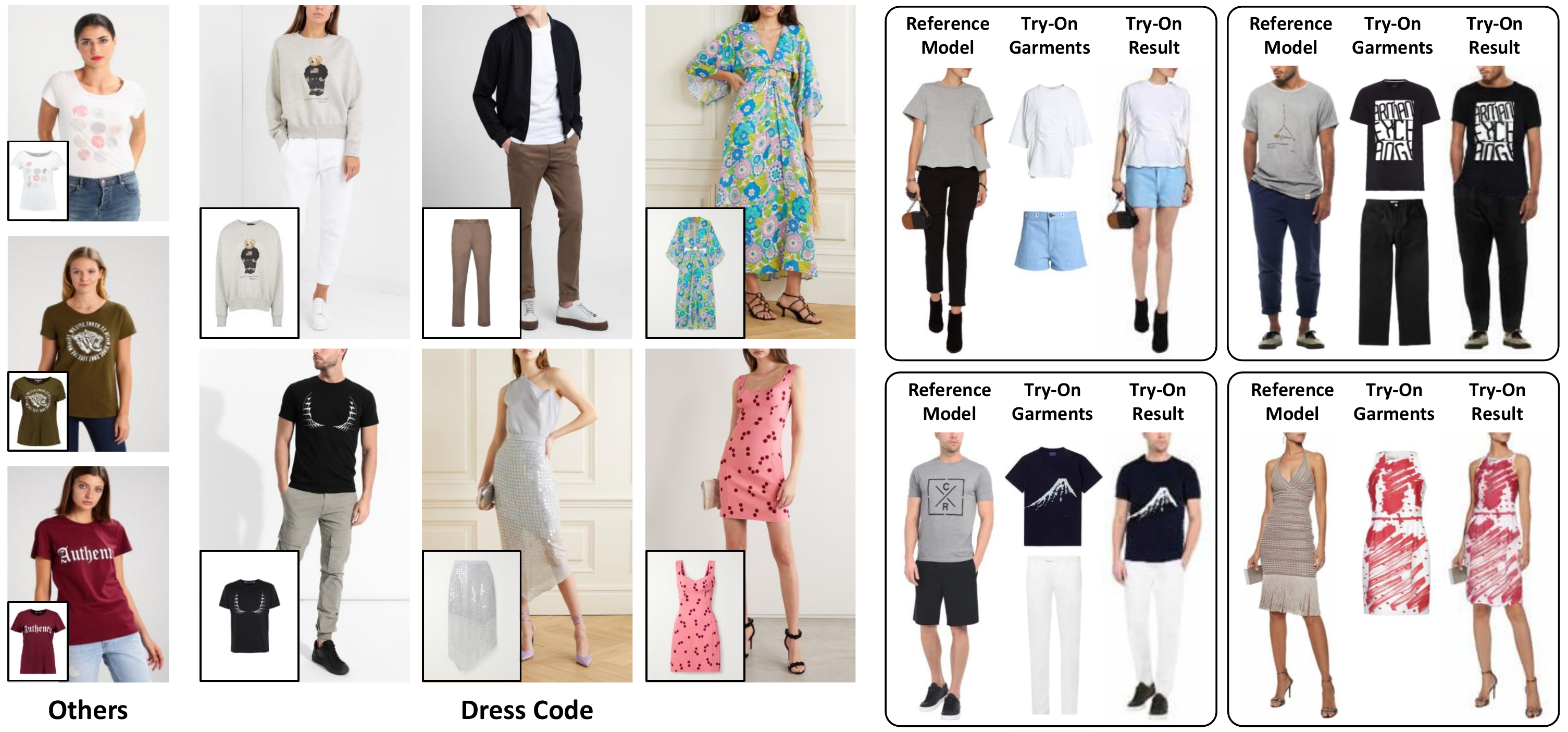}
\caption{Differently from existing publicly available datasets for virtual try-on, Dress Code features different garments, also belonging to lower-body and full-body categories, and high-resolution images.}
\label{fig:first_page}
\vspace{-0.35cm}
\end{figure}

Due to the strategic role that virtual try-on plays in e-commerce, many rich and potentially valuable datasets are proprietary and not publicly available to the research community~\cite{lewis2021tryongan,li2021toward,neuberger2020image,yildirim2019generating}. Public datasets, instead, either do not contain paired images of models and garments or feature a very limited number of images~\cite{han2018viton}. Moreover, the overall image resolution is low (mostly $256 \times 192$). Unfortunately, these drawbacks slow down progress in the field. In this paper, we present \textit{Dress Code}: a new dataset of high-resolution images ($1024 \times 768$) containing more than 50k image pairs of try-on garments and corresponding catalog images where each item is worn by a model. This makes Dress Code more than $3\times$ larger than VITON~\cite{han2018viton}, the most common benchmark for virtual try-on. Differently from existing publicly available datasets, which contain only upper-body clothes, Dress Code features upper-body, lower-body, and full-body clothes, as well as full-body images of human models (Fig.~\ref{fig:first_page}, \textit{left}). 

Current off-the-shelf architectures for virtual try-on are not optimized to work with clothes belonging to different macro-categories (\ie~upper-body, lower-body, and full-body clothes). In fact, this would require learning the correspondences between a particular garment class and the portion of the body involved in the try-on phase. For instance, trousers should match the legs pose, while a dress should match the pose of the entire body, from shoulders to hips and eventually knees. In this paper, we design an image-based virtual try-on architecture that can anchor the given garment to the right portion of the body. As a consequence, it is possible to perform a ``complete'' try-on over a given person by selecting different garments (Fig~\ref{fig:first_page}, \textit{right}).
In order to produce high-quality results rich in visual details, we introduce a parser-based discriminator~\cite{liu2019learning,park2019semantic,schonfeld2021you}. This component can increase the realism and visual quality of the results by learning an internal representation of the semantics of generated images, which is usually neglected by standard discriminator architectures~\cite{isola2017image,Wang_2018_CVPR}. This component works at pixel-level and predicts not only real/generated labels but also the semantic classes for each image pixel. We validate the effectiveness of the proposed approach by testing its performance on both our newly collected dataset and on the most widely used dataset for the task (\ie~VITON~\cite{han2018viton}). 

The contributions of this paper are summarized as follows: (1) We introduce Dress Code, a novel dataset for the virtual try-on task. To the best of our knowledge, it is the first publicly available dataset featuring lower-body and full-body clothes. As a plus, all images have high resolution ($1024\times 768$). (2) To address the challenges of generating high-quality images, we leverage a Pixel-level Semantic-Aware Discriminator (PSAD) that enhances the realism of try-on images. (3) With the aim of presenting a comprehensive benchmark on our newly collected dataset, we train and test up to nine state-of-the-art virtual try-on approaches and three different baselines. (4) Extensive experiments demonstrate that the proposed approach outperforms the competitors and other state-of-the-art architectures both quantitatively and qualitatively, also considering different image resolutions and a multi-garment setting.

\section{Related Work} \vspace{-0.1cm}
The first popular image-based virtual try-on model~\cite{han2018viton} builds upon a coarse-to-fine network. First, it predicts a coarse image of the reference person wearing the try-on garment, then it refines the texture and shape of the previously obtained result. Wang~\etal~\cite{wang2018toward} overcame the lack of shape-context precision (\ie~bad alignment between clothes and body shape) and proposed a geometric transformation module to learn the parameters of a thin-plate spline transformation to warp the input garment. Following this work, many different solutions were proposed to enhance the geometric transformation of the try-on garment. For instance, Liu~\etal~\cite{liu2016deepfashion} integrated a multi-scale patch adversarial loss to increase the realism in the warping phase. Minar~\etal~\cite{minar2020cpvton} and Yang~\etal~\cite{yang2020towards} proposed different regularization techniques to stabilize the warping process during training. Instead, other works~\cite{fincato2020viton,li2021toward} focused on the design of additional projections of the input garment to preserve details and textures of input clothing items.

Another line of work focuses on the improvement of the generation phase of final try-on images~\cite{choi2021viton,fenocchi2022dual,ge2021disentangled,he2022style,issenhuth2019end,jae2019viton}. Among them, Issenuth~\etal~\cite{issenhuth2019end} introduced a teacher-student approach: the teacher learns to generate the try-on results using image pairs (sampled from a paired dataset) and then teaches the student how to deal with unpaired data. This paradigm was further improved in~\cite{ge2021parser} with a student-tutor-teacher architecture where the network is trained in a parser-free way, exploiting both the tutor guidance and the teacher supervision. On a different line, Ge~\etal~\cite{ge2021disentangled} presented a self-supervised trainable network to reframe the virtual try-on task as clothes warping, skin synthesis, and image composition using a cycle-consistent framework.

A third direction of research estimates the person semantic layout to improve the visual quality of generated images~\cite{han2019clothflow,jandial2020sievenet,yang2020towards,yu2019vtnfp}. In this context, Jandial~\etal~\cite{jandial2020sievenet} proposed to generate a conditional segmentation mask to handle occlusions and complex body poses effectively. Very recently, Chen~\etal~\cite{Chen_2021_ICCV} introduced a new scenario where the try-on results are synthesized in sequential poses with spatio-temporal smoothness. Using a recurrent approach, Cui~\etal~\cite{Cui_2021_ICCV} designed a person generation framework for pose transfer, virtual try-on, and other fashion-related tasks.
While almost all these methods generate low-resolution results, a limited subset of works focuses on the generation of higher-resolution images instead. Unfortunately, these works employ non-public datasets to train and test the proposed architectures~\cite{li2021toward,yildirim2019generating}. 

\section{Dress Code Dataset} \vspace{-0.1cm}
\label{sec:dataset}

\begin{figure}[t]
\centering
\scriptsize
\setlength{\tabcolsep}{.25em}
\resizebox{\linewidth}{!}{
\begin{tabular}{cc cccc}
\rotatebox{90}{\parbox[t]{0.86in}{\hspace*{\fill}\textbf{Upper-body}\hspace*{\fill}}} & &
\includegraphics[height=0.18\linewidth]{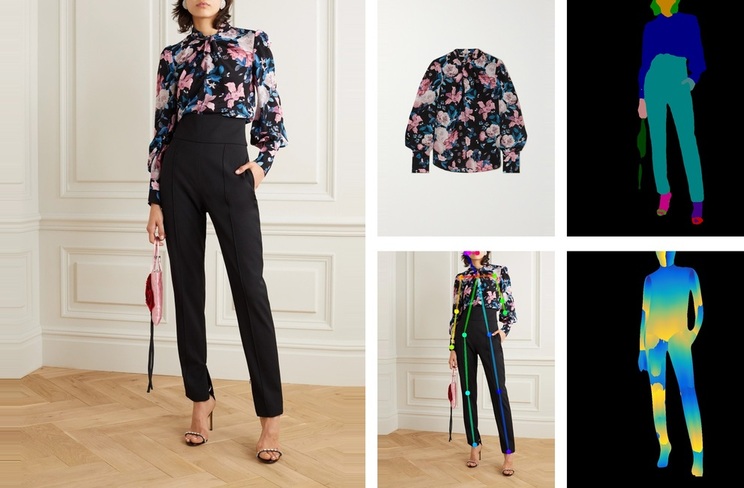} &
\includegraphics[height=0.18\linewidth]{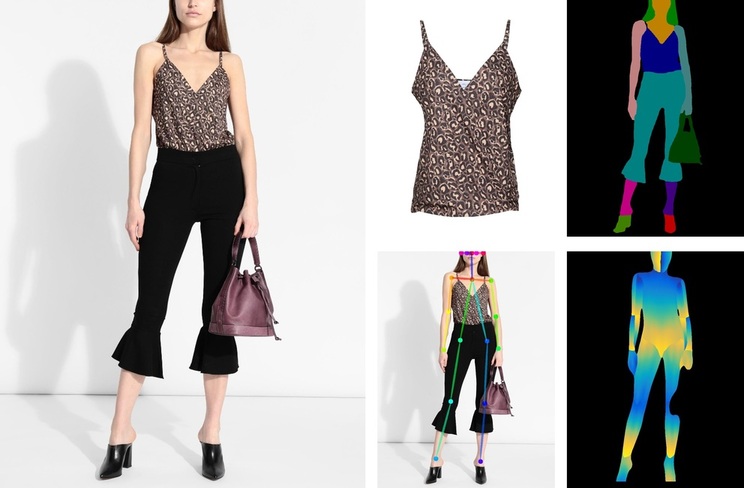} &
\includegraphics[height=0.18\linewidth]{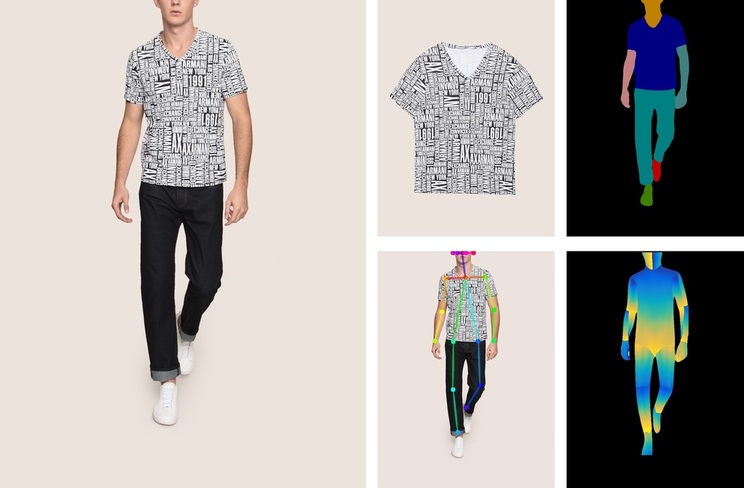} &
\includegraphics[height=0.18\linewidth]{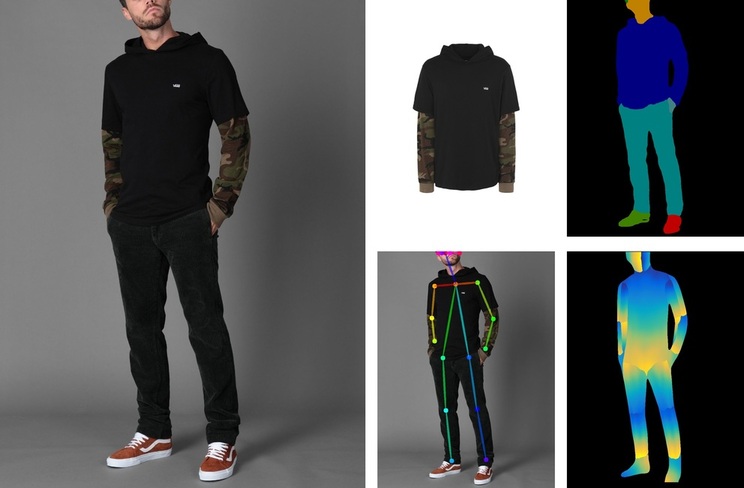} \\
\addlinespace[0.04cm]
\rotatebox{90}{\parbox[t]{0.86in}{\hspace*{\fill}\textbf{Lower-body}\hspace*{\fill}}} & &
\includegraphics[height=0.18\linewidth]{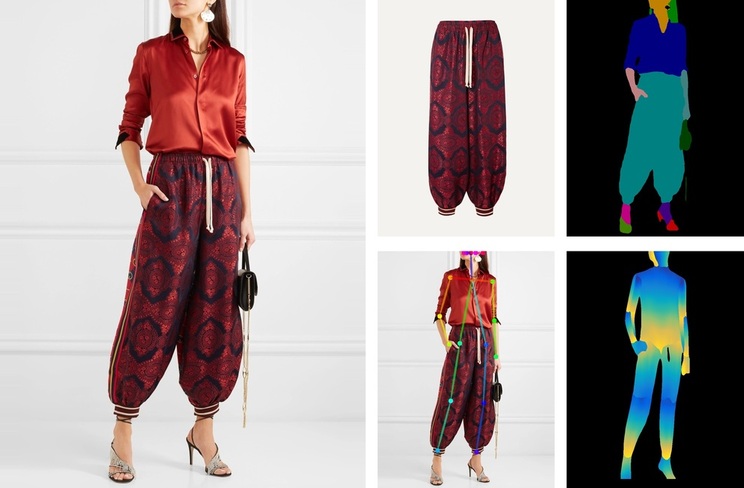} &
\includegraphics[height=0.18\linewidth]{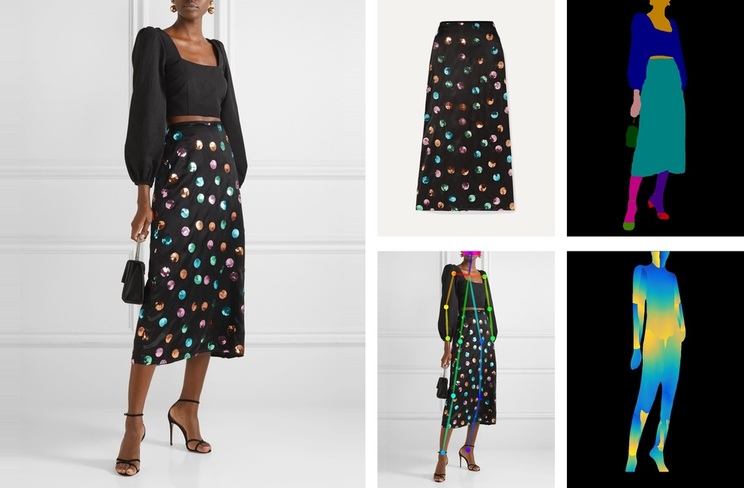} &
\includegraphics[height=0.18\linewidth]{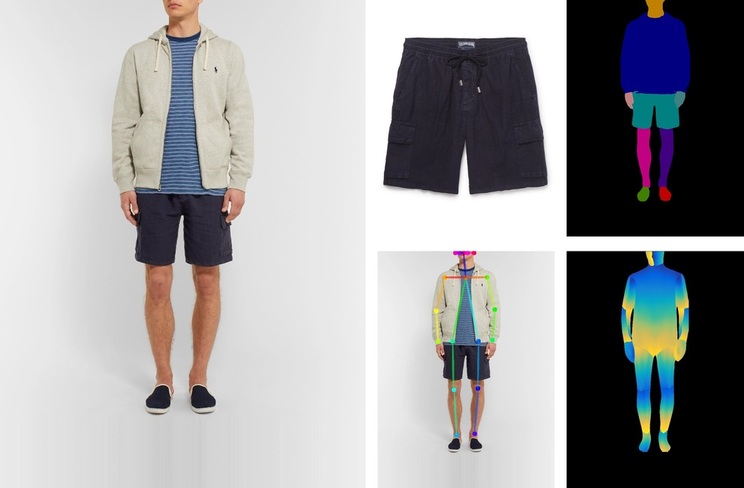} &
\includegraphics[height=0.18\linewidth]{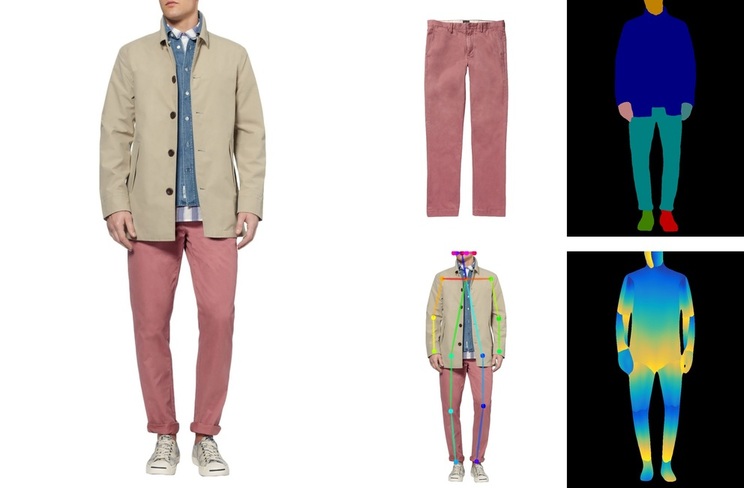} \\
\addlinespace[0.04cm]
\rotatebox{90}{\parbox[t]{0.86in}{\hspace*{\fill}\textbf{Dresses}\hspace*{\fill}}} & &
\includegraphics[height=0.18\linewidth]{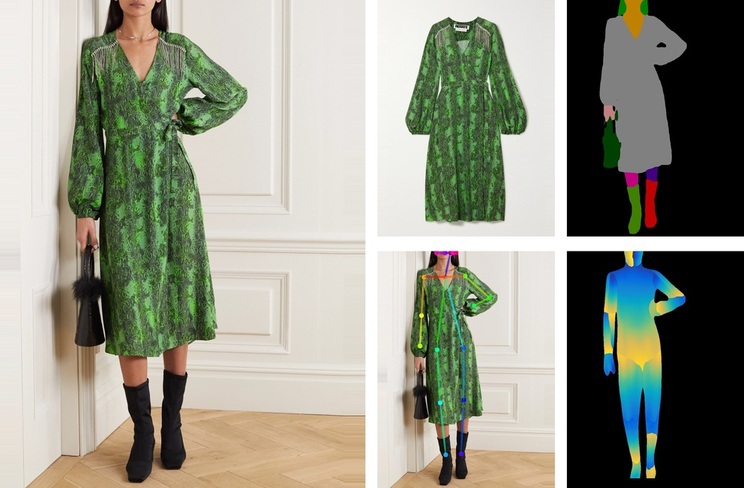} &
\includegraphics[height=0.18\linewidth]{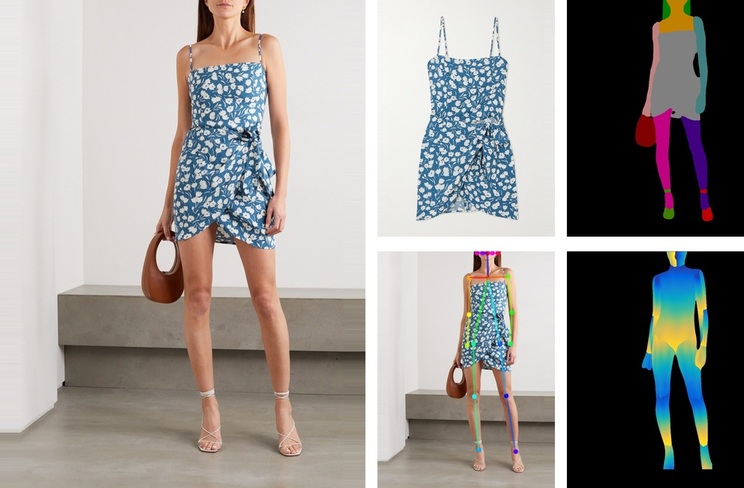} &
\includegraphics[height=0.18\linewidth]{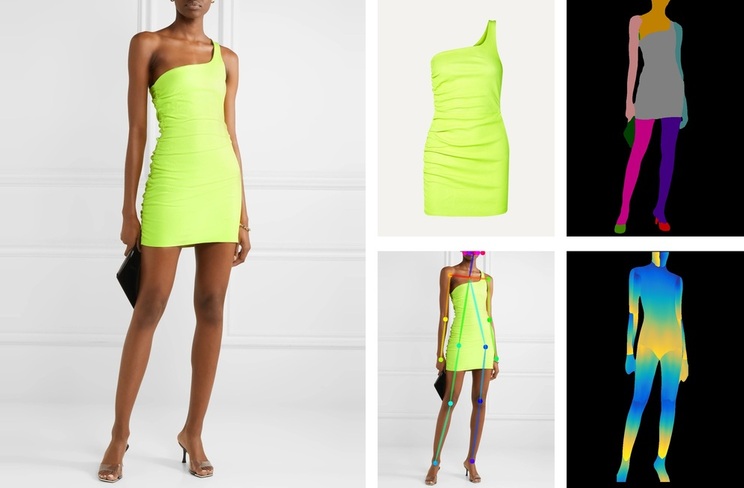} &
\includegraphics[height=0.18\linewidth]{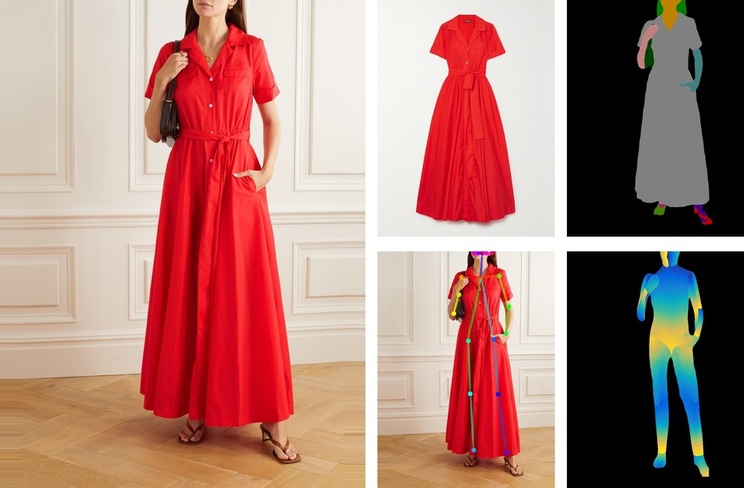} \\
\end{tabular}
}
\caption{Sample image pairs from the Dress Code dataset with pose keypoints, dense poses, and segmentation masks of human bodies.}
\label{fig:pose_parsing}
\vspace{-0.35cm}
\end{figure}

Publicly available datasets for virtual try-on are often limited by one or more factors such as lack of variety, small size, low-resolution images, privacy concerns, or from the fact of being proprietary. We identify four main desiderata that the ideal dataset for virtual try-on should possess: (1) it should be publicly available for research purposes; (2) it should have corresponding images of clothes and reference human models wearing them (\ie~the dataset should consist of paired images); (3) it should contain high-resolution images and (4) clothes belonging to different macro-categories (tops and t-shirts belong to the upper-body category, while skirts and trousers are examples of lower-body clothes and dresses are full-body garments). In addition to this, a dataset for virtual try-on with a large number of images is more preferable than other datasets with the same overall characteristics but smaller size.
By looking at Table~\ref{tab:comparison}, we can see that Dress Code complies with all of the above desiderata, while featuring more than three times the number of images of VITON~\cite{han2018viton}. To the best of our knowledge, this is the first publicly available virtual try-on dataset comprising multiple macro-categories and high-resolution image pairs. Additionally, it is the biggest available dataset for this task at present, as it includes more than 100k images evenly split between garments and human reference models.

\tit{Image collection and annotation}
All images are collected from different fashion catalogs of YOOX NET-A-PORTER containing both casual clothes and luxury garments. To create a coarse version of the dataset, we select images of different garment categories for a total of about 250k fashion items, each containing 2-5 images of different views of the same product. Since our goal is to create a dataset for virtual try-on and not all fashion items were released with the image pair required to perform the task, we select only those products where the front-view image of the garment and the corresponding full figure of the model are available. We exploit an automatic selection procedure: we only store the clothing items for which at least one image with the entire body of the model is present, using a human pose estimator to verify the presence of the neck and feet joints. In this way, all products without valid image pairs are automatically discarded. After this automatic stage, we manually validate all images and remove the remaining invalid image pairs, including those pairs for which the garment of interest is mostly occluded by other overlapping clothes. Finally, we group the annotated products into three categories: upper-body clothes (composed of tops, t-shirts, shirts, sweatshirts, and sweaters), lower-body clothes (composed of skirts, trousers, shorts, and leggings), and dresses. Overall, the dataset is composed of 53,795 image pairs: 15,366 pairs for upper-body clothes, 8,951 pairs for lower-body clothes, and 29,478 pairs for dresses. 

\begin{table}[t]
\centering
\footnotesize
\caption{Comparison between Dress Code and the most widely used datasets for virtual try-on and other related tasks.}
\label{tab:comparison}
\setlength{\tabcolsep}{.35em}
\resizebox{0.95\linewidth}{!}{
\begin{tabular}{l c ccccc}
\toprule
\textbf{Dataset} & & \textbf{Public} & \textbf{Multi-Category} & \textbf{\# Images} & \textbf{\# Garments}  & \textbf{Resolution} \\
\midrule
O-VITON~\cite{neuberger2020image}       & & \xmark & \cmark & 52,000 &    -    & $512 \times 256$ \\
TryOnGAN~\cite{lewis2021tryongan}       & & \xmark & \cmark & 105,000 &    -   & $512 \times 512$\\
Revery AI~\cite{li2021toward}           & & \xmark & \cmark & 642,000 & 321,000 & $512 \times 512$ \\
Zalando~\cite{yildirim2019generating}   & & \xmark & \cmark & 1,520,000 & 1,140,000 & $1024 \times 768$ \\
\midrule
VITON-HD~\cite{choi2021viton}           & & \cmark & \xmark & 27,358 &  13,679 & $1024 \times 768$ \\
FashionOn~\cite{hsieh2019fashionon}     & & \cmark & \xmark & 32,685 & 10,895 & $288 \times 192$ \\
DeepFashion~\cite{liu2016deepfashion}   & & \cmark & \xmark & 33,849 & 11,283 & $288 \times 192$ \\
MVP~\cite{dong2019towards}              & & \cmark & \xmark & 49,211 & 13,524 & $256 \times 192$ \\
FashionTryOn~\cite{zheng2019virtually}  & & \cmark & \xmark & 86,142 & 28,714 & $256 \times 192$ \\
LookBook~\cite{yoo2016pixel}            & & \cmark & \cmark & 84,748 &  9,732 & $256 \times 192$ \\
\midrule
VITON~\cite{han2018viton}               & & \cmark & \xmark & 32,506 & 16,253 &  $256 \times 192$ \\
\textbf{Dress Code}                     & & \cmark & \cmark & 107,584 & 53,792 & $1024 \times 768$ \\
\bottomrule
\end{tabular}
}
\vspace{-0.3cm}
\end{table}

Existing datasets for virtual try-on show the face and physiognomy of the human models. While this feature is not essential for virtual try-on, it also causes potential privacy issues. To preserve the models’ identity, we partially anonymize all images by cutting them at the level of the nose. In this way, information about the physiognomy of the human models is not available. To further enrich our dataset, we compute the joint coordinates, the dense pose, and the segmentation mask for the human parsing of each model. In particular, we use OpenPose~\cite{cao2017realtime} to extract 18 keypoints for each human body, DensePose~\cite{guler2018densepose} to compute the dense pose of each reference model, and SCHP~\cite{li2019self} to generate a segmentation mask representing the human parsing of model body parts and clothing items. Sample human model and garment pairs from our dataset with the corresponding additional information are shown in Figure~\ref{fig:pose_parsing}.

\tit{Comparison with other datasets}
Table~\ref{tab:comparison} reports the main characteristics of the Dress Code dataset in comparison with existing datasets for virtual try-on and fashion-related tasks. Although some proprietary and non-publicly available datasets have also been used~\cite{lewis2021tryongan,li2021toward,yildirim2019generating}, almost all virtual try-on literature~\cite{ge2021parser,wang2018toward,yang2020towards} employs the VITON dataset~\cite{han2018viton} to train the proposed models and perform experiments.
We believe that the use of Dress Code could greatly increase the performance and applicability of virtual try-on solutions. In fact, when comparing Dress Code with the VITON dataset, it can be seen that our dataset jointly features a larger number of image pairs (\ie~53,792 vs 16,253 of the VITON dataset), a wider variety of clothing items (\ie~VITON only contains t-shirts and upper-body clothes), a greater variance in model images (\ie~Dress Code images can contain challenging backgrounds, accessories like bags, scarfs, and belts, and both male and female models), and a greater image resolution (\ie~$1024\times768$ vs $256 \times 192$ of VITON images).

\section{Virtual Try-On with Pixel-level Semantics}
\label{sec:method}
Architectures for virtual try-on address the task of generating a new image of the reference person wearing the input try-on garment. Given the generative nature of this task, virtual try-on methods are usually trained using adversarial losses that typically work at image- or patch-level and do not consider the semantics of generated images. Differently from previous works, we introduce a Pixel-level Semantic Aware Discriminator (PSAD) that can build an internal representation of each semantic class and increase the realism of generated images. In this section, we first describe the baseline generative architecture and then detail PSAD which improves the visual quality and overall performance.

\subsection{Baseline Architecture}
\label{sec:architecture}
\begin{figure}[t]
\centering
\includegraphics[width=0.95\linewidth]{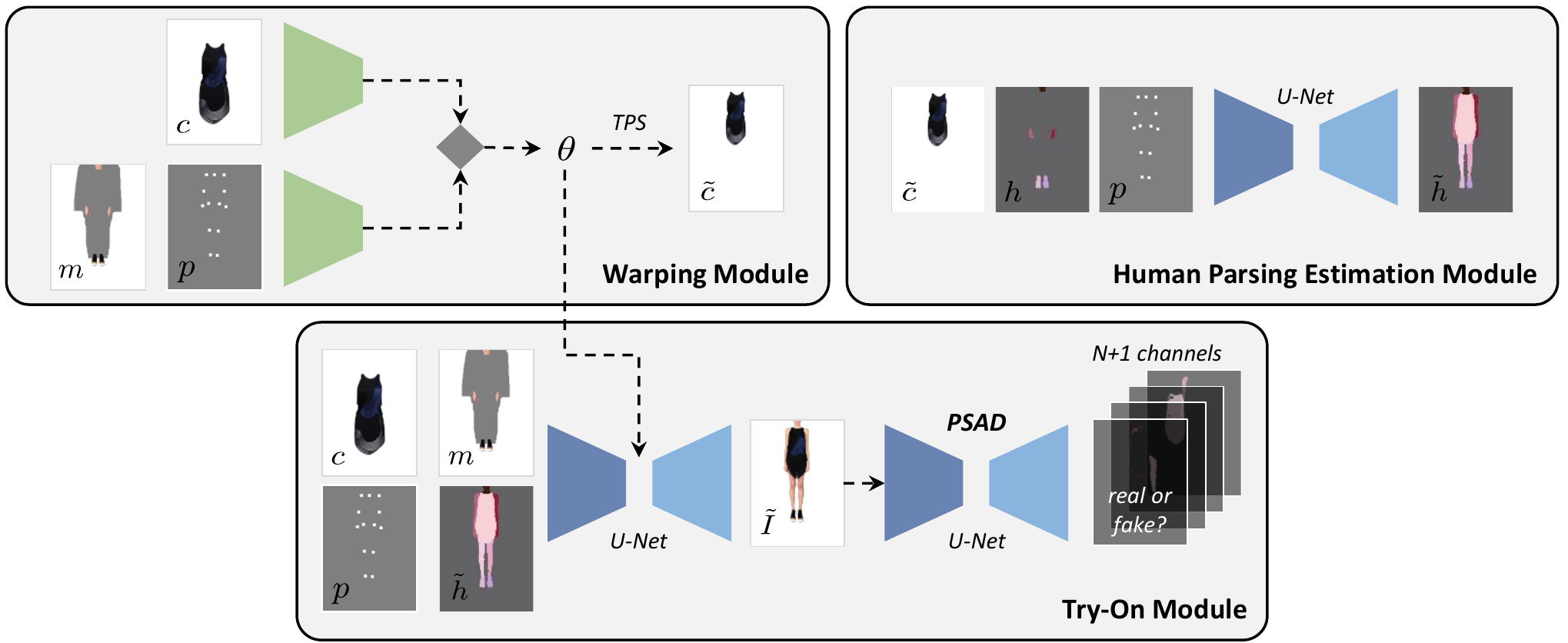}
\caption{Overview of the proposed architecture.}
\label{fig:model}
\vspace{-0.35cm}
\end{figure}

To tackle the virtual try-on task, we begin by building a baseline generative architecture that performs three main operations: (1) garment warping, (2) human parsing estimation, and finally (3) try-on.
First, the warping module employs geometric transformations to create a warped version of the input try-on garment. Then, the human parsing estimation module predicts a semantic map for the reference person.
Last, the try-on module generates the image of the reference person wearing the selected garment. Our baseline model is shown in Fig.~\ref{fig:model} and detailed in the following. 

\tit{Network Inputs and Notation}
Here, we define the different inputs for our network and related notation.
We denote with $c$ an image depicting a clothing item alone. This image contains information about the shape, texture, and color of the try-on garment. Details about the reference human model come in different forms: $p$, $m$, and $h$ are three images containing respectively the pose of that person ($p$), the background and appearance of the portions of the body and outfit that are not involved in the try-on phase such as hands, feet, and part of the face ($m$), and the semantic labels of each of these regions ($h$). 
Our architecture can employ two different representations for the body pose: keypoints or dense pose~\cite{guler2018densepose}. In this section, as well as in Fig.~\ref{fig:model}, we consider the case of pose keypoints. However, it is possible to switch and use dense pose representation by accounting for the different number of channels. Finally, we denote with $I$ the image depicting the person described by $(p, m, h)$ wearing the garment $c$.

\tit{Warping Module}
The warping module transforms the input try-on garment $c$ into a warped image of the same item that matches the body pose and shape expressed respectively by $p$ and $m$.
As warping function we use a thin-plate spline (TPS) geometric transformation~\cite{rocco2017convolutional}, which is commonly used in virtual try-on models~\cite{fincato2020viton,wang2018toward,yang2020towards}.
Inside this module, we aim to learn the correspondence between the inputs $(c, p, m)$ and the set $\theta$ of parameters to be used in the TPS transformation.
Specifically, we follow the warping module proposed in~\cite{wang2018toward} and compute a correlation map between the encoded representations of the try-on garment $c$ and the pose and cloth-agnostic person representation ($p$ and $m$), obtained using two separate convolutional networks. Then, we predict the spatial transformation parameters $\theta$ corresponding to the $(x, y)$-coordinate offsets of TPS anchor points. These parameters are used in the TPS function to generate the warped version $\tilde{c}$ of the input try-on garment:
\begin{equation}
    \tilde{c} = \text{TPS}_\theta(c).
    \label{eq:tildec}
\end{equation}

To train this network, we minimize the $L_1$ distance between the warped result $\tilde{c}$ and the cropped version of the garment $\hat{c}$ obtained from the ground-truth image $I$. In addition, to reduce visible distortions in the warped result, we employ the second-order difference constraint introduced in~\cite{yang2020towards}. Overall, the loss function used to train this module is defined as follows:
\begin{equation}
\mathcal{L}_{\textit{warp}}=\left \| \tilde{c} - \hat{c} \right \|_{1} + \lambda_{\textit{const}} \mathcal{L}_{\textit{const}} ,
\label{eq:loss_warp}
\end{equation}
where $\mathcal{L}_{\textit{const}}$ is the second-order difference constraint and $\lambda_{\textit{const}}$ is used to weigh the constraint loss function~\cite{yang2020towards}.

\tit{Human Parsing Estimation Module}
This module, based on the U-Net architecture~\cite{ronneberger2015u}, takes as input a concatenation of the warped try-on clothing item $\tilde{c}$ (Eq.~\ref{eq:tildec}), the pose image $p$, and the masked semantic image $h$, and predicts the complete semantic map $\tilde{h}$ containing the human parsing for the reference person:
\begin{equation}
    \tilde{h} = \text{U-Net}_\mu(c, h, p) ,
\end{equation}
where $\mu$ denotes the set of learnable weights in the network.
Every pixel of $\tilde{h}$ contains a probability distribution over 18 semantic classes, which include \textit{left/right arm}, \textit{left/right leg}, \textit{background}, \textit{dress}, \textit{shirt}, \textit{skirt}, \textit{neck}, and so on.
We optimize the set of weights $\mu$ of this module using a pixel-wise cross-entropy loss between the generated semantic map $\tilde{h}$ and the ground-truth $\hat{h}$.

\tit{Try-On Module}
This module produces the image $\tilde{I}$ depicting the reference person described by the triple $(p, m, \tilde{h})$ wearing the input try-on clothing item $c$. To this end, we employ a U-Net model~\cite{ronneberger2015u} which takes as input $c$, $p$, $m$, and the one-hot semantic image obtained by taking the pixel-wise argmax of $\tilde{h}$.
During training, instead, we employ the ground-truth human parsing $\hat{h}$. This artifice helps to stabilize training and brings better results.

At this stage, we take advantage of the previously learned geometric transformation $\text{TPS}_\theta$ to facilitate the generation of $\tilde{I}$.
Specifically, we employ a modified version of the U-Net model featuring a two-branch encoder that generates two different representations for the try-on garment $c$ and the reference person, and a decoder that combines these two representations to generate the final image $\tilde{I}$.
The input of the first branch is the original try-on garment $c$, while the input of the second branch is a concatenation of the reference model and corresponding additional information.
In the first branch, we apply the previously learned transformation $\text{TPS}_\theta$. Thus, the skip connections, which are typical of the U-Net design, no longer perform an identity mapping, but compute:
\begin{equation}
    E_i(c) = \text{TPS}_\theta (E_i(c)) ,
\end{equation}
where $E_i(c)$ are the features extracted from the $i^{\textit{th}}$ layer of the U-Net encoder.

During training, we exploit a combination of three different loss functions: an $L_1$ loss between the generated image $\tilde{I}$ and the ground-truth image $I$, a perceptual loss $\mathcal{L}_{\textit{vgg}}$, also know as VGG loss~\cite{johnson2016perceptual}, to compute the difference between the feature maps of $\tilde{I}$ and $I$ extracted with a VGG-19~\cite{simonyan2014very}, and the adversarial loss $\mathcal{L}_{\textit{adv}}$:
\begin{equation}
\mathcal{L}_{\textit{try-on}} = \left \| \tilde{I} - I \right \|_{1} + \mathcal{L}_{\textit{vgg}} + \lambda_{\textit{adv}}\mathcal{L}_{\textit{adv}} ,
\label{eq:loss_try_on}
\end{equation}
where $\lambda_{\textit{adv}}$ is used to weigh the adversarial loss. For a formulation of $\mathcal{L}_{\textit{adv}}$ using our proposed Pixel-level Parsing-Aware Discriminator (PSAD), we refer the reader to the next subsection (Eq.~\ref{eq:adv_loss}).

\subsection{Pixel-level Semantic-Aware Discriminator}
Virtual try-on models are usually enriched with adversarial training strategies to increase the realism of generated images. However, most of the existing discriminator architectures work at image- or patch-level, thus neglecting the semantics of generated images. To address this issue, we draw inspiration from semantic image synthesis literature~\cite{liu2019learning,park2019semantic,schonfeld2021you} and train our discriminator to predict the semantic class of each pixel using generated and ground-truth images as fake and real examples respectively. In this way, the discriminator can learn an internal representation of each semantic class (\eg~tops, skirts, body) and force the generator to improve the quality of synthesized images.

The discriminator is built upon the U-Net model~\cite{ronneberger2015u}, which is used as an encoder-decoder segmentation network. For each pixel of the input image, the discriminator predicts its semantic class and an additional label (real or generated). Overall, we have $N+1$ classes (\ie~$N$ classes corresponding to the ground-truth semantic classes plus one class for fake pixels) and thus we train the discriminator with a $(N+1)$-class pixel-wise cross-entropy loss. In this way, the discriminator prediction shifts from a patch-level classification, typical of standard patch-based discriminators~\cite{isola2017image,Wang_2018_CVPR}, to a per-pixel class-level prediction. 

Due to the unbalanced nature of the semantic classes, we weigh the loss class-wise using the inverse pixel frequency of each class. Formally, the loss function used to train this Pixel-level Parsing-Aware Discriminator (PSAD) can be defined as follows:
\begin{equation}
    \begin{split}
        \mathcal{L}_{adv} = - \mathbb{E}_{(I, \hat{h})} \left[ \sum_{k=1}^{N} w_k  \sum_{i,j}^{H \times W} \hat{h}_{i,j,k} \log{D(I)_{i, j, k}} \right] \\- \mathbb{E}_{(p, m, c, \hat{h})} \left[ \sum_{i, j}^{H \times W} \log{D(G(p, m, c, \hat{h}))}_{i,j,k=N+1} \right] ,
    \end{split}
\label{eq:adv_loss}
\end{equation}
where $I$ is the real image, $\hat{h}$ is the ground-truth human parsing, $p$ is the model pose, $m$ and $c$ are respectively the person representation and the try-on garment given as input to the generator, and $w_k$ is the class inverse pixel frequency.

\section{Experiments}
\label{sec:experiments}

\begin{table}[t]
\centering
\footnotesize
\setlength{\tabcolsep}{.3em}
\caption{Try-on results on the Dress Code test set. Top-1 results are highlighted in bold, underlined denotes second-best.}
\label{tab:try-on_res}
\resizebox{\linewidth}{!}{
\begin{tabular}{lc ccc c ccc c ccc c cccc}
\toprule
& & \multicolumn{3}{c}{\textbf{Upper-body}} & & \multicolumn{3}{c}{\textbf{Lower-body}} & & \multicolumn{3}{c}{\textbf{Dresses}} & & \multicolumn{4}{c}{\textbf{All}} \\
\cmidrule{3-5} \cmidrule{7-9} \cmidrule{11-13} \cmidrule{15-18}
\textbf{Model} & & \textbf{SSIM} $\uparrow$ & \textbf{FID} $\downarrow$ & \textbf{KID} $\downarrow$ & & \textbf{SSIM} $\uparrow$ & \textbf{FID} $\downarrow$ & \textbf{KID} $\downarrow$ & & \textbf{SSIM} $\uparrow$ & \textbf{FID} $\downarrow$ & \textbf{KID} $\downarrow$ & & \textbf{SSIM} $\uparrow$ & \textbf{FID} $\downarrow$ & \textbf{KID} $\downarrow$ & \textbf{IS} $\uparrow$ \\
\midrule
CP-VTON~\cite{wang2018toward} & & 0.812 & 46.99 & 3.236 & & 0.782 & 54.66 & 3.656 & & 0.816 & 34.95 & 1.759 & & 0.803 & 35.16 & 2.245 & 2.817 \\
CP-VTON+~\cite{minar2020cpvton} & & 0.863 & 28.93 & 1.856 & & 0.819 & 41.37 & 2.506 & & 0.826 & 32.27 & 1.630 & & 0.836 & 25.19 & 1.586 & 3.002 \\
CIT~\cite{ren2021cloth} & & 0.860 & 26.41 & 1.496 & & 0.834 & 31.77 & 1.753 & & 0.810 & 35.58 & 1.734 & & 0.835 & 21.99 & 1.313 & 3.022 \\
\midrule
CP-VTON$^\dagger$~\cite{wang2018toward} & & 0.898 & 23.03 & 1.338 & & 0.887 & 26.96 & 1.409 & & 0.838 & 33.04 & 1.668 & & 0.874 & 18.99 & 1.117 & \textbf{3.058} \\
CIT$^\dagger$~\cite{ren2021cloth} & & 0.912 & 17.66 & 0.895 & & 0.896 & 23.15 & 1.005 & & 0.855 & 23.87 & 0.969 & & 0.888 & 13.97 & 0.761 & 3.014 \\
VITON-GT~\cite{fincato2020viton} & & 0.922 & 18.90 & 0.994 & & 0.916 & 21.88 & 0.949 & & 0.864 & 29.45 & 1.402 & & 0.899 & 13.80 & 0.711 & 3.042 \\
WUTON~\cite{issenhuth2019end} & & 0.924 & 17.74 & 0.893 & & 0.918 & 22.57 & 1.008 & & 0.866 & 28.93 & 1.304 & & 0.902 & 13.28 & 0.771 & 3.005 \\
ACGPN~\cite{yang2020towards} & & 0.889 & 19.03 & 1.028 & & 0.874 & 24.46 & 1.208 & & 0.845 & 22.42 & 0.944 & & 0.868 & 13.79 & 0.818 & 2.924 \\
PF-AFN~\cite{ge2021parser} & & 0.918 & 19.03 & 1.237 & & 0.907 & 23.43 & 1.018 & & 0.869 & 21.94 & 0.723 & & 0.902 & 14.36 & 0.756 & 3.023 \\
\midrule
\textit{Dense Pose} \\
\hspace{0.4cm}\textbf{Ours (Patch)} & & \ul{0.930} & 18.21 & 0.929 & & \ul{0.922} & 21.95 & 0.992 & & \ul{0.875} & 21.84 & 0.768 & & \ul{0.908} & 12.82 & 0.692 & 3.042 \\
\hspace{0.4cm}\textbf{Ours (PSAD)} & & 0.928 & \ul{17.18} & \ul{0.793} & & 0.921 & \ul{20.49} & \ul{0.896} & & 0.872 & \textbf{19.63} & \textbf{0.635} & & 0.906 & \ul{11.47} & \ul{0.619} & 2.987 \\
\midrule
\textit{Pose Keypoints} \\
\hspace{0.4cm}\textbf{Ours (NoDisc)} & & 0.926 & 18.84 & 0.943 & & 0.915 & 22.48 & 0.943 & & 0.873 & 23.71 & 0.937 & & 0.907 & 13.51 & 0.704 & 3.041 \\
\hspace{0.4cm}\textbf{Ours (Binary)} & & 0.925 & 18.39 & 0.872 & & 0.914 & 22.52 & 0.98 & & 0.871 & 22.35 & 0.816 & & 0.906 & 12.89 & 0.645 & 3.017 \\
\hspace{0.4cm}\textbf{Ours (Patch)} & & \textbf{0.931} & 18.40 & 0.841 & & \textbf{0.923} & 21.46 & 0.955 & & \textbf{0.876} & 21.94 & 0.814 & & \textbf{0.909} & 12.53 & 0.666 & \ul{3.043} \\
\hspace{0.4cm}\textbf{Ours (PSAD)} & & 0.928 & \textbf{17.04} & \textbf{0.762} & & 0.921 & \textbf{20.04} & \textbf{0.795} & & 0.872 & \ul{20.98} & \ul{0.672} & & 0.906 & \textbf{11.40} & \textbf{0.570} & 3.036 \\
\bottomrule
\end{tabular}
}
\vspace{-0.35cm}
\end{table}

\subsection{Experimental Setup}
\tinytit{Datasets}
First, we perform experiments on our newly proposed dataset, Dress Code, using 48,392 image pairs as training set and the remaining as test set (\ie~5,400 pairs, 1,800 for each category). During evaluation, image pairs of the test set are rearranged to form unpaired pairs of clothes and front-view models. 
On Dress Code, we use three different image resolutions: $256\times192$ (\ie~the one typical used by virtual try-on models), $512\times384$, and $1024\times768$.
Following our experiments on Dress Code, we evaluate our model on the standard VITON dataset~\cite{han2018viton}, composed of 16,253 image pairs. We employ this dataset to evaluate our solution in comparison with other state-of-the-art architectures on a widely-employed benchmark. In VITON, all images have a resolution of $256\times192$ and are divided into training and test set with 14,221 and 2,032 image pairs respectively. 

\tit{Evaluation metrics} Following recent literature, we employ evaluation metrics that either compare the generated images with the corresponding ground-truths, \ie~Structural Similarity (SSIM), or measure the realism and the visual quality of the generation, \ie~Frechét Inception Distance (FID)~\cite{heusel2017gans}, Kernel Inception Distance (KID)~\cite{binkowski2018demystifying}, and Inception Score (IS)~\cite{salimans2016improved}.

\tit{Training}
We train the three components of our model separately. Specifically, we first train the warping module and then the human parsing estimation module for 100k and 50k iterations respectively. Finally, we train the try-on module for other 150k iterations. We set the weight of the second-order difference constraint $\lambda_{\textit{const}}$ to $0.01$ and the weight of the adversarial loss $\lambda_{\textit{adv}}$ to $0.1$. All experiments are performed using Adam~\cite{kingma2015adam} as optimizer and a learning rate equal to $10^{-4}$. More details on the architecture and training stage are reported in the supplementary material.

\subsection{Experiments on Dress Code}

\tinytit{Baselines and Competitors}
In this set of experiments, we compare with CP-VTON~\cite{wang2018toward}, CP-VTON+~\cite{minar2020cpvton}, CIT~\cite{ren2021cloth}, VITON-GT~\cite{fincato2020viton}, WUTON~\cite{issenhuth2019end}, ACGPN~\cite{yang2020towards}, and PF-AFN~\cite{ge2021parser}, that we re-train from scratch on our dataset using source codes provided by the authors, when available, or our re-implementations. In addition to these methods, we implement an improved version of~\cite{wang2018toward} (\ie~CP-VTON$^\dagger$) and of~\cite{ren2021cloth} (\ie~CIT$^\dagger$) in which we use, as an additional input to the model, the person representation $m$.  
To validate the effectiveness of the Pixel-level Semantic Aware Discriminator (PSAD), we also test a model trained with a patch-based discriminator~\cite{isola2017image} (Patch), a model trained by removing in our discriminator the $N$ semantic channels and only keeping the real/fake one~\cite{schonfeld2020u} (Binary), and a baseline trained without the adversarial loss (NoDisc).

\begin{figure}[t]
\scriptsize
\centering
\setlength{\tabcolsep}{.2em}
\resizebox{\linewidth}{!}{
\begin{tabular}{ccc c ccc c ccc}
&  \textbf{Ours} & \textbf{Ours} & & & \textbf{Ours} & \textbf{Ours} & & & \textbf{Ours} & \textbf{Ours} \\
&  \textbf{(Patch)} & \textbf{(PSAD)} & & & \textbf{(Patch)} & \textbf{(PSAD)} & & & \textbf{(Patch)} & \textbf{(PSAD)} \\
\addlinespace[0.08cm]
\includegraphics[width=0.16\linewidth]{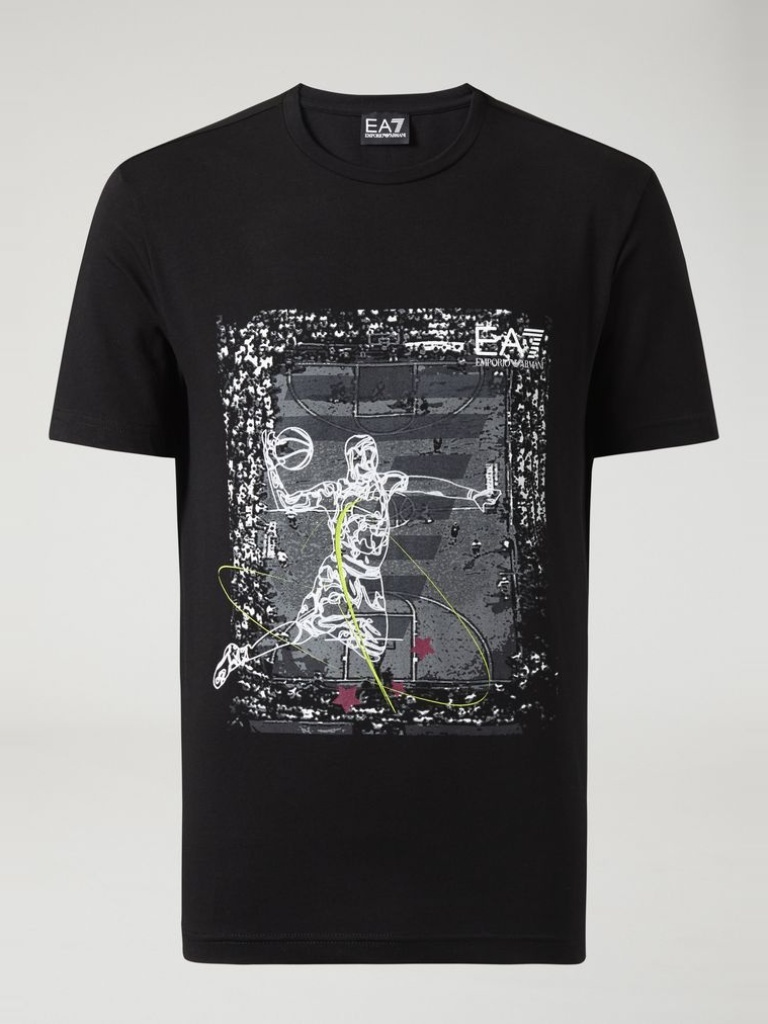} & 
\includegraphics[width=0.16\linewidth]{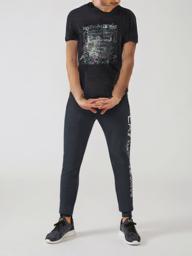} & 
\includegraphics[width=0.16\linewidth]{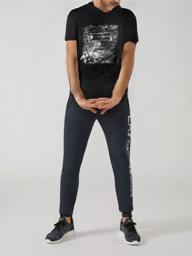} & &
\includegraphics[width=0.16\linewidth]{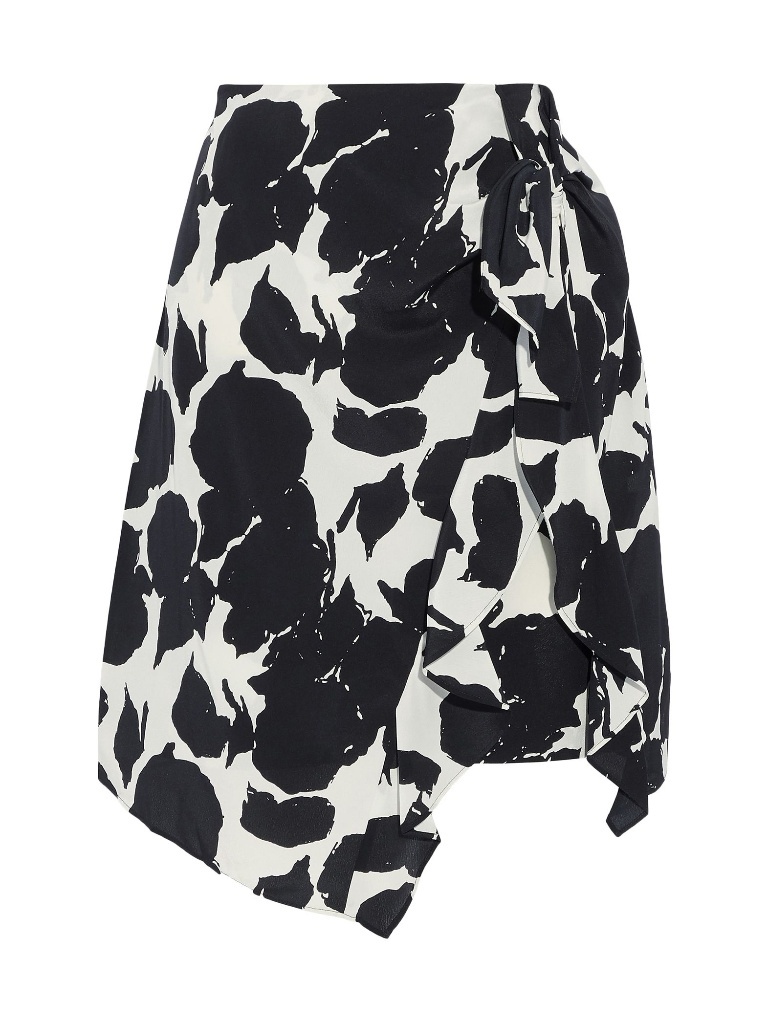} & 
\includegraphics[width=0.16\linewidth]{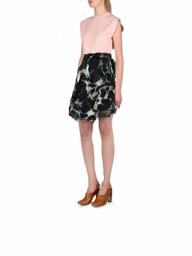} & 
\includegraphics[width=0.16\linewidth]{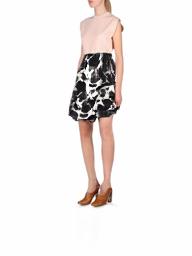} & &
\includegraphics[width=0.16\linewidth]{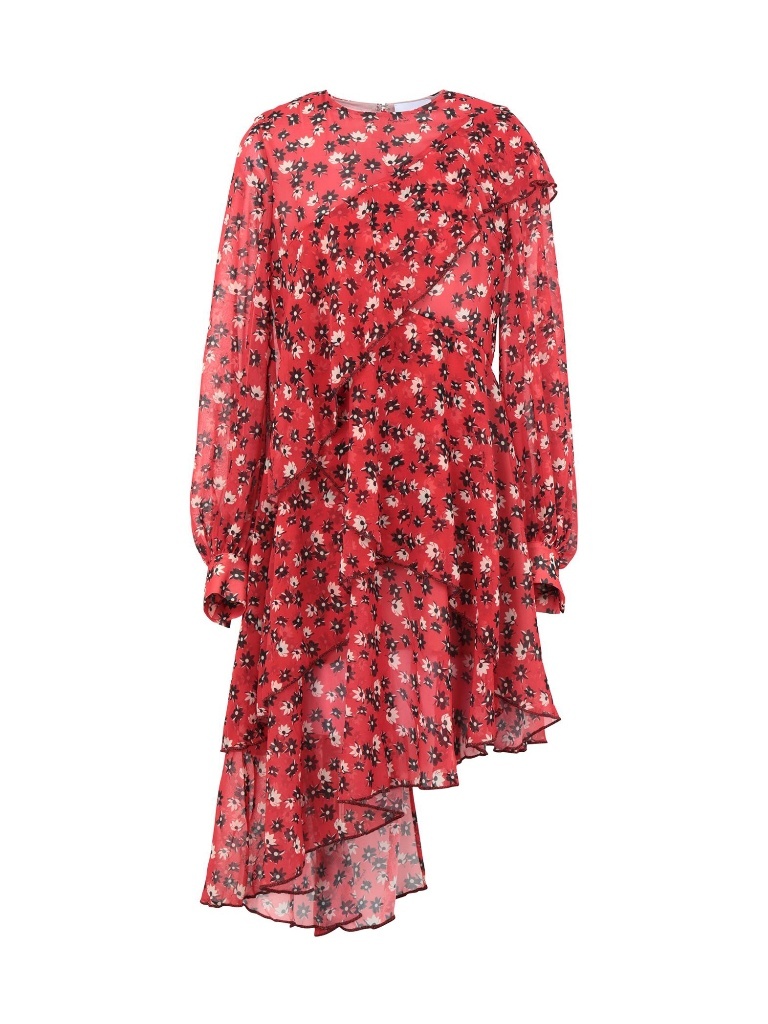} & 
\includegraphics[width=0.16\linewidth]{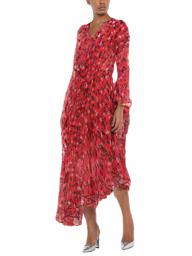} & 
\includegraphics[width=0.16\linewidth]{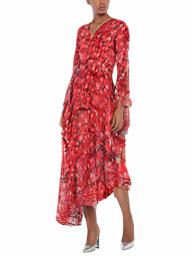} \\
\includegraphics[width=0.16\linewidth]{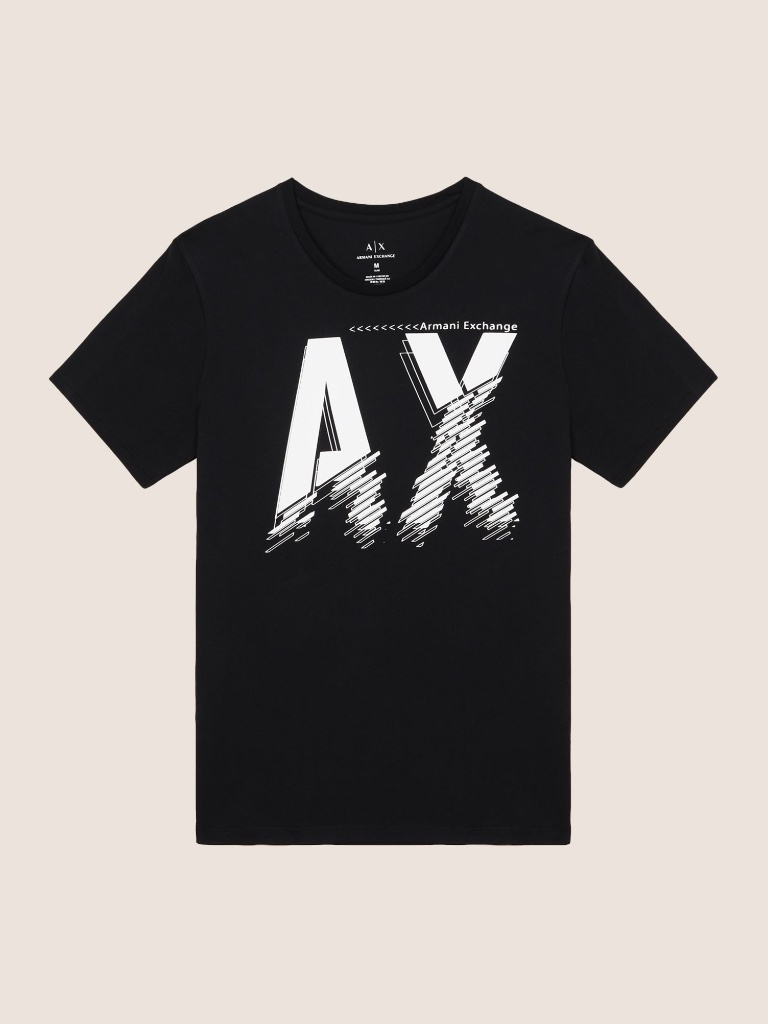} & 
\includegraphics[width=0.16\linewidth]{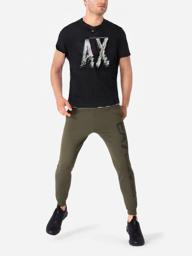} & 
\includegraphics[width=0.16\linewidth]{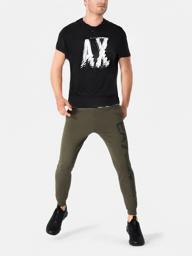} & &
\includegraphics[width=0.16\linewidth]{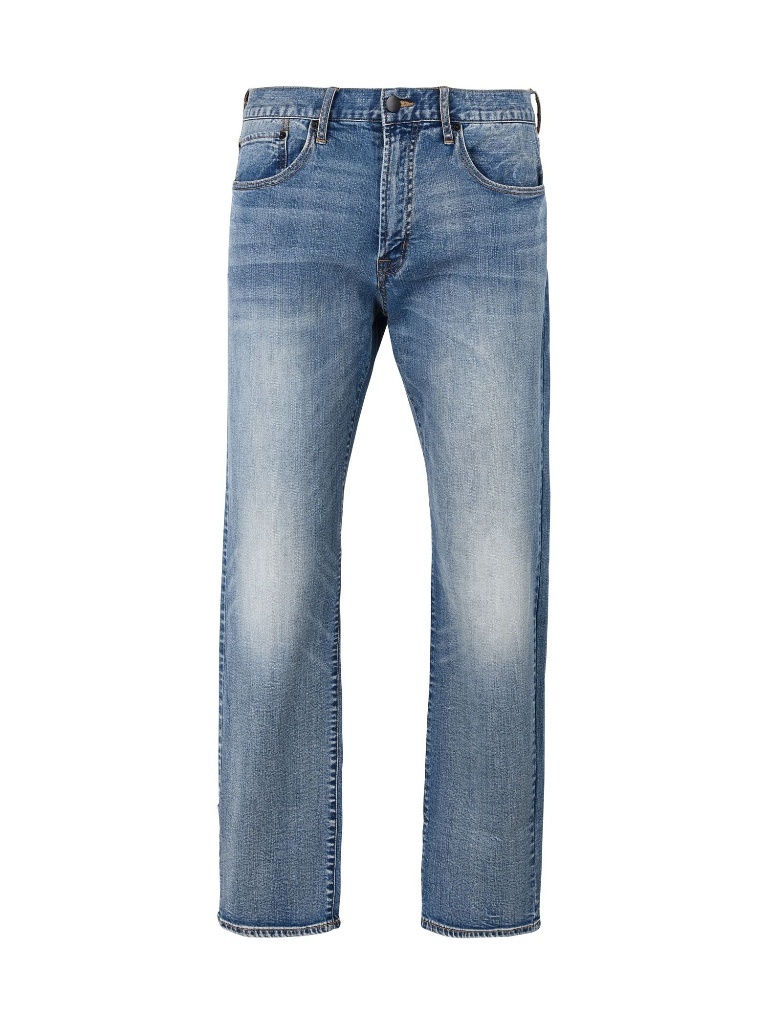} & 
\includegraphics[width=0.16\linewidth]{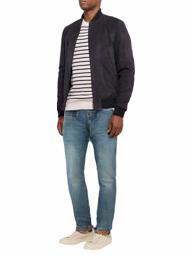} & 
\includegraphics[width=0.16\linewidth]{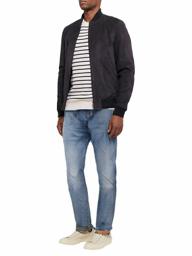} & &
\includegraphics[width=0.16\linewidth]{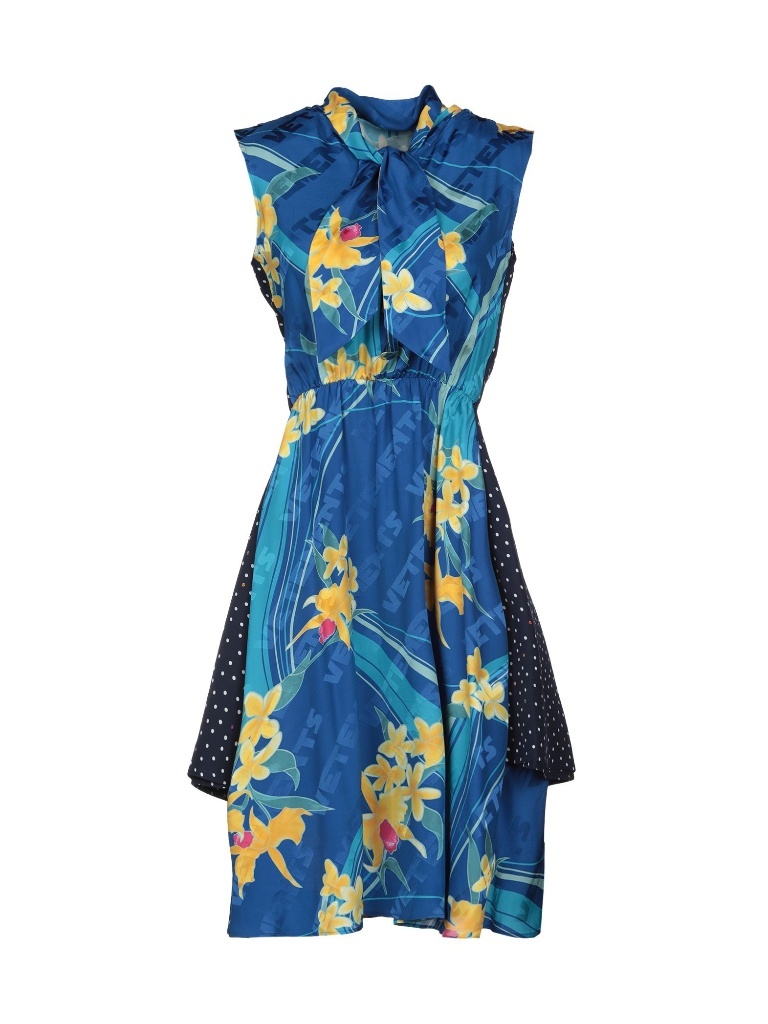} & 
\includegraphics[width=0.16\linewidth]{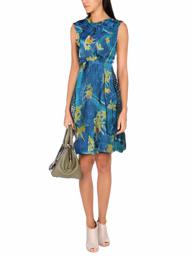} & 
\includegraphics[width=0.16\linewidth]{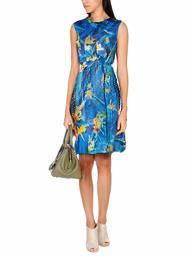} \\
\end{tabular}
}
\caption{Qualitative comparison between Patch and PSAD.}
\label{fig:comparison_hpad}
\vspace{-0.2cm}
\end{figure}

\begin{figure}[t]
\centering
\scriptsize
\setlength{\tabcolsep}{.2em}
\resizebox{\linewidth}{!}{
\begin{tabular}{ccccc c ccccc}
& & \textbf{WUTON} & \textbf{ACGPN} & \textbf{Ours} & & & & \textbf{WUTON} & \textbf{ACGPN} & \textbf{Ours} \\
& & \cite{issenhuth2019end} & \cite{yang2020towards} & \textbf{(PSAD)} & & & &  \cite{issenhuth2019end} & \cite{yang2020towards} & \textbf{(PSAD)} \\
\addlinespace[0.08cm]
\includegraphics[width=0.16\linewidth]{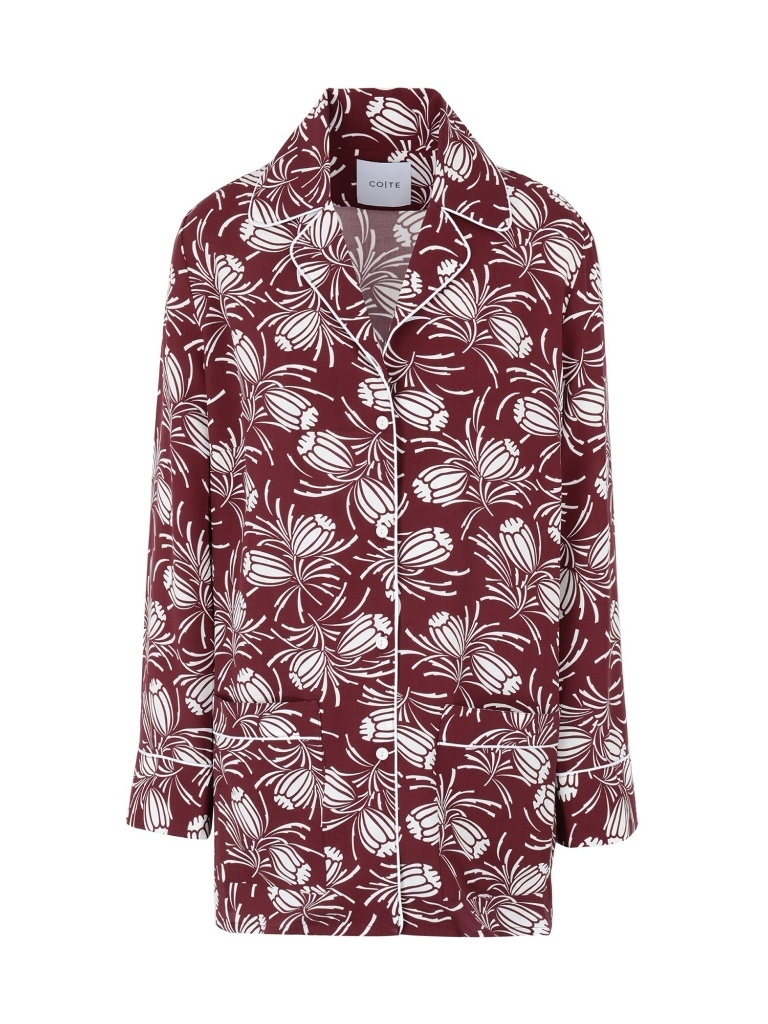} &
\includegraphics[width=0.16\linewidth]{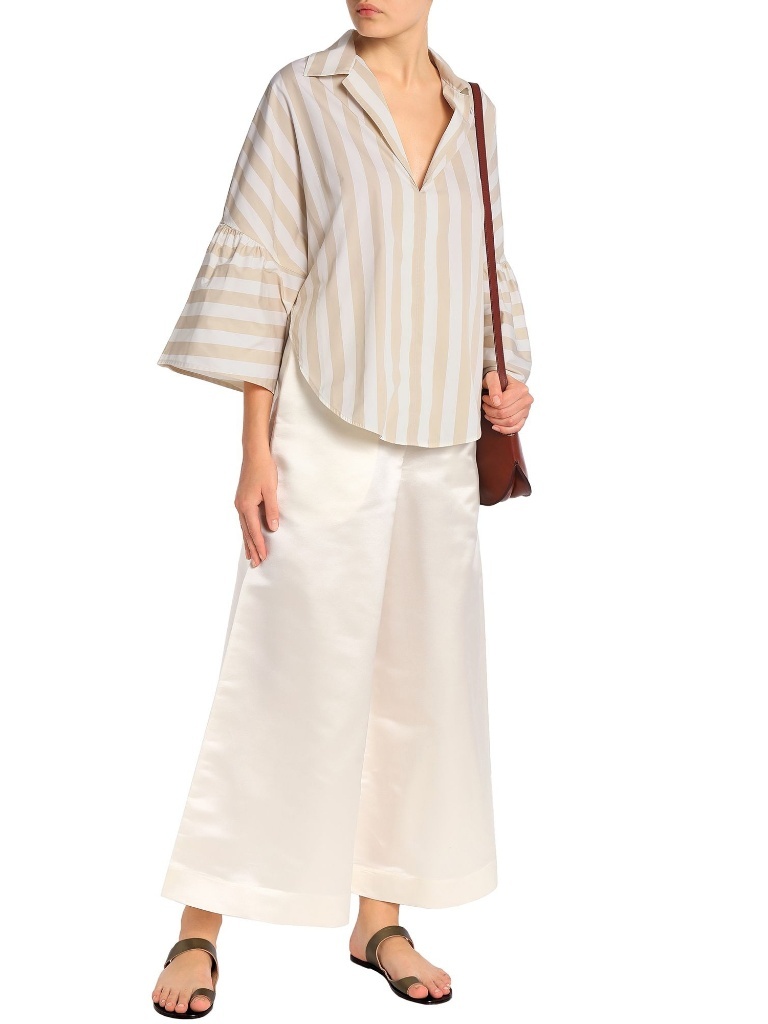} & 
\includegraphics[width=0.16\linewidth]{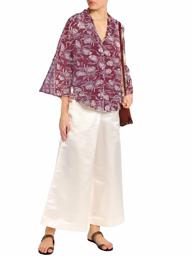} &
\includegraphics[width=0.16\linewidth]{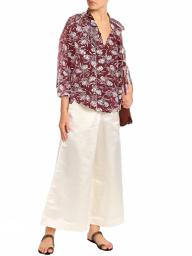} &
\includegraphics[width=0.16\linewidth]{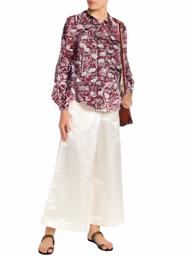} & &
\includegraphics[width=0.16\linewidth]{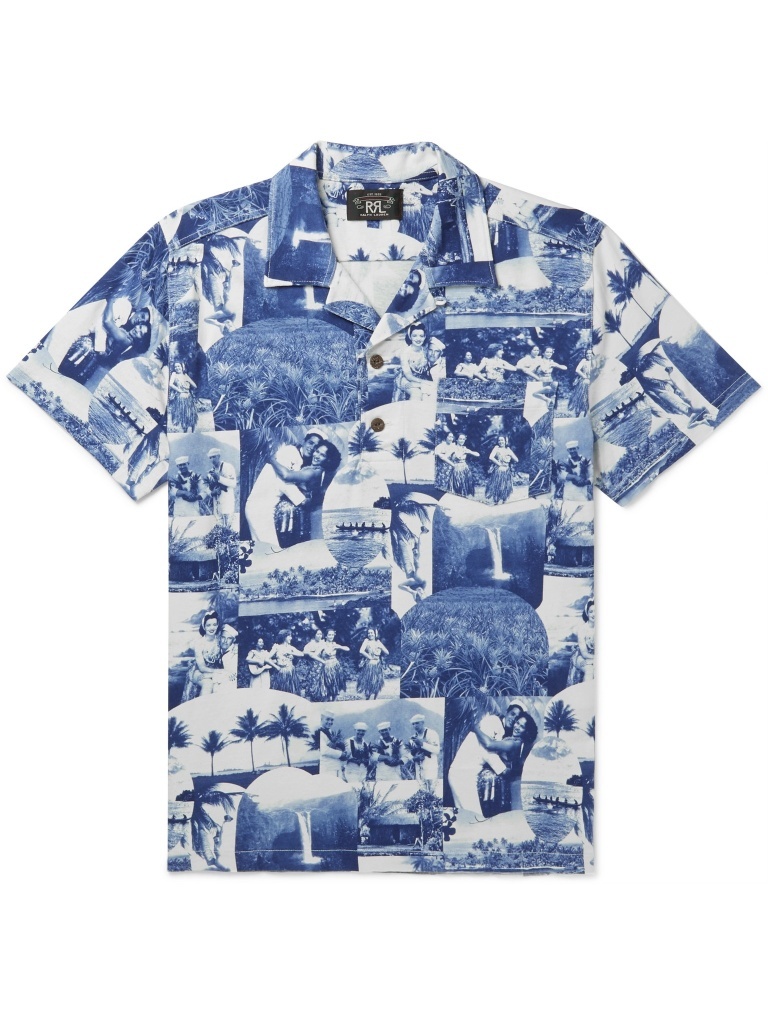} &
\includegraphics[width=0.16\linewidth]{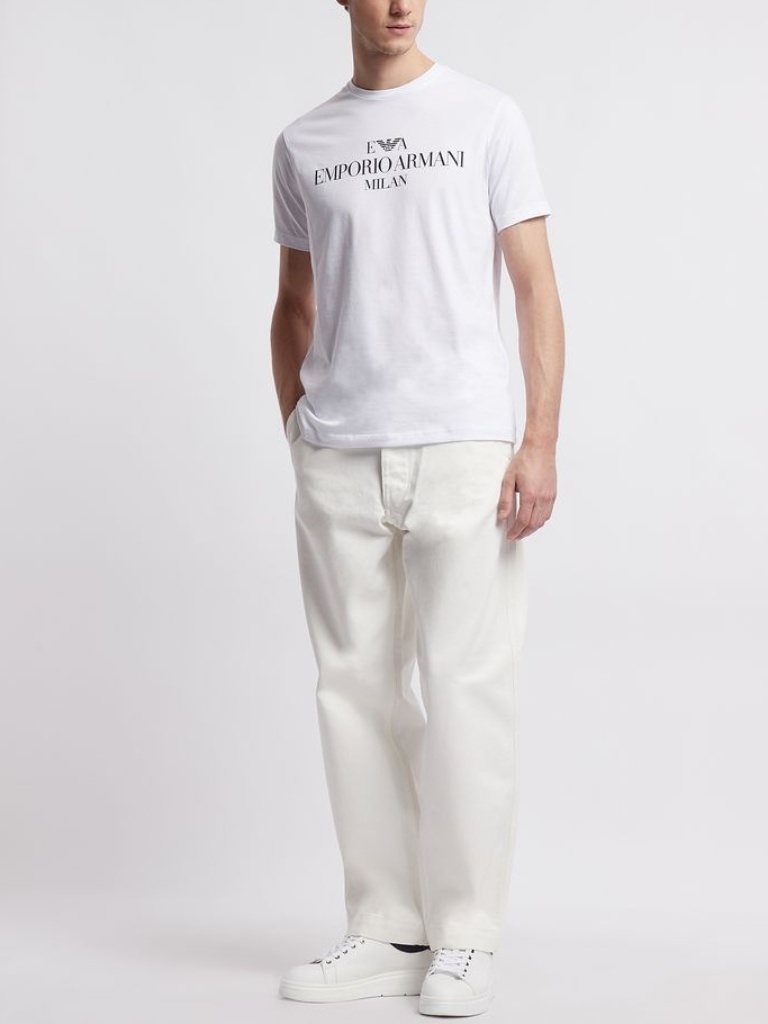} & 
\includegraphics[width=0.16\linewidth]{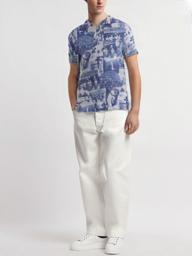} &
\includegraphics[width=0.16\linewidth]{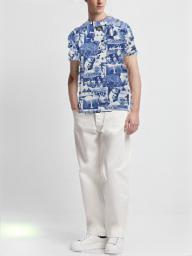} &
\includegraphics[width=0.16\linewidth]{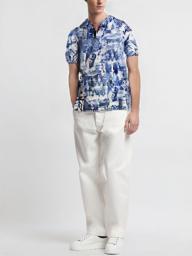} \\
\includegraphics[width=0.16\linewidth]{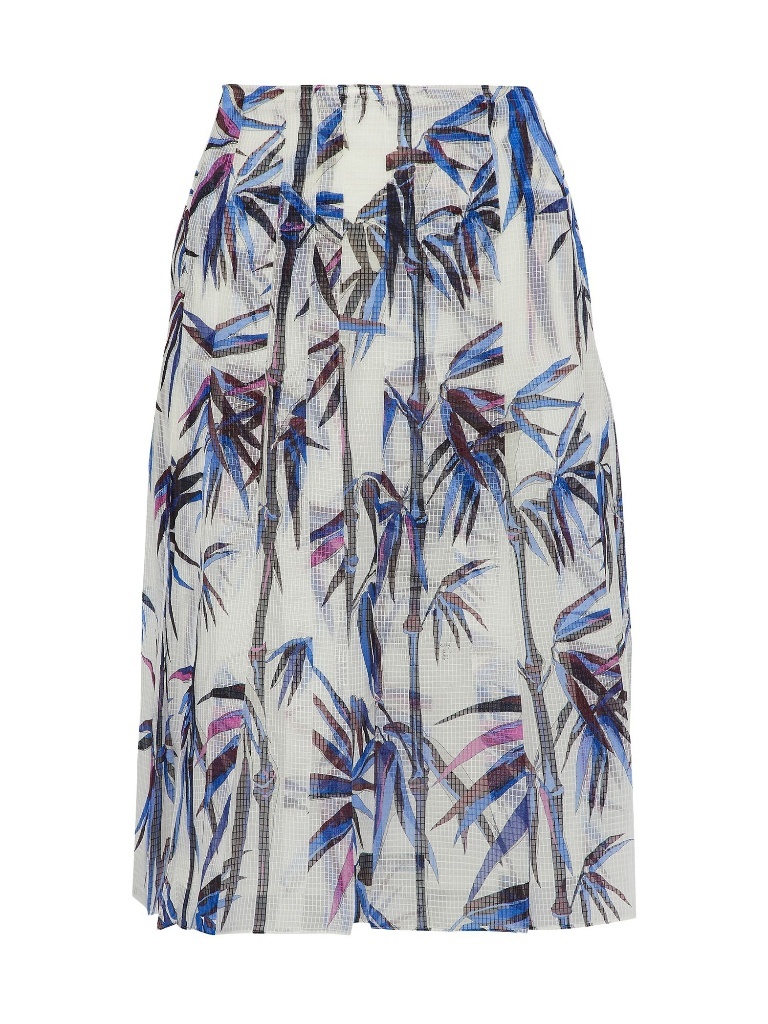} &
\includegraphics[width=0.16\linewidth]{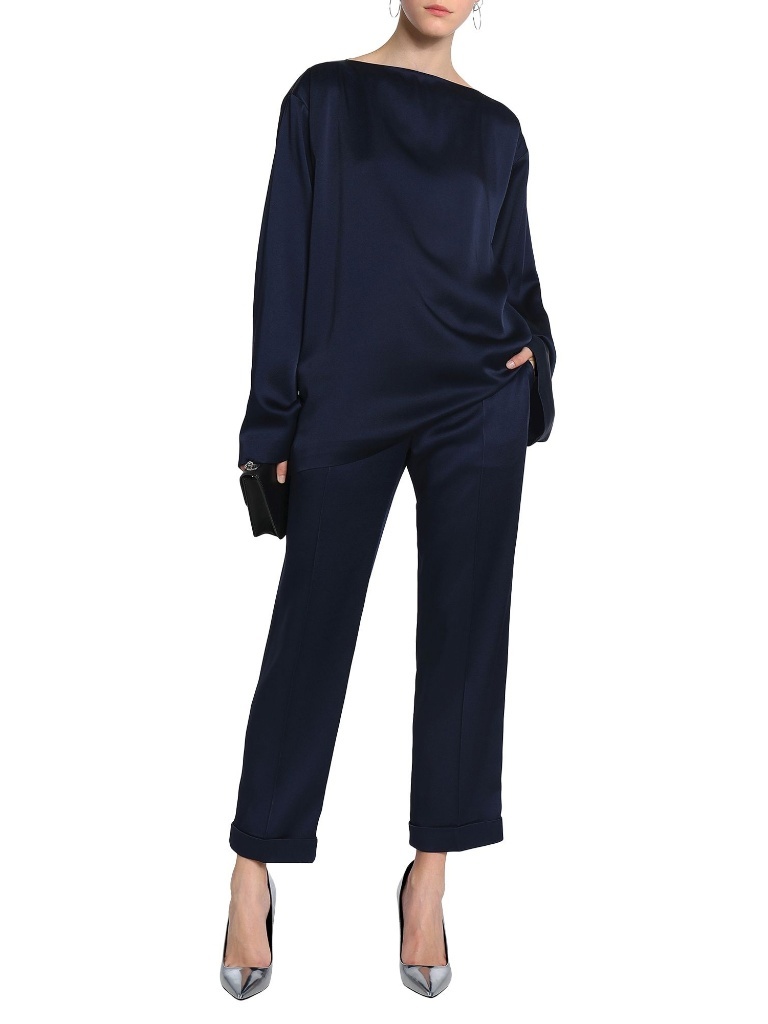} & 
\includegraphics[width=0.16\linewidth]{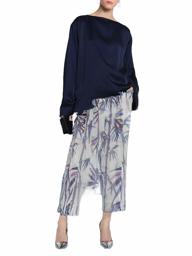} &
\includegraphics[width=0.16\linewidth]{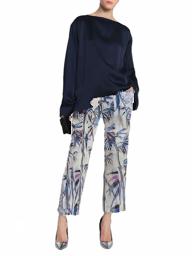} &
\includegraphics[width=0.16\linewidth]{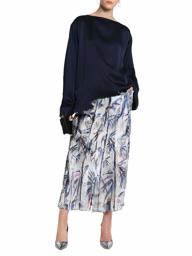} & &
\includegraphics[width=0.16\linewidth]{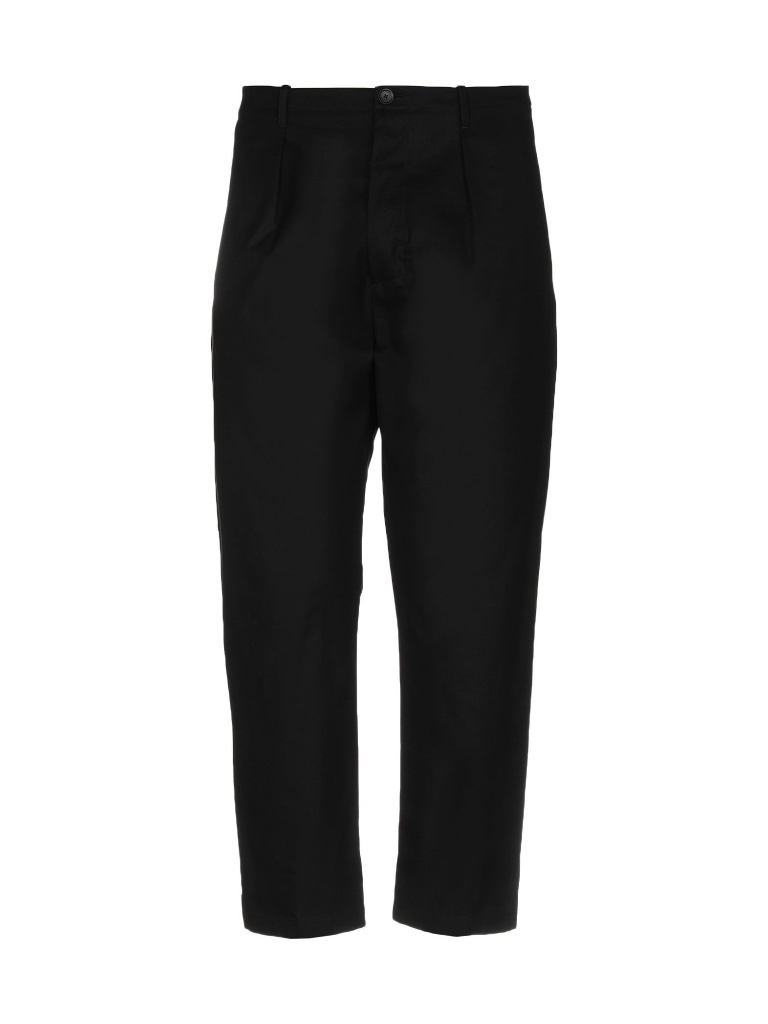} &
\includegraphics[width=0.16\linewidth]{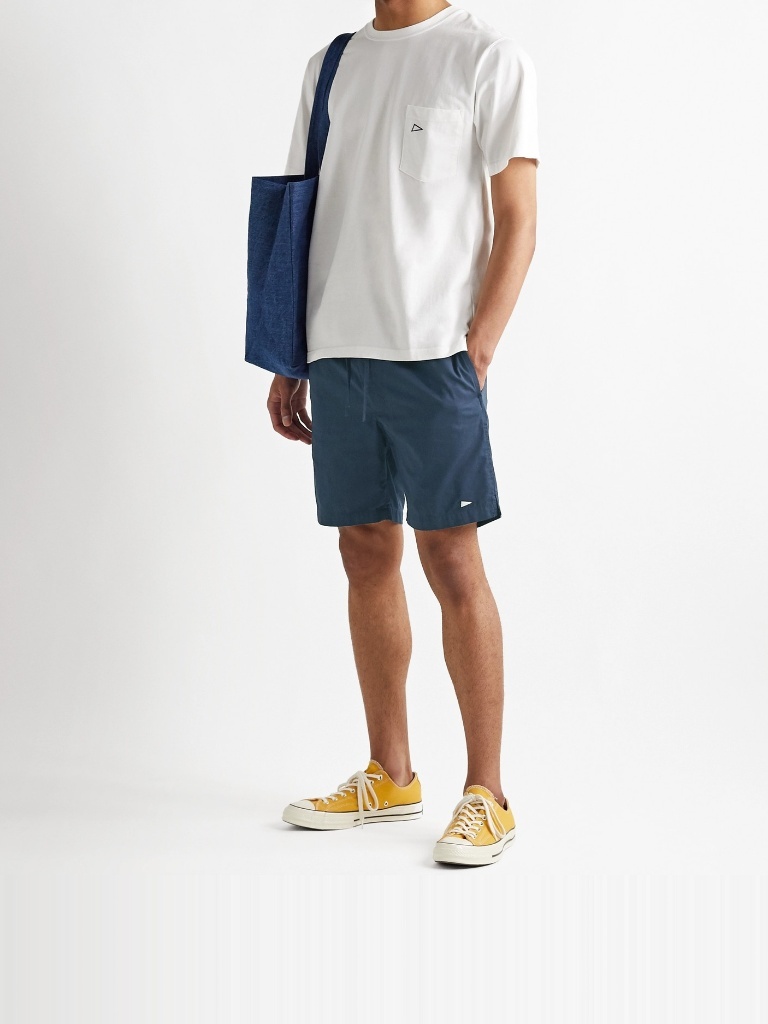} & 
\includegraphics[width=0.16\linewidth]{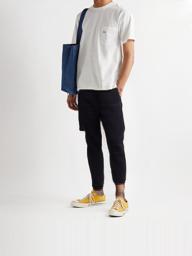} &
\includegraphics[width=0.16\linewidth]{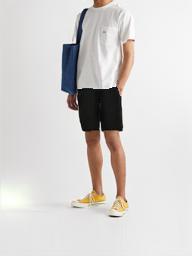} &
\includegraphics[width=0.16\linewidth]{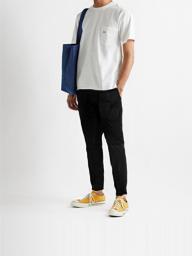} \\
\includegraphics[width=0.16\linewidth]{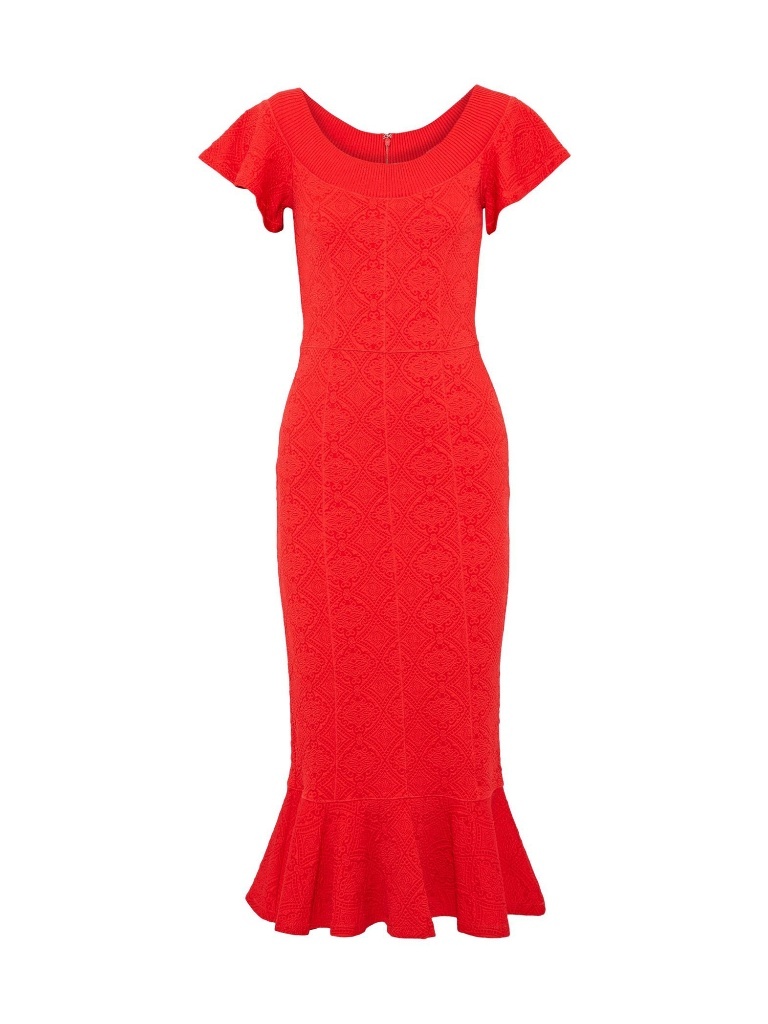} &
\includegraphics[width=0.16\linewidth]{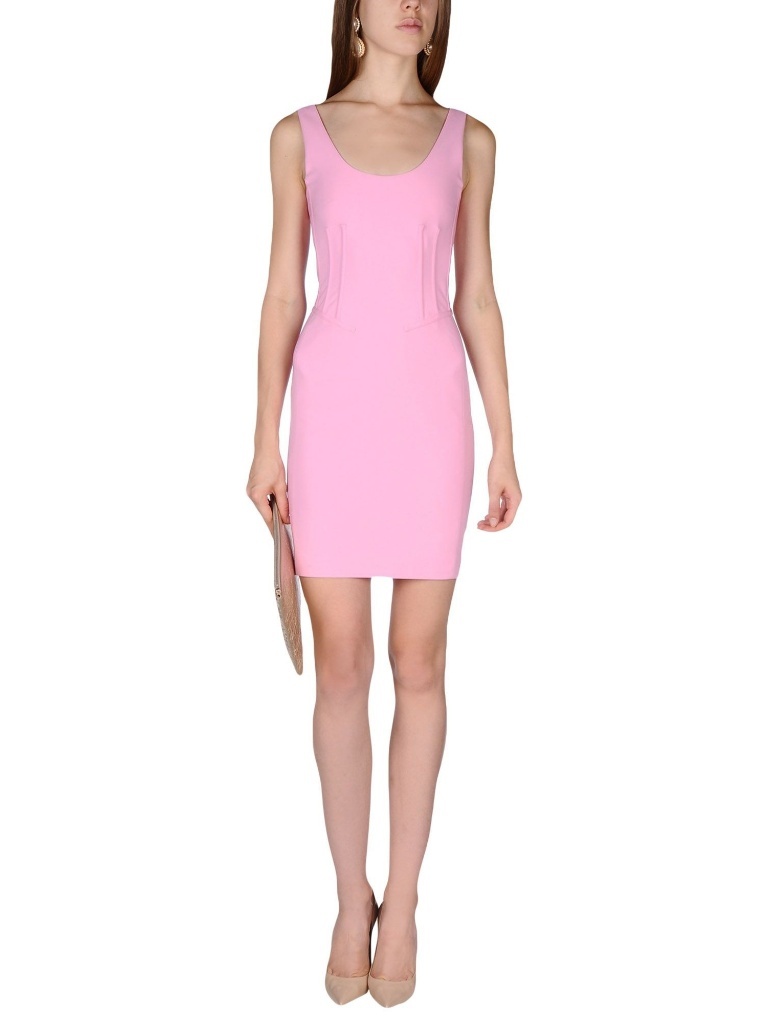} & 
\includegraphics[width=0.16\linewidth]{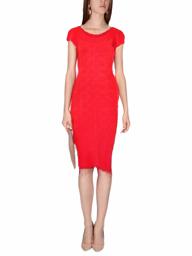} &
\includegraphics[width=0.16\linewidth]{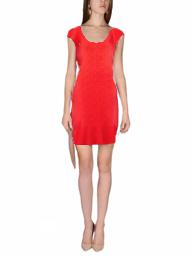} &
\includegraphics[width=0.16\linewidth]{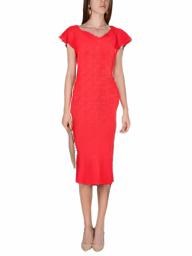} & & 
\includegraphics[width=0.16\linewidth]{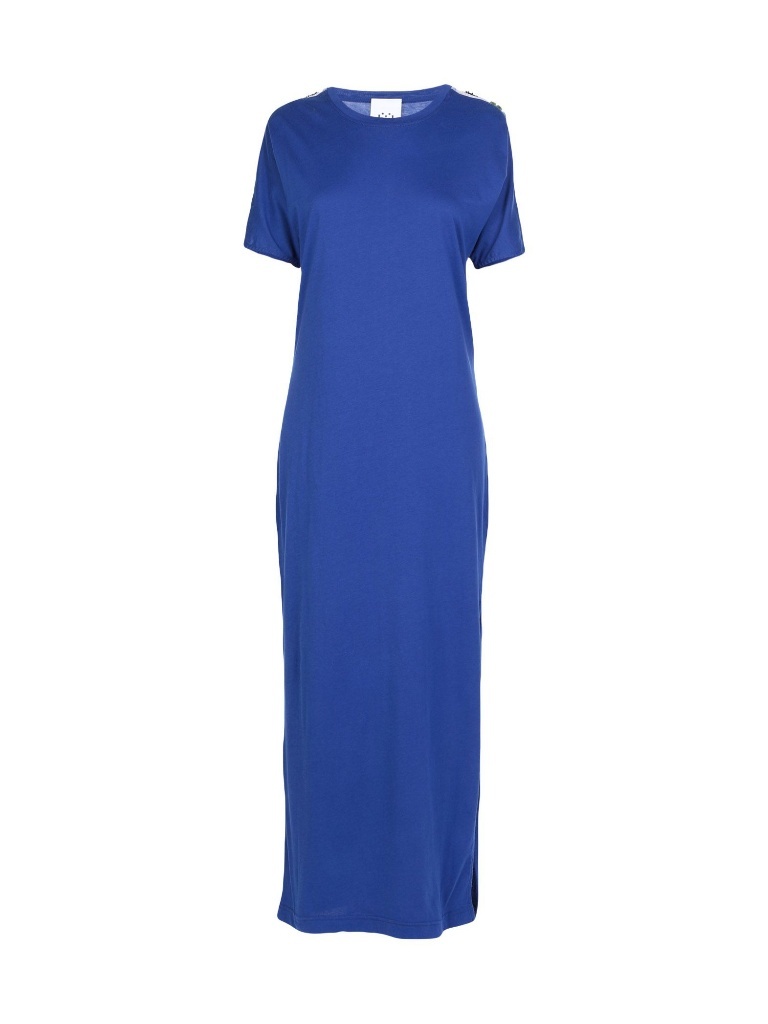} &
\includegraphics[width=0.16\linewidth]{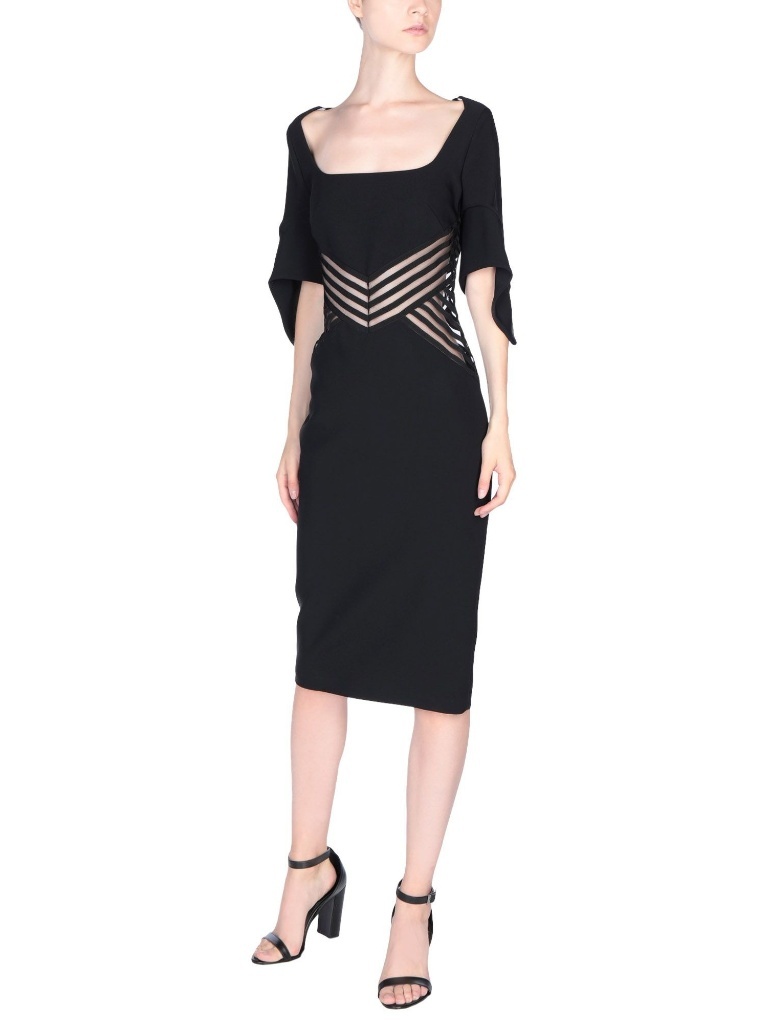} & 
\includegraphics[width=0.16\linewidth]{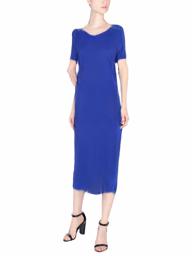} &
\includegraphics[width=0.16\linewidth]{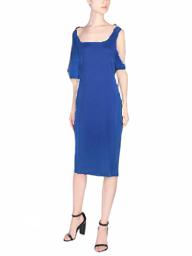} &
\includegraphics[width=0.16\linewidth]{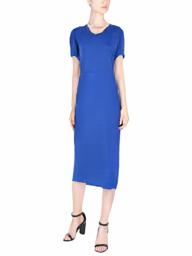} \\
\end{tabular}
}
\caption{Sample try-on results on the Dress Code test set.}
\label{fig:dress_code_results}
\vspace{-0.35cm}
\end{figure}

\tit{Low-Resolution Results and Ablative Analysis} 
In this experiment, we compare our complete model (PSAD) with state-of-the-art architectures for virtual try-on and with the Patch, Binary, and NoDisc baselines. For these comparisons, we consider the standard resolution for virtual try-on ($256 \times 192$). In Table~\ref{tab:try-on_res}, we report numerical results on the Dress Code test set. As it can be seen, our model obtains better results than competitors on all clothing categories in terms of almost all considered evaluation metrics. Quantitative results also confirm the effectiveness of PSAD in comparison with a pixel-level discriminator without semantics and with a standard patch-based discriminator, especially in terms of the realism of the generated images (\ie~FID and KID). PSAD is second to the Patch model only in terms of SSIM, and by a very limited margin. All discriminator-based configurations outperform the NoDisc baseline, thus showing the importance of incorporating a discriminator in a virtual try-on architecture. 
We also test the effect of using different representations for the body pose (human keypoints and dense pose). When comparing the two versions of our model, we find out that using dense pose helps to deal with full-body clothes (\ie~dresses), but does not bring a consistent improvement over the use of human keypoints in our architecture. For this reason, we keep the latter model version for all the next experiments. 
In Fig.~\ref{fig:comparison_hpad}, we report a qualitative comparison between the results obtained with our Patch model and the PSAD version.
In Fig.~\ref{fig:dress_code_results}, we compare our results with those obtained by state-of-the-art competitors. Overall, our model with PSAD can better preserve the characteristics of the original clothes such as colors, textures, and shapes, and reduce artifacts and distortions, increasing the realism and visual quality of the generated images.

\begin{table}[t]
\centering
\footnotesize
\caption{High-resolution results on the Dress Code test set.}
\label{tab:try-on_res_hd}
\setlength{\tabcolsep}{.35em}
\resizebox{0.82\linewidth}{!}{
\begin{tabular}{lc cccc c cccc}
\toprule
& & \multicolumn{4}{c}{$512\times384$} & & \multicolumn{4}{c}{$1024\times768$} \\
\cmidrule{3-6} \cmidrule{8-11}
\textbf{Model} & & \textbf{SSIM} $\uparrow$ & \textbf{FID} $\downarrow$ & \textbf{KID} $\downarrow$ & \textbf{IS} $\uparrow$ & & \textbf{SSIM} $\uparrow$ & \textbf{FID} $\downarrow$ & \textbf{KID} $\downarrow$ & \textbf{IS} $\uparrow$ \\
\midrule
CP-VTON~\cite{wang2018toward} & & 0.831 & 29.24 & 1.671 & 3.096 & & 0.853 & 36.68 & 2.379 & 3.155  \\
CP-VTON$^\dagger$~\cite{wang2018toward} & & 0.896 & 10.08 & 0.425 & 3.277 & & 0.912 & 9.96 & 0.338 & 3.300 \\
\textbf{Ours (Patch)} & & \textbf{0.923} & 9.44 & \textbf{0.246} & 3.310 & & \textbf{0.922} & 9.99 & 0.370 & 3.344 \\
\textbf{Ours (PSAD)} & & 0.916 & \textbf{7.27} & 0.394 & \textbf{3.320} & & 0.919 & \textbf{7.70} & \textbf{0.236} & \textbf{3.357} \\
\bottomrule
\end{tabular}
}
\vspace{-0.2cm}
\end{table}

\begin{table}[t]
\centering
\footnotesize
\caption{Multi-garment try-on results on the Dress Code test set.}
\label{tab:multi_garment}
\setlength{\tabcolsep}{.45em}
\resizebox{0.56\linewidth}{!}{
\begin{tabular}{lc cccc}
\toprule
\textbf{Model} & & \textbf{Resolution} & \textbf{FID} $\downarrow$ & \textbf{KID} $\downarrow$ & \textbf{IS} $\uparrow$ \\
\midrule
CP-VTON$^\dagger$~\cite{wang2018toward} & & $256\times192$ & 30.29 & 1.935 & \textbf{2.912} \\
VITON-GT~\cite{fincato2020viton}        & & $256\times192$ & 21.06 & 1.176 & 2.762 \\
WUTON~\cite{issenhuth2019end}           & & $256\times192$ & 20.13 & 1.084 & 2.753 \\
\textbf{Ours (Patch)}                   & & $256\times192$ & 19.86 & 1.006 & 2.784 \\
\textbf{Ours (PSAD)}                    & & $256\times192$ & \textbf{17.52} & \textbf{0.749} & 2.832 \\
\midrule
CP-VTON$^\dagger$~\cite{wang2018toward}     & & $512\times384$ & 22.96 & 1.327 & \textbf{3.273} \\
\textbf{Ours (Patch)}                    & & $512\times384$ & 21.90 & 1.155 & 3.073 \\
\textbf{Ours (PSAD)}                     & & $512\times384$ & \textbf{16.90} & \textbf{0.690} & 3.160 \\
\midrule
CP-VTON$^\dagger$~\cite{wang2018toward}     & & $1024\times768$ & 23.30 & 1.393 & 3.261 \\
\textbf{Ours (Patch)}                    & & $1024\times768$ & 20.26 & 0.841 & \textbf{3.498} \\
\textbf{Ours (PSAD)}                     & & $1024\times768$ & \textbf{17.19} & \textbf{0.681} & 3.340 \\
\bottomrule
\end{tabular}
}
\vspace{-0.35cm}
\end{table}

\tit{High-Resolution Results} 
For this experiment, we train and test our models and competitors using higher-resolution images ($512 \times 384$ and $1024 \times 768$). We compare with CP-VTON~\cite{wang2018toward} and its improved version (CP-VTON$^\dagger$). Quantitative results for this setting are reported in Table~\ref{tab:try-on_res_hd} and refer to the entire test set of the Dress Code dataset. As it can be seen, our method outperforms the competitors. When generating images with resolution $1024 \times 768$, PSAD achieves the best results in terms of FID, KID, and IS with respect to the competitors and the Patch baseline.

\begin{table}[t]
\centering
\footnotesize
\caption{User study results. Our model is always preferred more than 50\% of the time.}
\label{tab:user_study}
\setlength{\tabcolsep}{.35em}
\resizebox{0.9\linewidth}{!}{
\begin{tabular}{lccccccc}
\toprule
& & CP-VTON & VITON-GT & WUTON & ACGPN & PF-AFN & Ours (Patch)\\
\midrule
\textbf{Realism} & & 
10.1 / \textbf{89.9} & 
46.4 / \textbf{53.6} & 
42.0 / \textbf{58.0} & 
35.9 / \textbf{64.1} & 
29.4 / \textbf{70.6} & 
34.8 / \textbf{65.2} \\
\midrule
\textbf{Coherency} & & 
11.5 / \textbf{88.5} & 
32.1 / \textbf{67.9} & 
41.6 / \textbf{58.4} &
23.1 / \textbf{76.9} & 
25.0 / \textbf{75.0} & 
36.9 / \textbf{63.1} \\
\bottomrule
\end{tabular}
}
\vspace{-.2cm}
\end{table}

\begin{figure}[t]
\centering
\includegraphics[width=\linewidth]{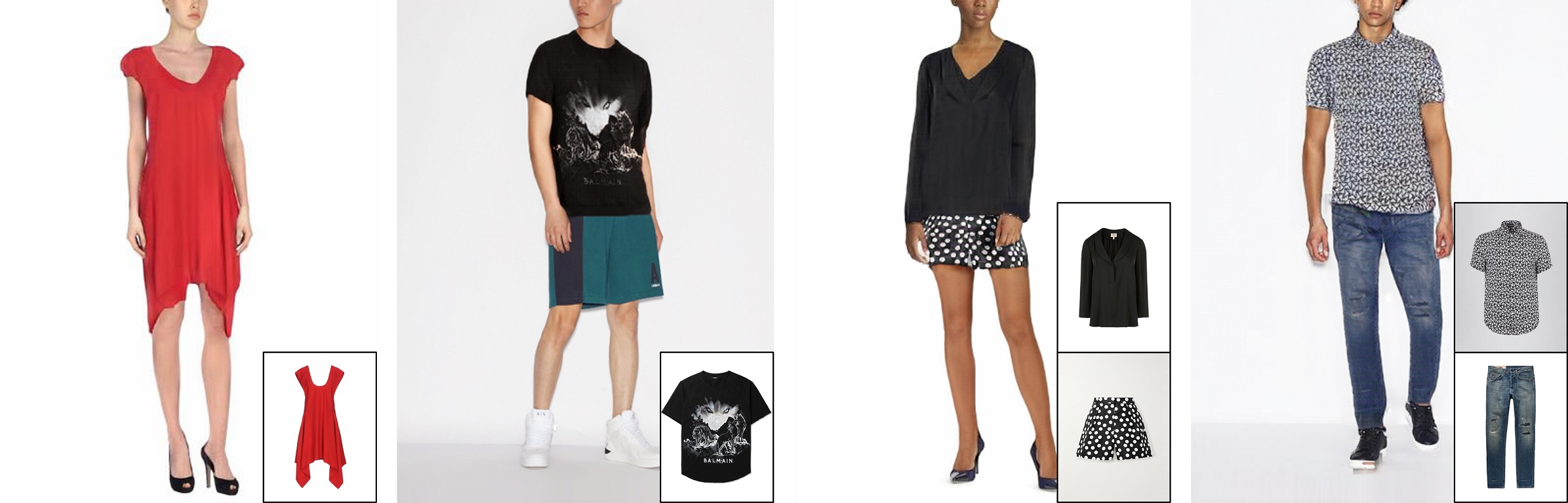}
\caption{High-resolution results on the Dress Code test set in both single- and multi-garment try-on settings.}
\label{fig:hd_results}
\vspace{-0.35cm}
\end{figure}

\tit{Multi-Garment Try-On Results}
As an additional experiment on the Dress Code dataset, we propose a novel setting in which the try-on is performed twice: first with an upper-body garment, and then with a lower-body item. This fully-unpaired setting aims to push further the difficulty of image-based virtual try-on, as it doubles the number of operations required to generate the resulting image. We remind that this experiment would have not been possible on the standard VITON dataset~\cite{han2018viton}, as it contains only upper-body clothes. In Table~\ref{tab:multi_garment}, we report numerical results at varying image resolution.
We can observe that PSAD outperforms the competitors and baselines on almost all the metrics for all the different image resolutions, with the only exception of the IS metric. Notably, the improvement of PSAD with respect to the Patch baseline ranges from $2.34$ to $5.00$ and from $0.16$ to $0.46$ in terms of FID and KID respectively.

\tit{User Study}
While quantitative metrics used in the previous experiments can capture fine-grained variations in the generated images, the overall realism and visual quality of the results can be effectively assessed by human evaluation.
To further evaluate the quality of generated images, we conduct a user study measuring both the realism of our results and their coherence with the input try-on garment and reference person. In the first test (Realism test), we show two generated images, one generated by our model and the other by a competitor, and ask to select the more realistic one. In the second test (Coherency test), in addition to the two generated images, we include the images of the try-on garment and the reference person used as input to the try-on network. In this case, we ask the user to select the image that is more coherent with the given inputs. All images are randomly selected from the Dress Code test set. Overall, this study involves a total of 30 participants, including researchers and non-expert people, and we collect more than 3,000 different evaluations (\ie~1,500 for each test). Results are shown in Table~\ref{tab:user_study}. For each test, we report the percentage of votes obtained by the competitor / by our model. We also include a comparison with the Patch baseline. Our complete model is always selected more than 50\% of the time against all considered competitors, thus further demonstrating the effectiveness of our solution.

\subsection{Experiments on VITON}
To conclude, we train the try-on networks on the widely used VITON dataset~\cite{han2018viton}. For this experiment, we compare our PSAD and Patch models with other state-of-the-art architectures. In particular, we report results from CP-VTON~\cite{wang2018toward} and CP-VTON+~\cite{minar2020cpvton} using source codes and pre-trained models provided by the authors. For SieveNet~\cite{jandial2020sievenet}, ACGPN~\cite{yang2020towards}, and DCTON~\cite{ge2021disentangled}, we use the results reported in the papers. Table~\ref{tab:viton_res} shows the quantitative results on the test set, while in Fig.~\ref{fig:comparison_viton} we report four examples of the generated try-on results. Also in this setting, PSAD contributes to increasing the realism and visual quality of synthesized images.

\begin{table}[t]
\centering
\footnotesize
\caption{Try-on results on the VITON test set~\cite{han2018viton}. Note that all models are trained exclusively on VITON.}
\label{tab:viton_res}
\setlength{\tabcolsep}{.45em}
\resizebox{0.7\linewidth}{!}{
\begin{tabular}{lc ccccc}
\toprule
\textbf{Model} & & \textbf{Resolution} & \textbf{SSIM} $\uparrow$ & \textbf{FID} $\downarrow$ & \textbf{KID} $\downarrow$ & \textbf{IS} $\uparrow$ \\
\midrule
CP-VTON~\cite{wang2018toward} & & $256\times192$ & 0.798 & 19.06 & 0.906 & 2.601 \\
CP-VTON+~\cite{minar2020cpvton} & & $256\times192$ & 0.828 & 16.31 & 0.784 & 2.821 \\
SieveNet~\cite{jandial2020sievenet} & & $256\times192$ & 0.766 & 14.65 & - & 2.820 \\
ACGPN~\cite{yang2020towards} & & $256\times192$ & 0.845 & - & - & 2.829 \\
DCTON~\cite{ge2021disentangled} & & $256\times192$ & 0.830 & 14.82 & - & \textbf{2.850} \\
\midrule
\textbf{Ours (Patch)} & & $256\times192$ & \textbf{0.893} & 14.76 & 0.495 & 2.733 \\
\textbf{Ours (PSAD)} & & $256\times192$ & 0.885 & \textbf{13.71} & \textbf{0.412} & 2.840 \\
\bottomrule
\end{tabular}
}
\vspace{-0.2cm}
\end{table}

\begin{figure}[t]
\small
\centering
\setlength{\tabcolsep}{.2em}
\resizebox{\linewidth}{!}{
\begin{tabular}{cc ccc c cc ccc}
& & \textbf{CP-VTON} & \textbf{ACGPN} & \textbf{Ours} & & & & \textbf{CP-VTON} & \textbf{ACGPN} & \textbf{Ours} \\
& & \cite{wang2018toward} & \cite{yang2020towards} & \textbf{(PSAD)} & & & & \cite{wang2018toward} & \cite{yang2020towards} & \textbf{(PSAD)} \\
\addlinespace[0.08cm]
\includegraphics[width=0.16\linewidth]{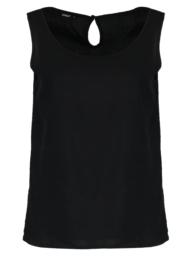} & 
\includegraphics[width=0.16\linewidth]{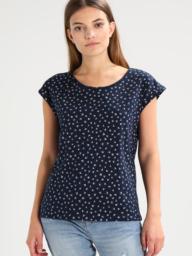} & 
\includegraphics[width=0.16\linewidth]{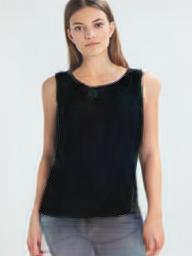} & 
\includegraphics[width=0.16\linewidth]{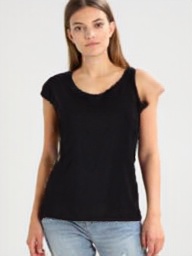} & 
\includegraphics[width=0.16\linewidth]{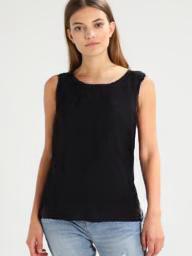} & &
\includegraphics[width=0.16\linewidth]{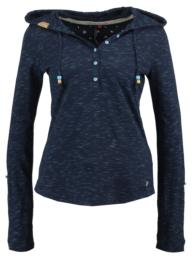} & 
\includegraphics[width=0.16\linewidth]{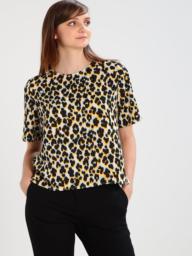} & 
\includegraphics[width=0.16\linewidth]{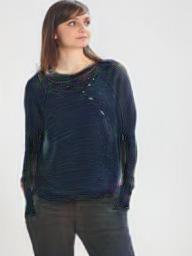} & 
\includegraphics[width=0.16\linewidth]{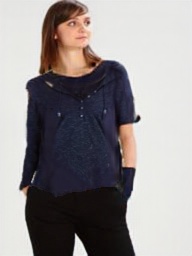} & 
\includegraphics[width=0.16\linewidth]{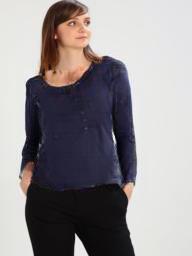} \\
\includegraphics[width=0.16\linewidth]{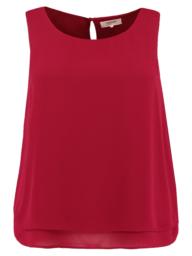} & 
\includegraphics[width=0.16\linewidth]{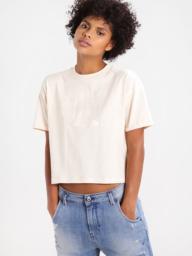} & 
\includegraphics[width=0.16\linewidth]{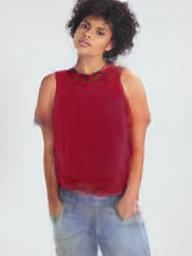} & 
\includegraphics[width=0.16\linewidth]{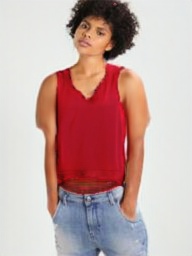} & 
\includegraphics[width=0.16\linewidth]{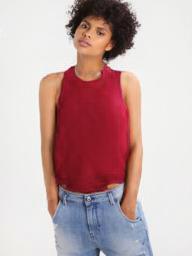} & & 
\includegraphics[width=0.16\linewidth]{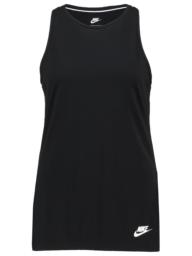} & 
\includegraphics[width=0.16\linewidth]{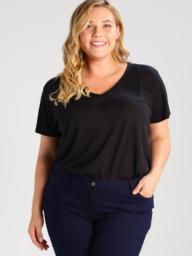} & 
\includegraphics[width=0.16\linewidth]{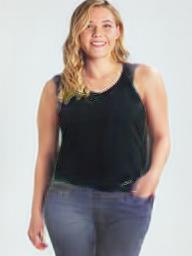} & 
\includegraphics[width=0.16\linewidth]{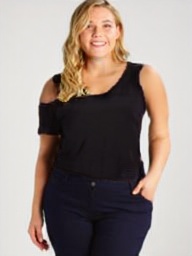} & 
\includegraphics[width=0.16\linewidth]{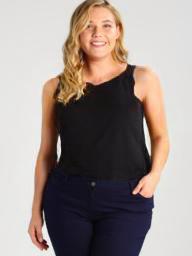} \\
\end{tabular}
}
\caption{Sample generated results on the VITON test set.}
\label{fig:comparison_viton}
\vspace{-.3cm}
\end{figure}

\section{Conclusion}
In this paper, we presented Dress Code: a new dataset for image-based virtual try-on. Dress Code, while being more than $3\times$ larger than the most common dataset for virtual try-on, is the first publicly available dataset for this task featuring clothes of multiple macro-categories and high-resolution images. We also presented a comprehensive benchmark with up to nine state-of-the-art virtual try-on approaches and different baselines, and introduced a Pixel-level Semantic-Aware Discriminator (PSAD) that improves the generation of high-quality images and the realism of the results.

\medskip\bigskip
\noindent\textbf{Acknowledgments.}
We thank CINECA, the Italian Supercomputing Center, for providing computational resources. This work has been supported by the PRIN project ``CREATIVE: CRoss-modal understanding and gEnerATIon of Visual and tExtual content'' (CUP B87G22000460001), co-funded by the Italian Ministry of University and Research.

\clearpage
%
%
\bibliographystyle{splncs04}
\bibliography{bibliography}

\clearpage
\section*{Supplementary Material}
\subsection*{Dress Code Dataset}
In Table~\ref{tab:stats}, we report the total number of images and the dimensions of the train/test splits in the Dress Code dataset. Additionally, we detail the dimension of each split with respect to the different macro-categories of the dataset (upper-body clothes, lower-body clothes, and dresses). In our experiments, we train our models on all training pairs of the dataset and test both on each category separately and on the entire test set. Additional sample image pairs from Dress Code are shown in Fig.~\ref{fig:dataset_upperbody},~\ref{fig:dataset_lowerbody}, and~\ref{fig:dataset_dresses} with the corresponding pose keypoints, dense pose of the reference model, and segmentation mask of the human body. As it can be seen, images of our dataset have a great variety considering both the body pose of the reference models and category and textures of try-on garments. This can lead to virtual try-on architectures becoming more general and adapting to more challenging scenarios.

\subsection*{Additional Implementation Details}

\tinytit{Data Pre-Processing}
To extract the person pose representation $p$, we employ either OpenPose~\cite{cao2017realtime} or DensePose~\cite{guler2018densepose}. Specifically, the keypoints of the human body extracted with OpenPose~\cite{cao2017realtime} are used to compute the $18$-channel pose heatmap, where each channel corresponds to one body keypoint represented as an $11 \times 11$ white rectangle. While both the $25$ channels label map and the $2$ channels UV map estimated by DensePose~\cite{guler2018densepose} are concatenated and used with no further processing.

In order to create the masked person representation $m$, we remove the information regarding the target clothes and the interested part of the body from $I$. Hence, the model only sees the face, the hair, and the target person part of the body which do not contain ground-truth information. 
To produce such masked representation, we use both the target label map to extract the clothes area and the pose map to extract the area of the limbs. These areas are then merged to form the mask which is then dilated to avoid the model getting information about the target shape. Finally, all the non-modifiable areas in the image (\eg~face, hands, hairs, etc.) are subtracted from the generated mask. The final mask is then applied to the image $I$. Note that while dilating the mask introduces more complexity in the paired setting generation task, it is essential in the unpaired one, especially when trying to substitute a garment with another whose shape covers a much larger area of the image.  

\begin{table}[t]
\centering
\footnotesize
\caption{Number of train and test pairs for each category of the Dress Code dataset.}
\label{tab:stats}
\resizebox{0.7\linewidth}{!}{
\setlength{\tabcolsep}{.4em}
\begin{tabular}{lc ccc}
\toprule
& & \textbf{Images} & \textbf{Training Pairs} & \textbf{Test Pairs} \\
\midrule
\textbf{Upper-body Clothes} & & 30,726 & 13,563 & 1,800 \\
\textbf{Lower-body Clothes} & & 17,902 & 7,151 & 1,800 \\
\textbf{Dresses} & & 58,956 & 27,678 & 1,800 \\
\midrule
\textbf{All} & & 107,584 & 48,392 & 5,400 \\
\bottomrule
\end{tabular}
}
\vspace{-0.4cm}
\end{table}

\tit{Warping Module} Two feature extraction networks are included in the warping module, with four $2$-strided down-sampling convolutional layers with a kernel size of $4$ plus two $1$-strided ones with a kernel size of $3$. The first extraction network takes as input the try-on clothing item $c$, while the second one works on the concatenation between the person representation $m$ and the pose of the reference person $p$. Following~\cite{wang2018toward}, a correlation map is then computed between the outputs of the two feature extraction networks and then fed to a convolutional network, consisting of two $2$-strided convolutional layers with a kernel size of $4$ and two $1$-strided convolutional layers with a kernel size of $3$. The output is forwarded through a fully connected layer that predicts the parameters of the geometric transformation. In particular, these parameters are the TPS anchor point coordinate offsets having a size of $2\times5\times5 = 50$. Batch normalization is applied to all convolutional layers. For the high-resolution versions of our model, we add an additional $2$-strided down-sampling convolutional layer with a kernel size of $4$ to both feature extraction networks.

\tit{Human Parsing Estimation Module}
It is based on the U-Net architecture with four blocks in both encoder and decoder. Each block is composed of two sequences of a convolutional layer with a kernel size of $3$, instance normalization, and a ReLU activation function. Each encoding block is followed by a $2$-strided max pooling layer with a kernel size of $2$, while each decoding block is preceded by a 2-strided transposed convolutional layer with a kernel size of $2$ to upsample feature maps. Each encoding block is connected to the corresponding decoding block using skip connections. When training with high-resolution images, we add a U-Net block in both encoder and decoder.

\tit{Try-On Module} The encoder has four U-Net blocks, each having two convolutional layers with a kernel size of $3$ and a $2$-strided max pooling layer with a kernel size of $2$. The decoder is symmetric but, instead of max pooling, the feature maps are up-sampled using a 2-strided transposed convolutional layer with a kernel size of $2$. Also in this case, when training with high-resolution images, we add a U-Net block in both encoder and decoder.

\tit{Discriminator} PSAD works at pixel-level, classifying each pixel as one of the $N$ semantic classes of the human parser or as fake. The architecture is composed of $6$ downsampling and $6$ upsampling blocks arranged according to the U-Net architecture. The last layer is a $1\times1$ spatial convolution that brings the feature dimensionality to $N+1$. 

When training the Patch-based baseline, we instead employ PatchGAN~\cite{isola2017image} as our discriminator, which does not operate at pixel-level but instead classifies square image patches as real or fake, averaging all predictions to get the final result. It consists of three 2-strided down-sampling convolutional layers and one 1-strided down-sampling convolutional layer, all having a kernel size of $4$. We use a convolutional layer to generate a scalar output in the last layer. Except for the first, we utilize batch normalization and apply Leaky ReLU with a $0.2$ slope after each layer.

\begin{table*}[t]
\centering
\footnotesize
\caption{High-resolution results on the Dress Code test set}
\label{tab:hd_try-on_res}
\setlength{\tabcolsep}{.3em}
\resizebox{\linewidth}{!}{
\begin{tabular}{lc ccc c ccc c ccc c cccc}
\toprule
& & \multicolumn{3}{c}{\textbf{Upper-body}} & & \multicolumn{3}{c}{\textbf{Lower-body}} & & \multicolumn{3}{c}{\textbf{Dresses}} & & \multicolumn{4}{c}{\textbf{All}} \\
\cmidrule{3-5} \cmidrule{7-9} \cmidrule{11-13} \cmidrule{15-18}
\textbf{Model ($512 \times 384$)} & & \textbf{SSIM} $\uparrow$ & \textbf{FID} $\downarrow$ & \textbf{KID} $\downarrow$ & & \textbf{SSIM} $\uparrow$ & \textbf{FID} $\downarrow$ & \textbf{KID} $\downarrow$ & & \textbf{SSIM} $\uparrow$ & \textbf{FID} $\downarrow$ & \textbf{KID} $\downarrow$ & & \textbf{SSIM} $\uparrow$ & \textbf{FID} $\downarrow$ & \textbf{KID} $\downarrow$ & \textbf{IS} $\uparrow$ \\
\midrule
CP-VTON~\cite{wang2018toward} & & 0.850 & 48.24 & 3.365 & & 0.826 & 57.38 & 4.00 & & 0.845 & 24.04 & 0.891 & & 0.831 & 29.24 & 1.671 & 3.096 \\
CP-VTON$^\dagger$~\cite{wang2018toward} & & 0.916 & 14.37 & 0.442 & & 0.910 & \textbf{12.54} & \textbf{0.432} & & 0.861 & 21.82 & 0.720 & & 0.896 & 10.08 & 0.425 & 3.277 \\
\textbf{Ours (Patch)}  & & \textbf{0.943} & 13.48 & 0.346 & & \textbf{0.936} & 19.86 & 0.717 & & \textbf{0.893} & 20.35 & 0.623 & & \textbf{0.923} & 9.44 & \textbf{0.246} & 3.310 \\
\textbf{Ours (PSAD)} & & 0.936 & \textbf{11.65} & \textbf{0.180} & & 0.931 & 17.83 & 0.643 & & 0.884 & \textbf{15.99} & \textbf{0.324} & & 0.916 & \textbf{7.27} & 0.394 & \textbf{3.320} \\
\midrule
& & \multicolumn{3}{c}{\textbf{Upper-body}} & & \multicolumn{3}{c}{\textbf{Lower-body}} & & \multicolumn{3}{c}{\textbf{Dresses}} & & \multicolumn{4}{c}{\textbf{All}} \\
\cmidrule{3-5} \cmidrule{7-9} \cmidrule{11-13} \cmidrule{15-18}
\textbf{Model ($1024 \times 768$)} & & \textbf{SSIM} $\uparrow$ & \textbf{FID} $\downarrow$ & \textbf{KID} $\downarrow$ & & \textbf{SSIM} $\uparrow$ & \textbf{FID} $\downarrow$ & \textbf{KID} $\downarrow$ & & \textbf{SSIM} $\uparrow$ & \textbf{FID} $\downarrow$ & \textbf{KID} $\downarrow$ & & \textbf{SSIM} $\uparrow$ & \textbf{FID} $\downarrow$ & \textbf{KID} $\downarrow$ & \textbf{IS} $\uparrow$ \\
\midrule
CP-VTON~\cite{wang2018toward} & & 0.862 & 60.40 & 4.730 & & 0.840 & 60.35 & 4.236 & & 0.858 & 24.44 & 0.873 & & 0.853 & 36.68 & 2.379 & 3.155 \\
CP-VTON$^\dagger$~\cite{wang2018toward} & & 0.931 & 14.63 & 0.387 & & 0.930 & \textbf{16.46} & \textbf{0.393} & & 0.877 & 23.80 & 0.832 & & 0.912 & 9.96 & 0.338 & 3.300 \\
\textbf{Ours (Patch)} & & \textbf{0.944} & 13.38 & 0.273 & & 0.933 & 19.97 & 0.654 & & \textbf{0.890} & 24.14 & 0.807 & & \textbf{0.922} & 9.99 & 0.370 & 3.34 \\
\textbf{Ours (PSAD)} & & 0.941 & \textbf{12.10} & \textbf{0.171} & & \textbf{0.935} & 19.02 & 0.641 & & 0.882 & \textbf{17.93} & \textbf{0.425} & & 0.919 & \textbf{7.70} & \textbf{0.236} & \textbf{3.357} \\
\bottomrule
\end{tabular}
}
\vspace{-0.3cm}
\end{table*}

\tinytit{Training} The experiments with low-resolution images are performed using a batch size of $32$, while we use a batch size of $16$ when training with high-resolution images for both $512\times384$ and $1024\times768$ resolutions. All experiments with $256\times192$ and $512\times384$ images are performed on 4 NVIDIA V100 GPUs, taking $10$ hours to train the human parsing estimation module, one day for the warping module training stage, and around two days to train the try-on module. When instead training with full-resolution images, we split the batch size on 16 GPUs.

\begin{figure*}[t]
\centering
\footnotesize
\setlength{\tabcolsep}{.3em}
\resizebox{\linewidth}{!}{
\begin{tabular}{ccc}
\includegraphics[height=0.18\linewidth]{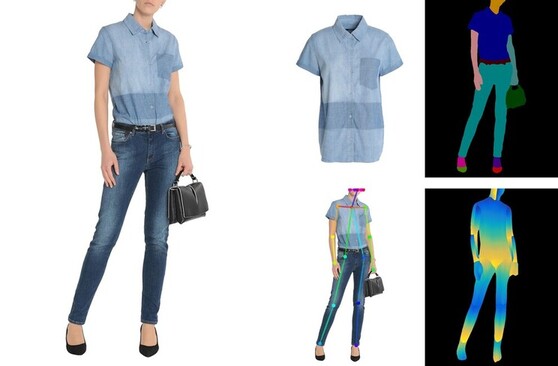}
\includegraphics[height=0.18\linewidth]{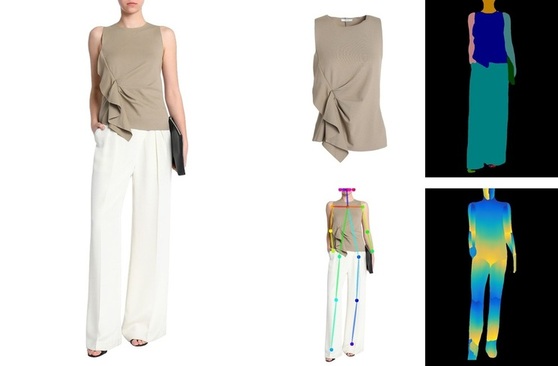}
\includegraphics[height=0.18\linewidth]{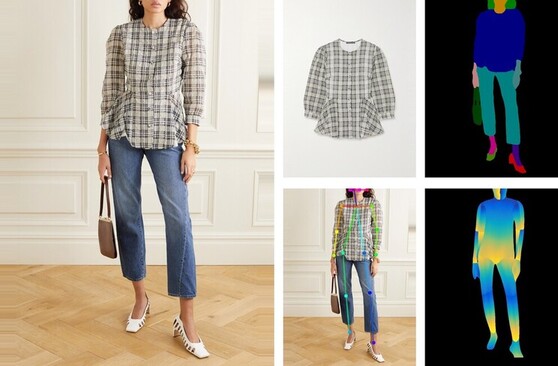} \\
\addlinespace[0.05cm]
\includegraphics[height=0.18\linewidth]{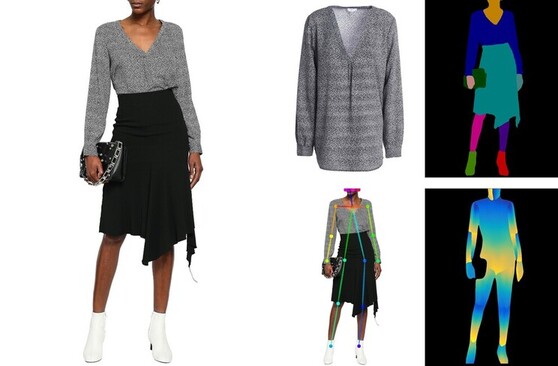}
\includegraphics[height=0.18\linewidth]{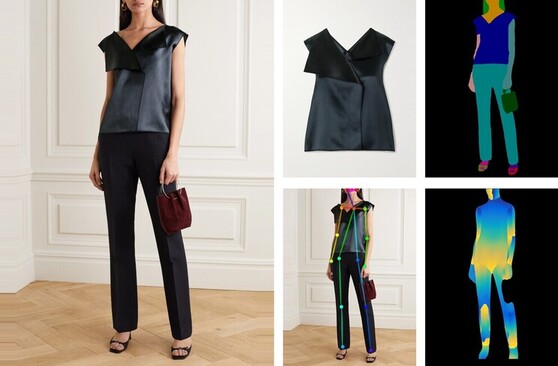}
\includegraphics[height=0.18\linewidth]{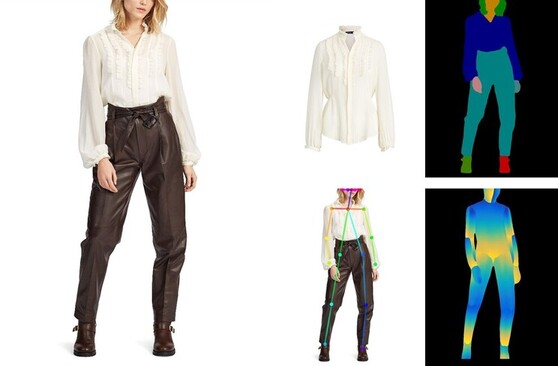} \\
\addlinespace[0.05cm]
\includegraphics[height=0.18\linewidth]{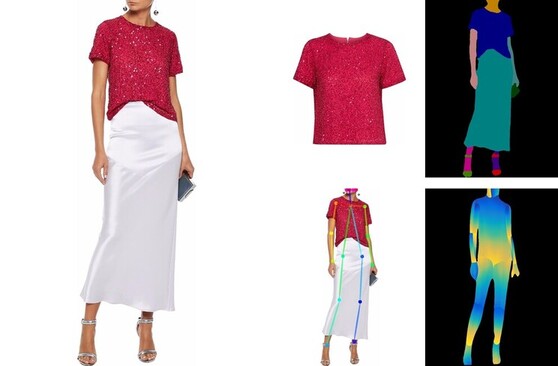}
\includegraphics[height=0.18\linewidth]{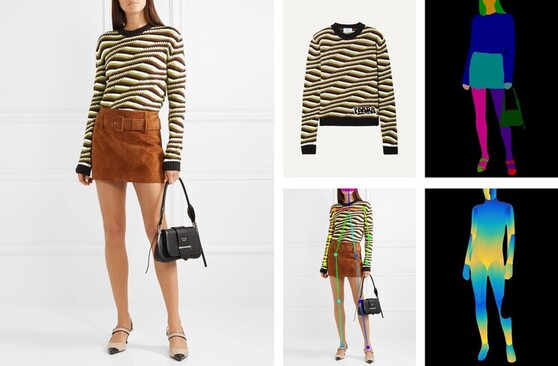}
\includegraphics[height=0.18\linewidth]{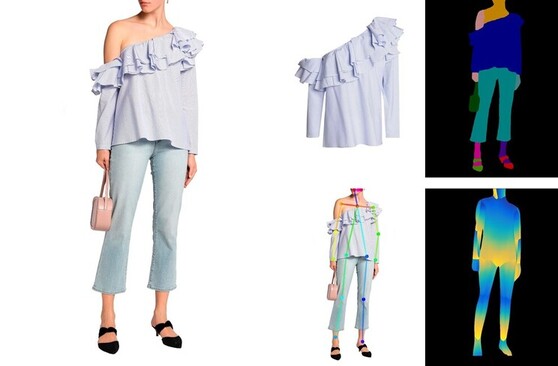} \\
\addlinespace[0.05cm]
\includegraphics[height=0.18\linewidth]{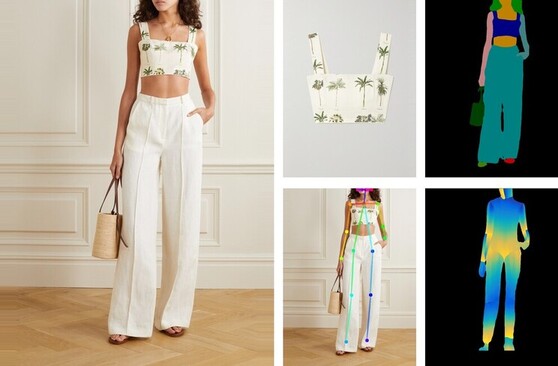}
\includegraphics[height=0.18\linewidth]{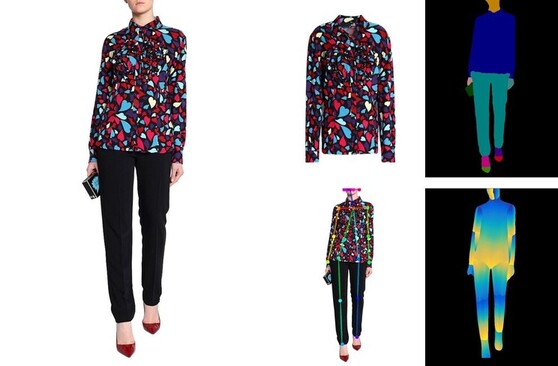}
\includegraphics[height=0.18\linewidth]{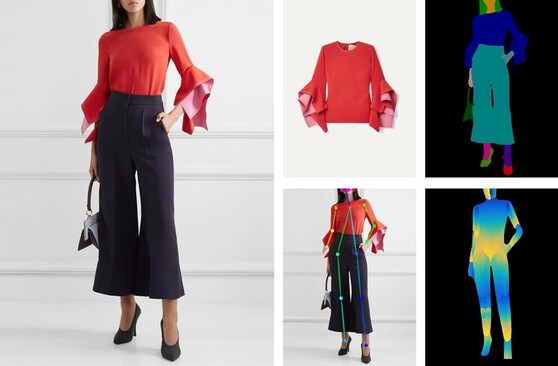} \\
\addlinespace[0.05cm]
\includegraphics[height=0.18\linewidth]{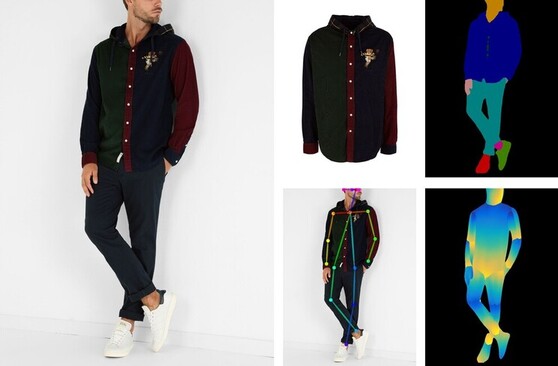}
\includegraphics[height=0.18\linewidth]{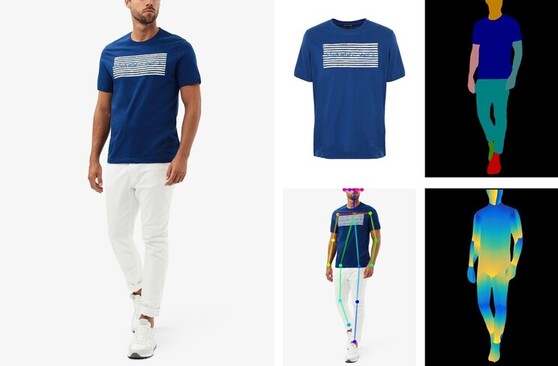}
\includegraphics[height=0.18\linewidth]{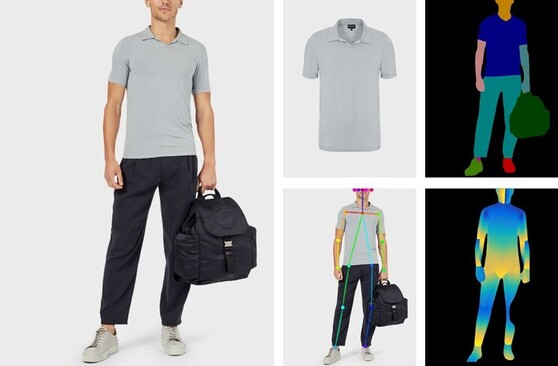} \\
\addlinespace[0.05cm]
\includegraphics[height=0.18\linewidth]{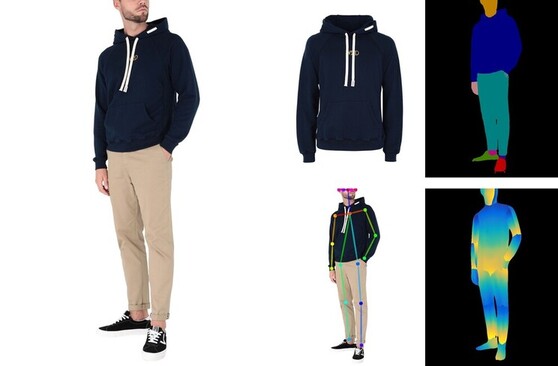}
\includegraphics[height=0.18\linewidth]{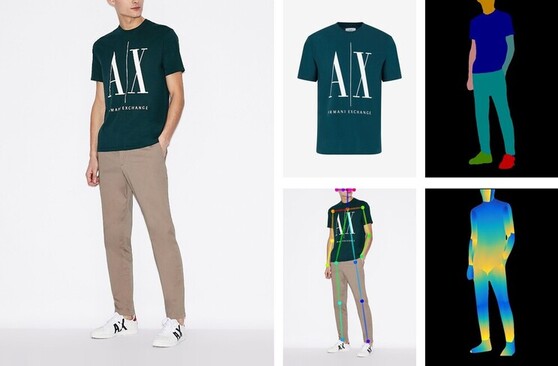}
\includegraphics[height=0.18\linewidth]{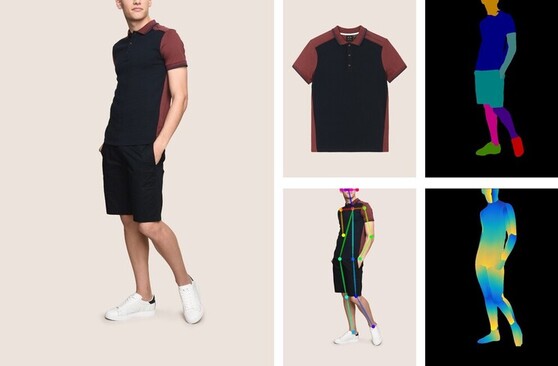} \\
\addlinespace[0.05cm]
\includegraphics[height=0.18\linewidth]{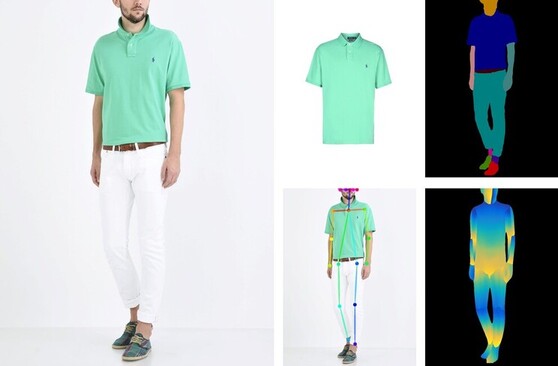}
\includegraphics[height=0.18\linewidth]{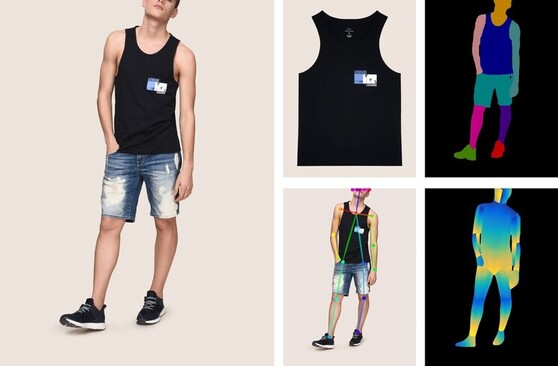}
\includegraphics[height=0.18\linewidth]{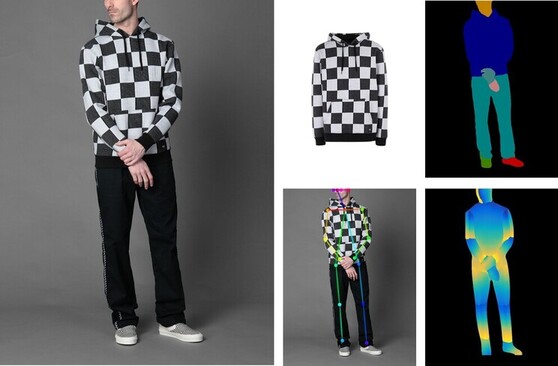} \\
\end{tabular}
}
\caption{Sample images of upper-body clothes and reference models from Dress Code.}
\label{fig:dataset_upperbody}
\end{figure*}

\begin{figure*}[t]
\centering
\footnotesize
\setlength{\tabcolsep}{.3em}
\resizebox{\linewidth}{!}{
\begin{tabular}{ccc}
\includegraphics[height=0.18\linewidth]{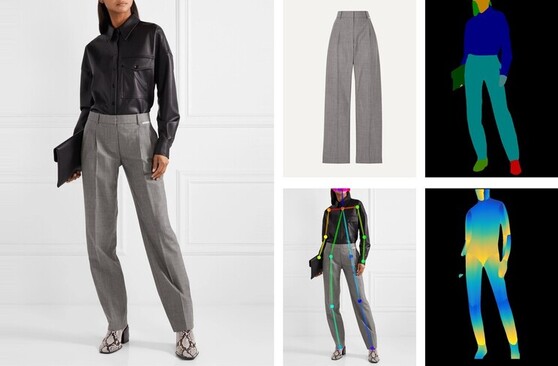} &
\includegraphics[height=0.18\linewidth]{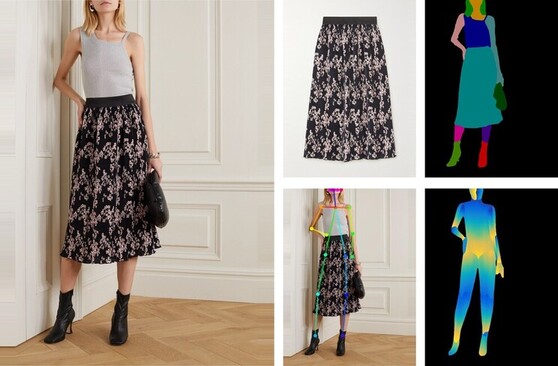} &
\includegraphics[height=0.18\linewidth]{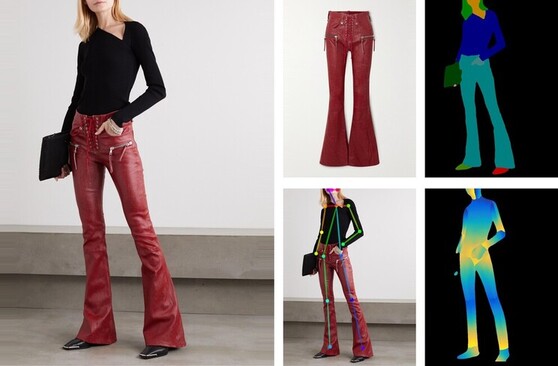} \\
\addlinespace[0.05cm]
\includegraphics[height=0.18\linewidth]{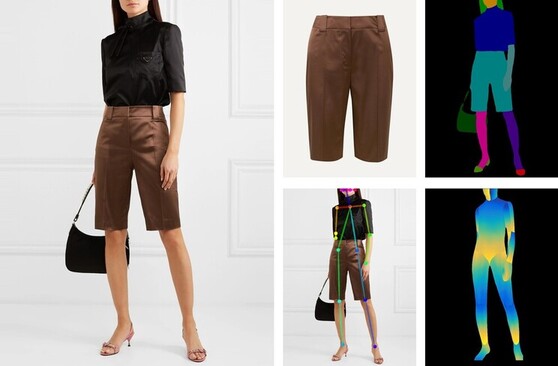} &
\includegraphics[height=0.18\linewidth]{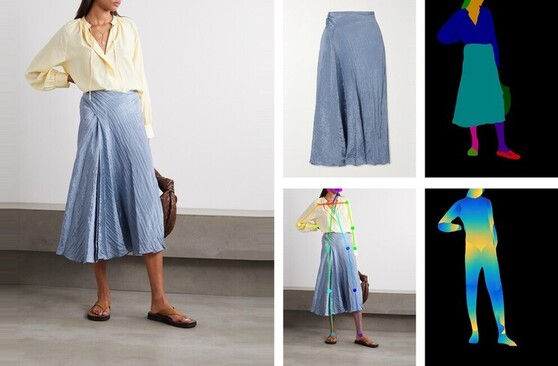} &
\includegraphics[height=0.18\linewidth]{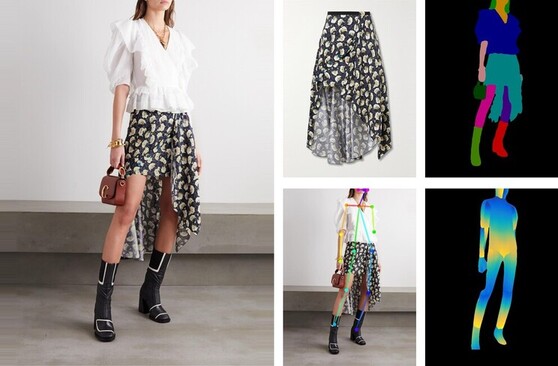} \\
\addlinespace[0.05cm]
\includegraphics[height=0.18\linewidth]{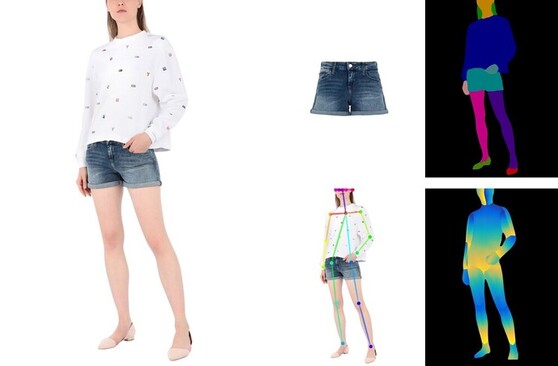} &
\includegraphics[height=0.18\linewidth]{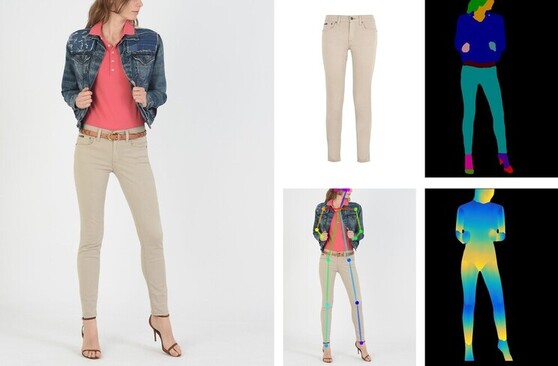} &
\includegraphics[height=0.18\linewidth]{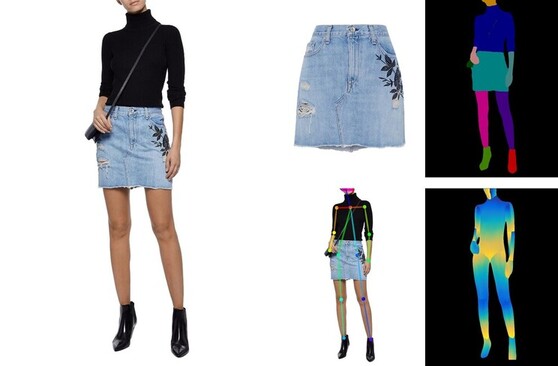} \\
\addlinespace[0.05cm]
\includegraphics[height=0.18\linewidth]{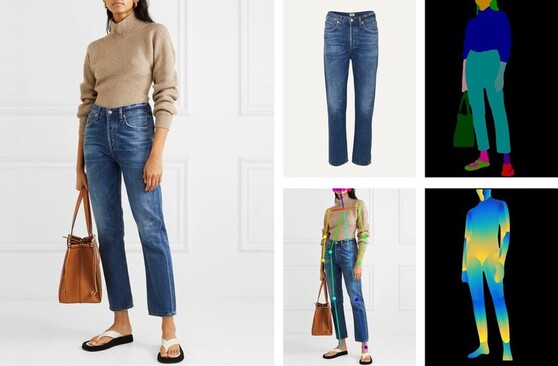} &
\includegraphics[height=0.18\linewidth]{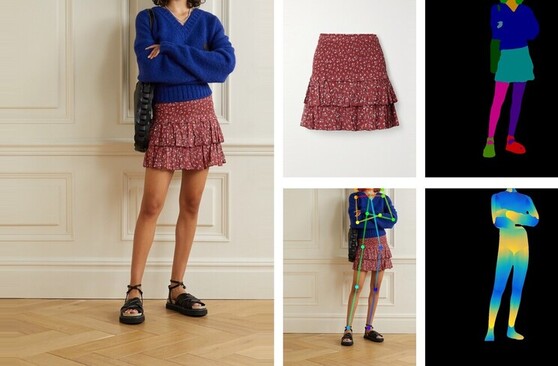} &
\includegraphics[height=0.18\linewidth]{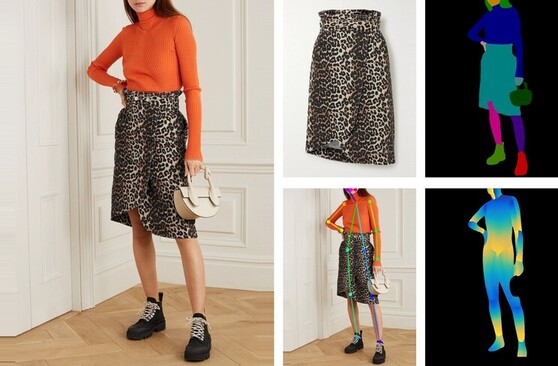} \\
\addlinespace[0.05cm]
\includegraphics[height=0.18\linewidth]{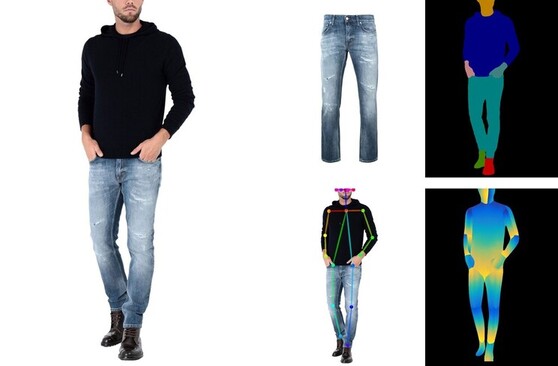} &
\includegraphics[height=0.18\linewidth]{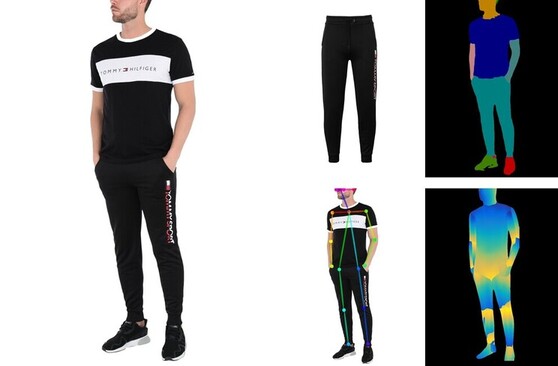} &
\includegraphics[height=0.18\linewidth]{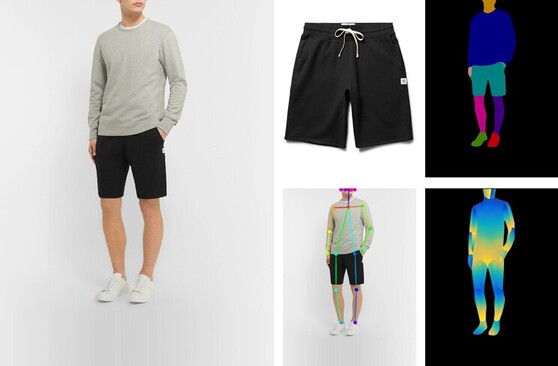} \\
\addlinespace[0.05cm]
\includegraphics[height=0.18\linewidth]{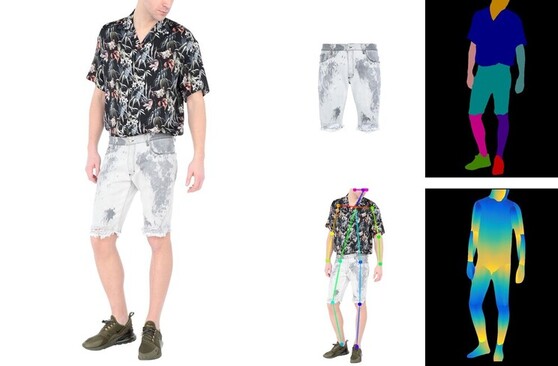} &
\includegraphics[height=0.18\linewidth]{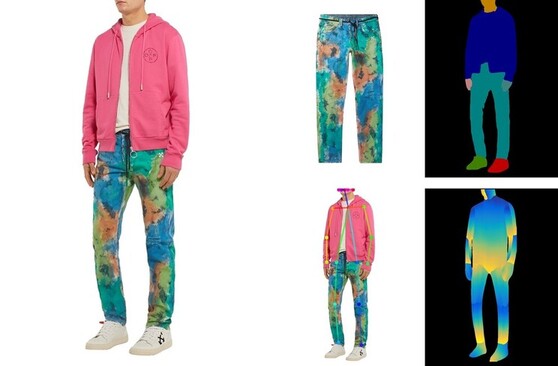} &
\includegraphics[height=0.18\linewidth]{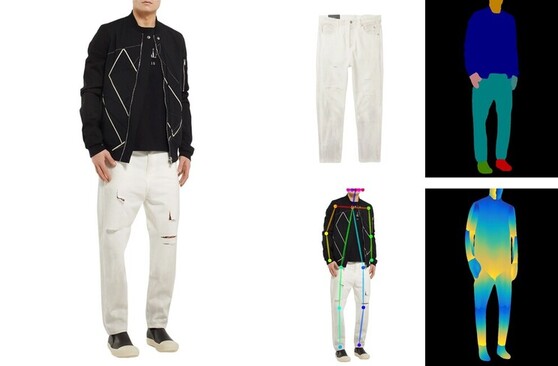} \\
\addlinespace[0.05cm]
\includegraphics[height=0.18\linewidth]{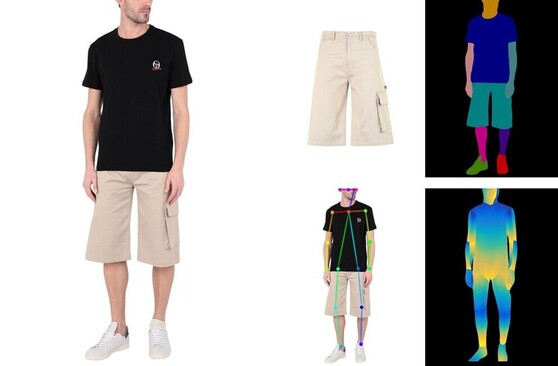} &
\includegraphics[height=0.18\linewidth]{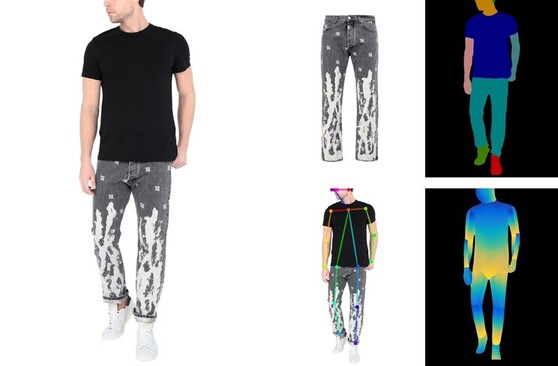} &
\includegraphics[height=0.18\linewidth]{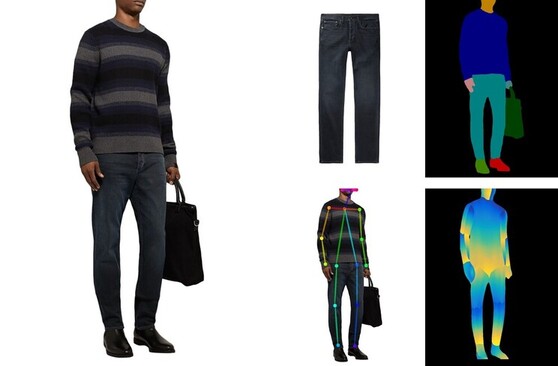} \\
\end{tabular}
}
\caption{Sample images of lower-body clothes and reference models from Dress Code.}
\label{fig:dataset_lowerbody}
\end{figure*}

\begin{figure*}[t]
\centering
\footnotesize
\setlength{\tabcolsep}{.3em}
\resizebox{\linewidth}{!}{
\begin{tabular}{ccc}
\includegraphics[height=0.18\linewidth]{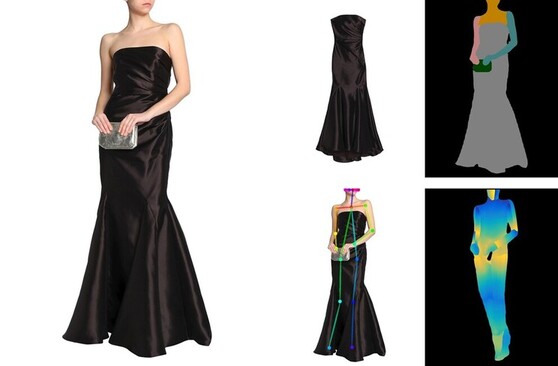} &
\includegraphics[height=0.18\linewidth]{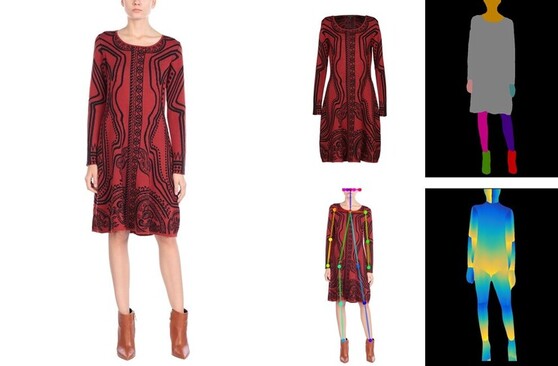} &
\includegraphics[height=0.18\linewidth]{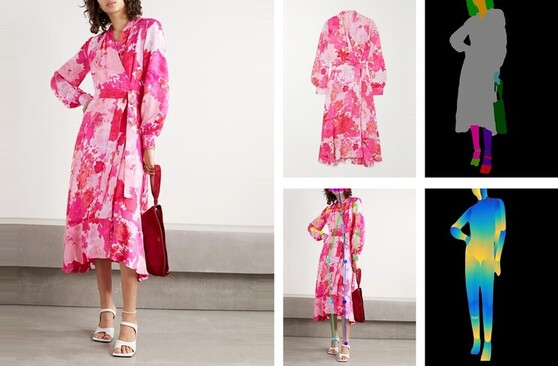} \\
\addlinespace[0.05cm]
\includegraphics[height=0.18\linewidth]{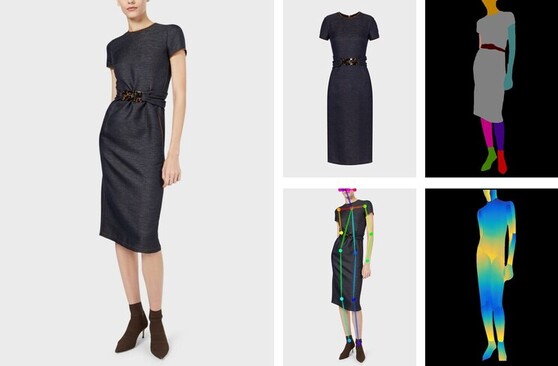} &
\includegraphics[height=0.18\linewidth]{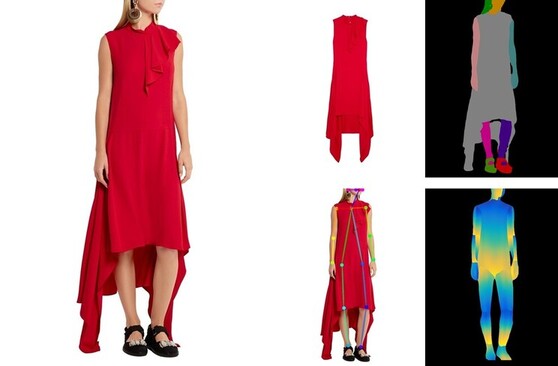} &
\includegraphics[height=0.18\linewidth]{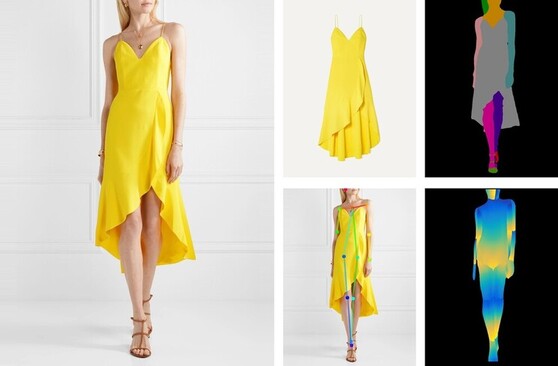} \\
\addlinespace[0.05cm]
\includegraphics[height=0.18\linewidth]{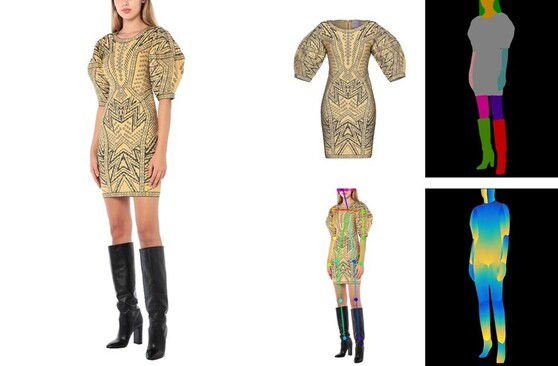} &
\includegraphics[height=0.18\linewidth]{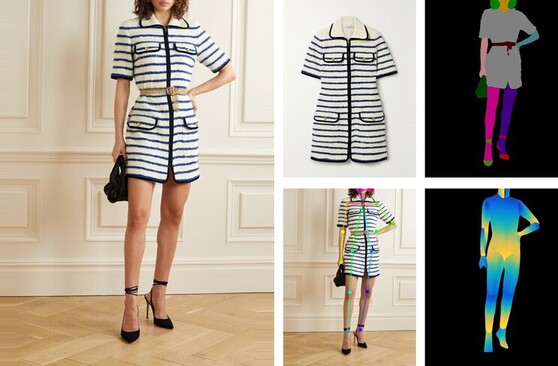} &
\includegraphics[height=0.18\linewidth]{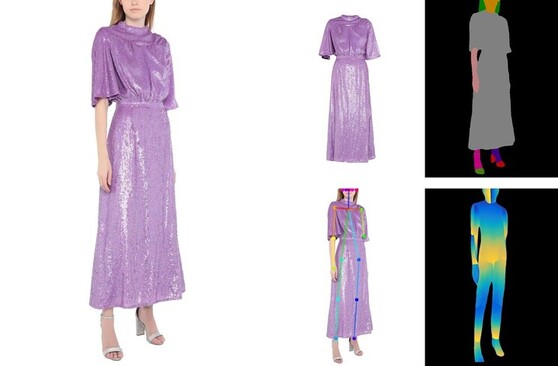} \\
\addlinespace[0.05cm]
\includegraphics[height=0.18\linewidth]{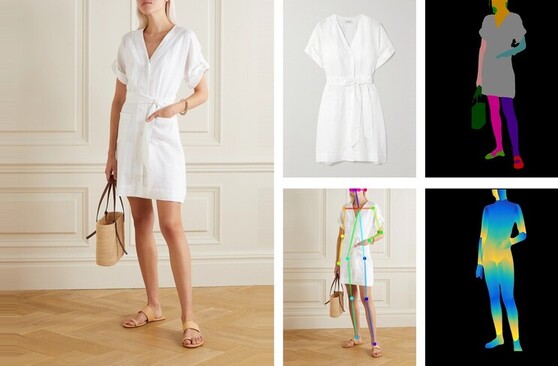} &
\includegraphics[height=0.18\linewidth]{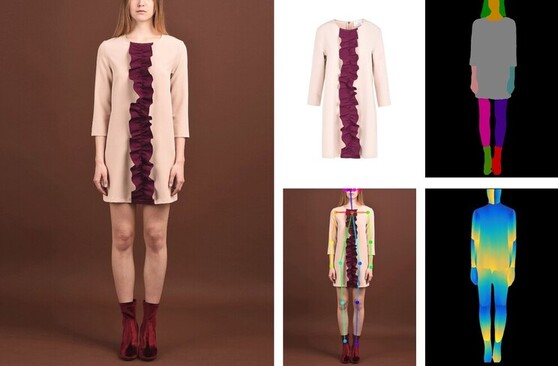} &
\includegraphics[height=0.18\linewidth]{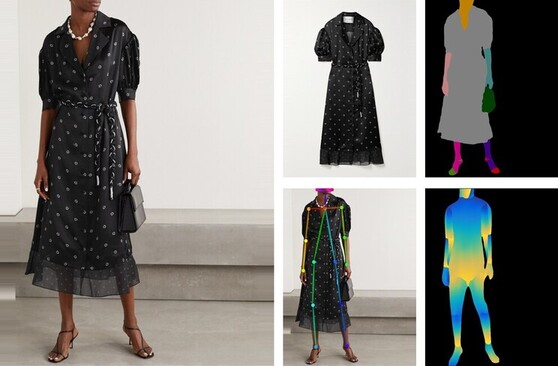} \\
\addlinespace[0.05cm]
\includegraphics[height=0.18\linewidth]{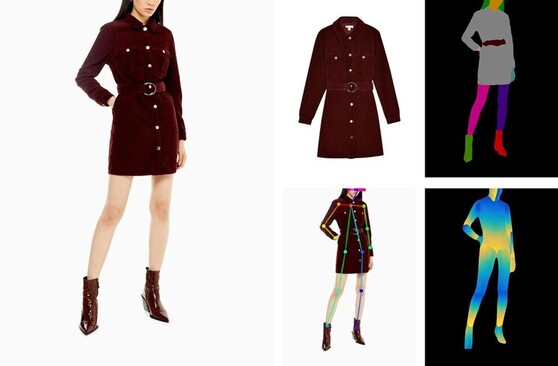} &
\includegraphics[height=0.18\linewidth]{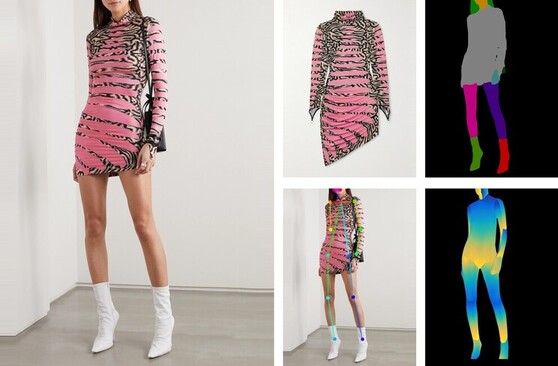} &
\includegraphics[height=0.18\linewidth]{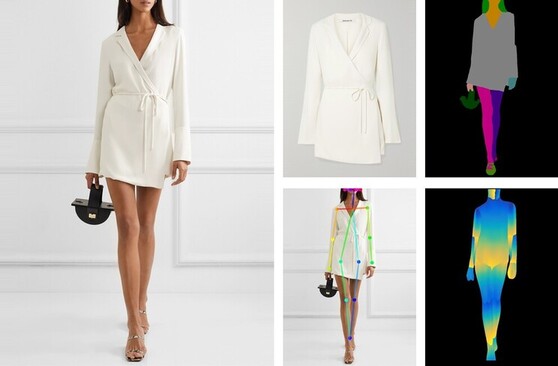} \\
\addlinespace[0.05cm]
\includegraphics[height=0.18\linewidth]{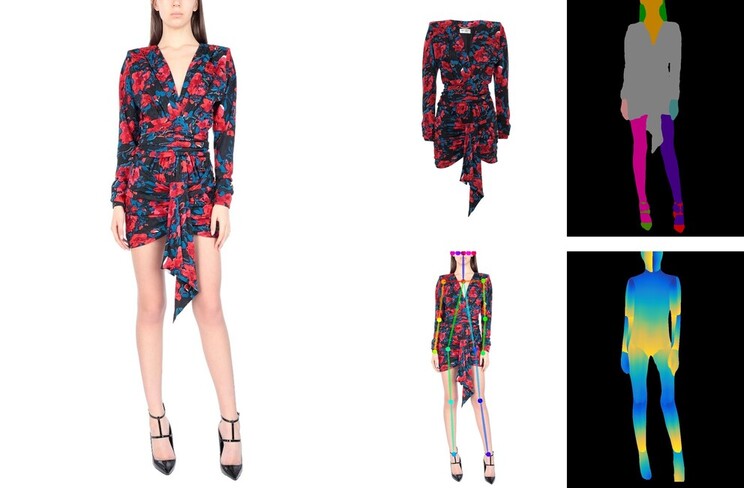} &
\includegraphics[height=0.18\linewidth]{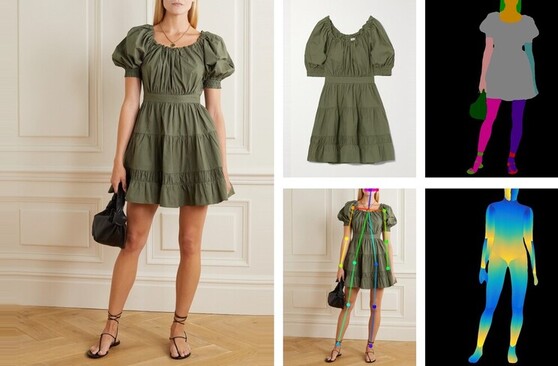} &
\includegraphics[height=0.18\linewidth]{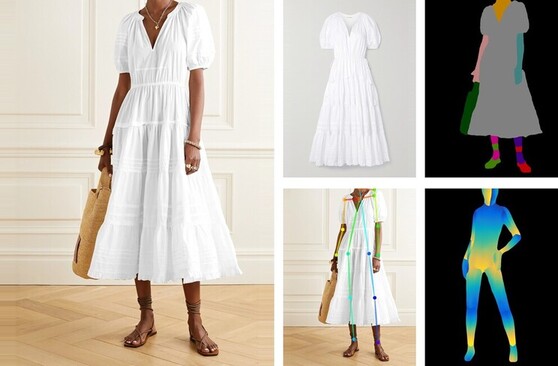} \\
\addlinespace[0.05cm]
\includegraphics[height=0.18\linewidth]{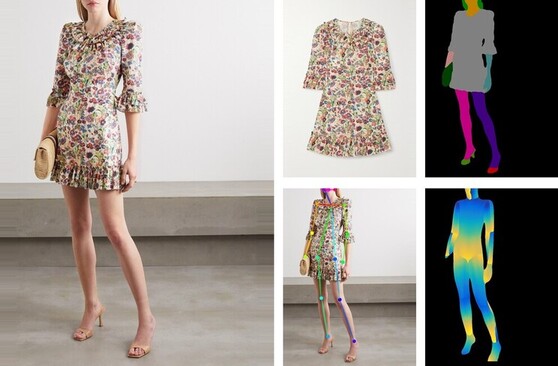} &
\includegraphics[height=0.18\linewidth]{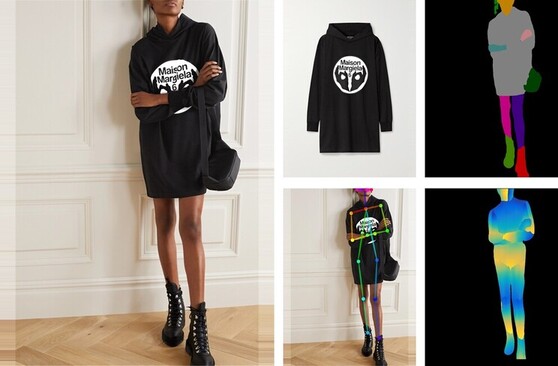} &
\includegraphics[height=0.18\linewidth]{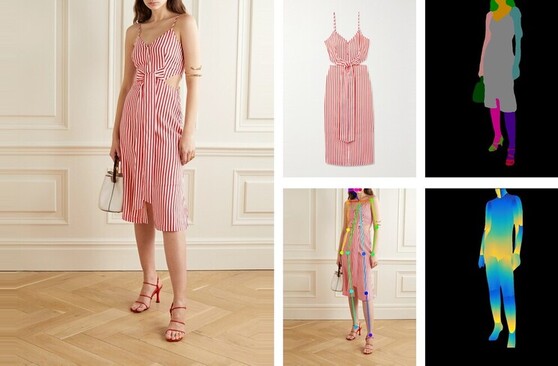} \\
\end{tabular}
}
\caption{Sample images of dresses and reference models from Dress Code.}
\label{fig:dataset_dresses}
\end{figure*}

\subsection*{Additional Results}

\begin{figure*}[t]
\centering
\footnotesize
\setlength{\tabcolsep}{.2em}
\resizebox{\linewidth}{!}{
\begin{tabular}{cc c cc c cc c cc} 
& \textbf{Ours} & & & \textbf{Ours} & & & \textbf{Ours} & & & \textbf{Ours} \\ 
& \textbf{(PSAD)} & & & \textbf{(PSAD)} & & & \textbf{(PSAD)} & & & \textbf{(PSAD)} \\ 
\addlinespace[0.08cm]
\includegraphics[width=0.16\linewidth]{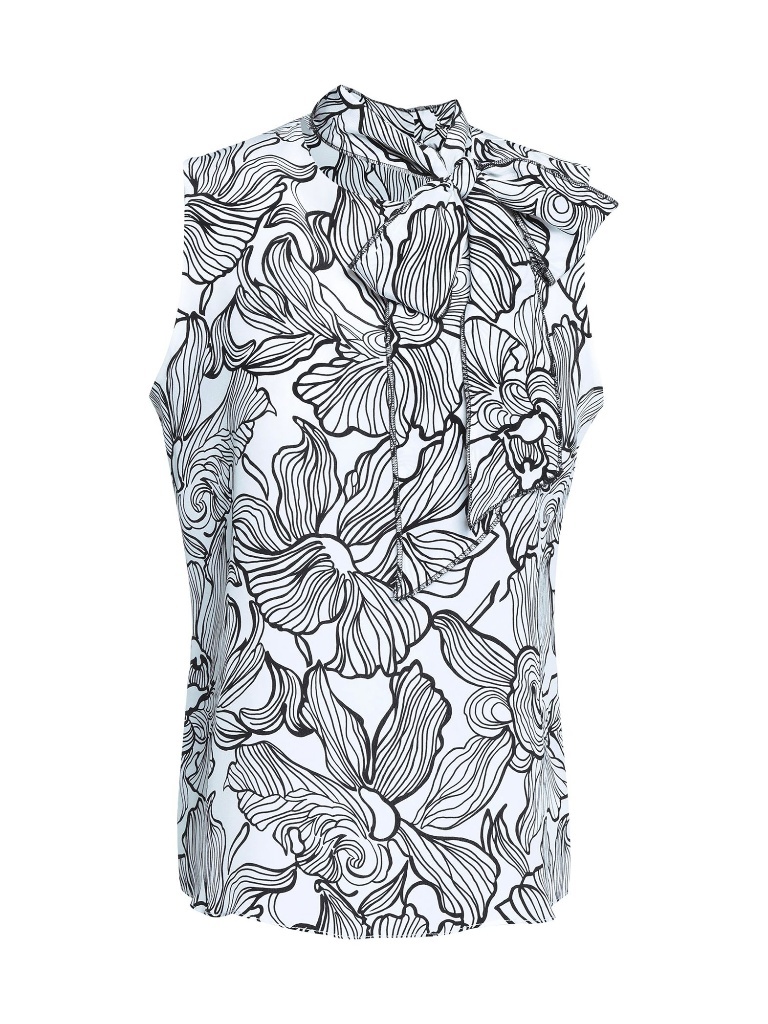} &
\includegraphics[width=0.16\linewidth]{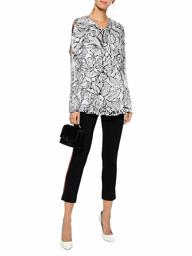} & &
\includegraphics[width=0.16\linewidth]{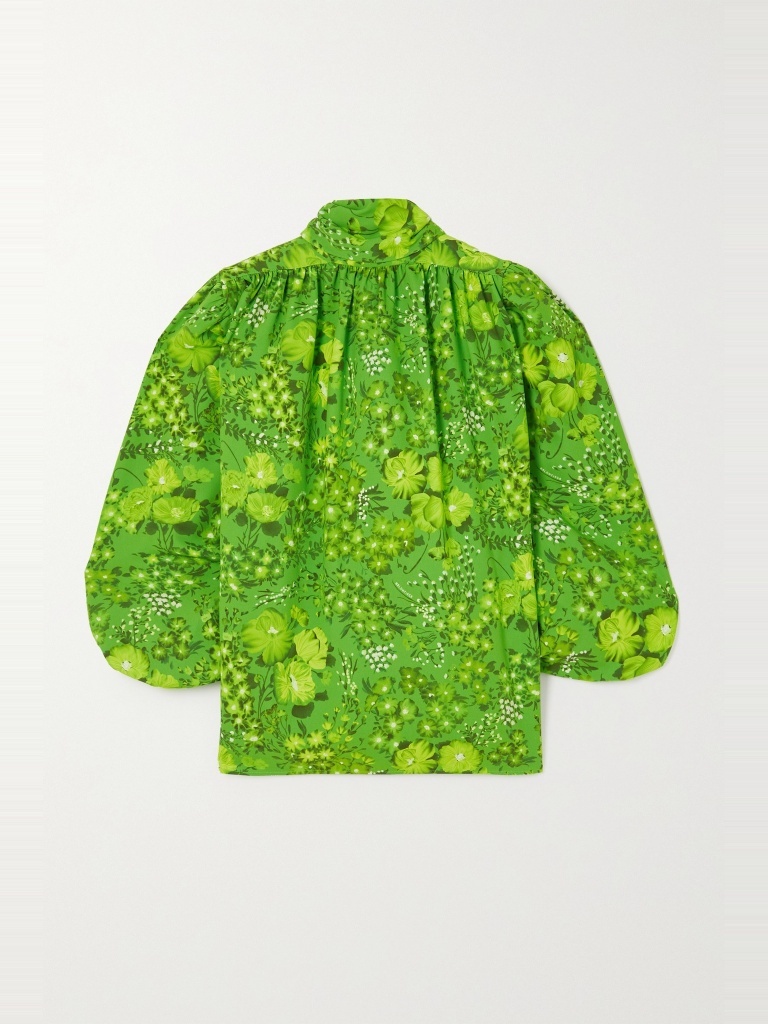} &
\includegraphics[width=0.16\linewidth]{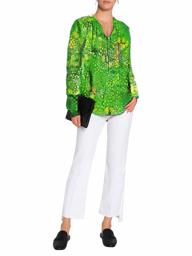} & &
\includegraphics[width=0.16\linewidth]{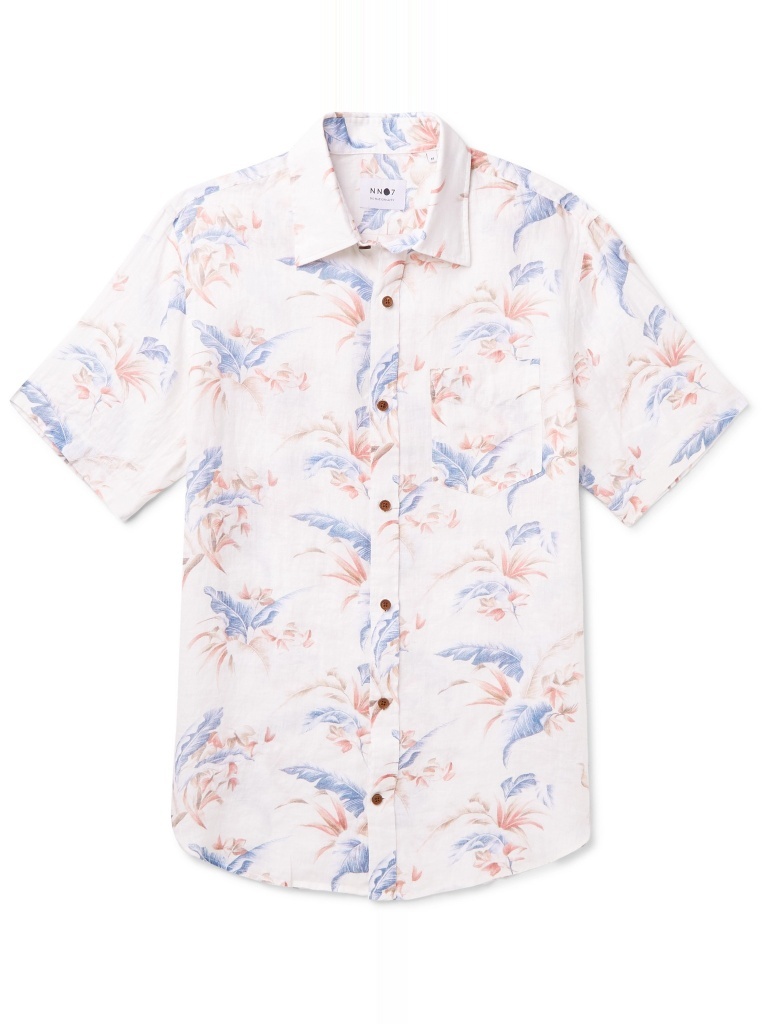} &
\includegraphics[width=0.16\linewidth]{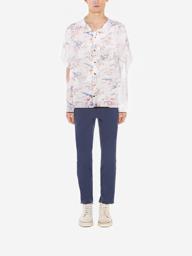} & &
\includegraphics[width=0.16\linewidth]{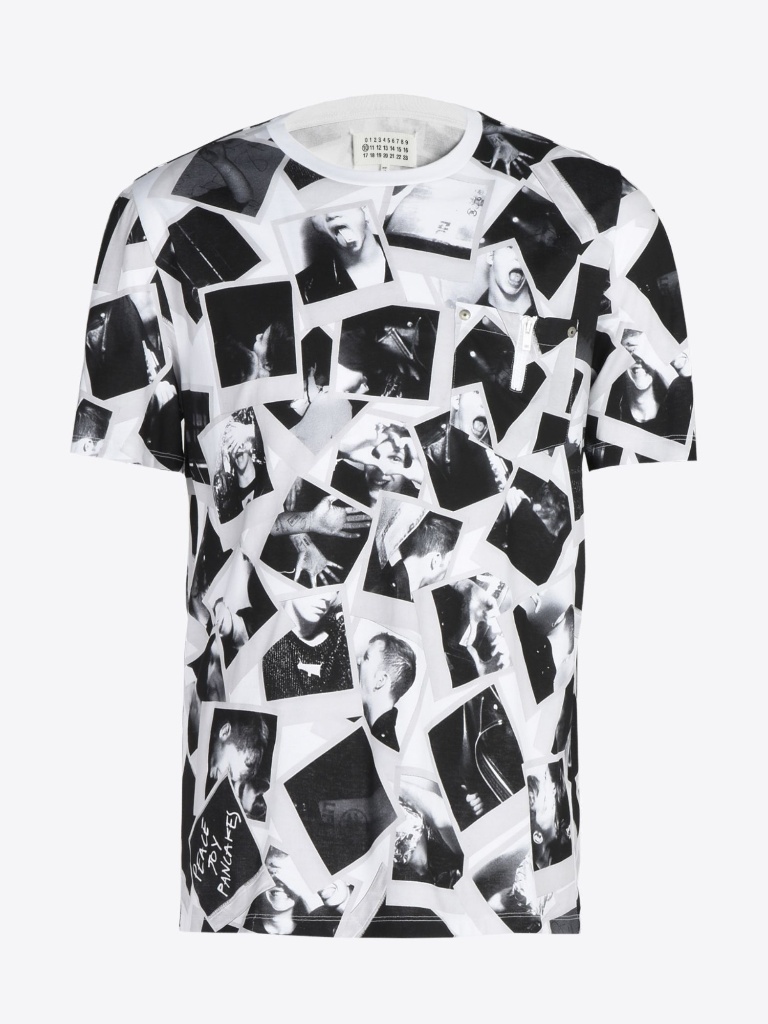} &
\includegraphics[width=0.16\linewidth]{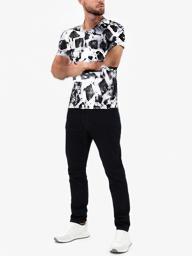} \\
\includegraphics[width=0.16\linewidth]{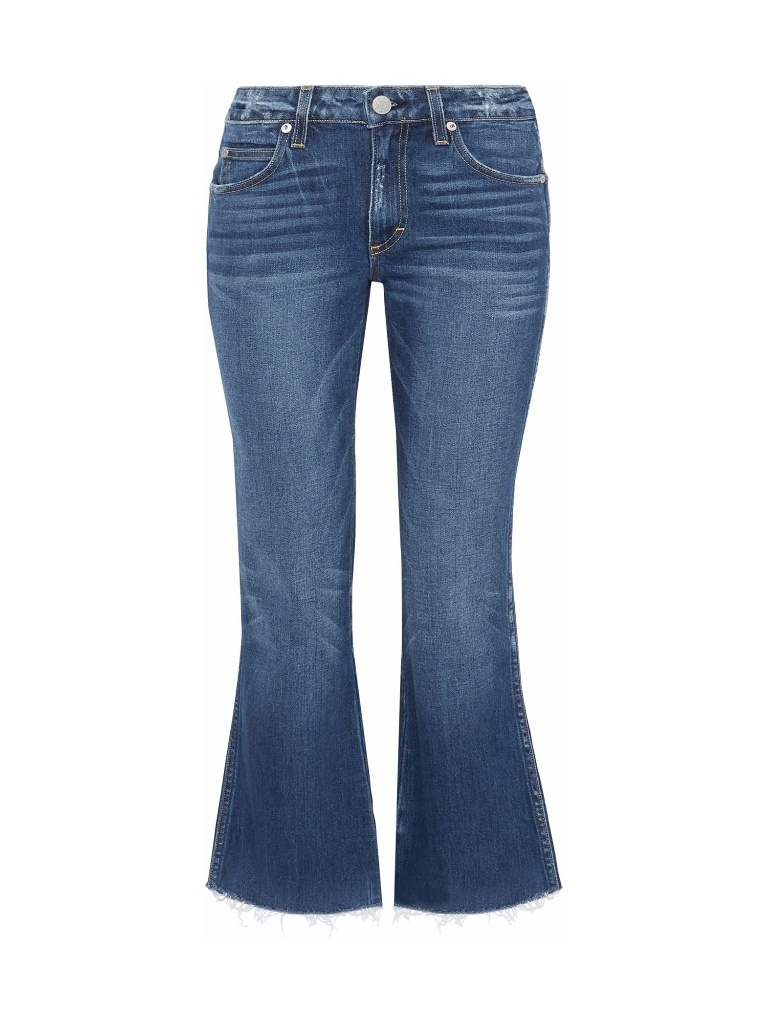} &
\includegraphics[width=0.16\linewidth]{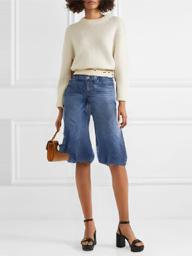} & &
\includegraphics[width=0.16\linewidth]{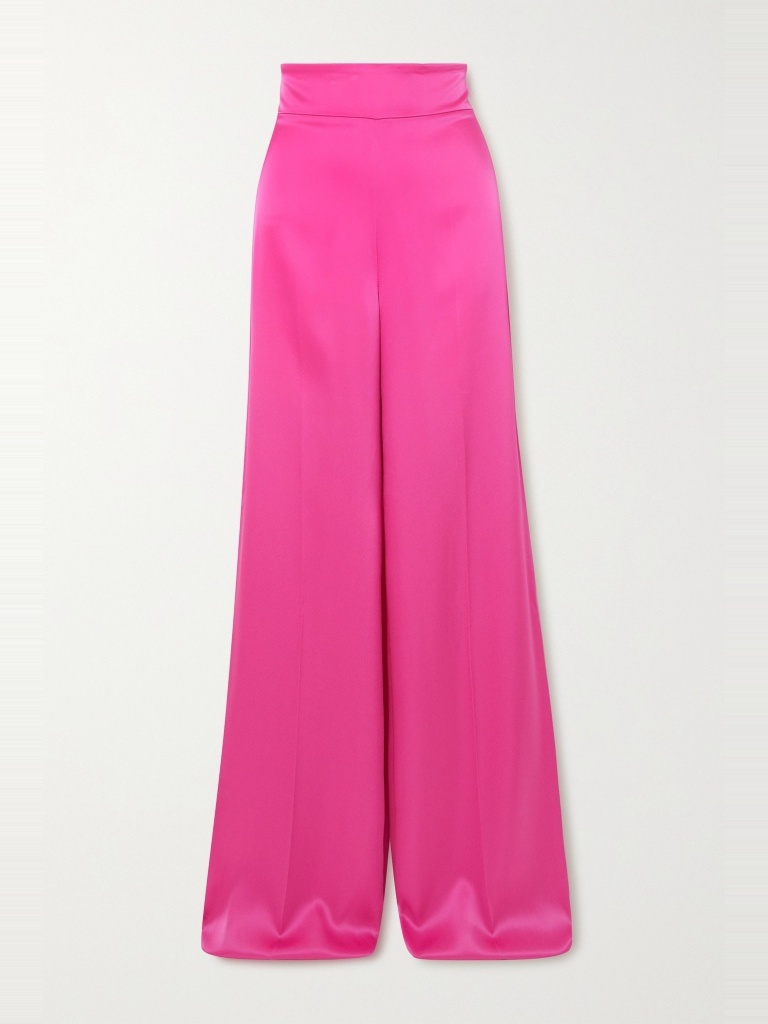} &
\includegraphics[width=0.16\linewidth]{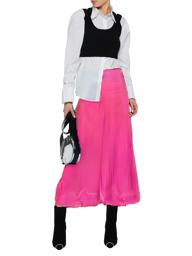} & &
\includegraphics[width=0.16\linewidth]{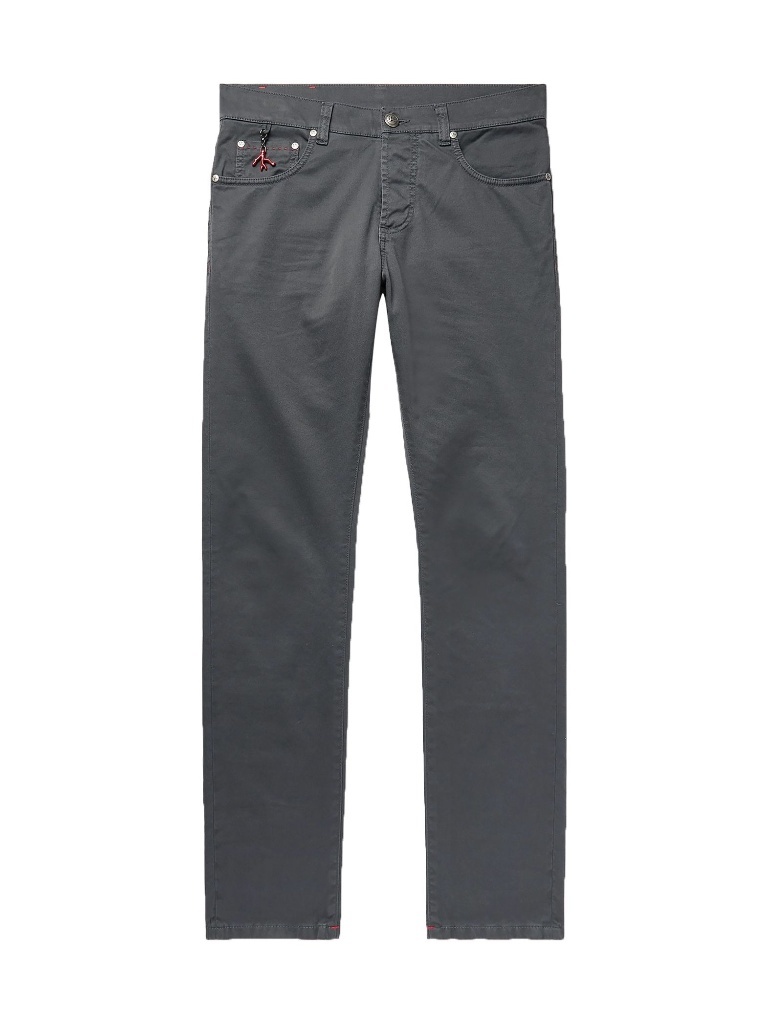} &
\includegraphics[width=0.16\linewidth]{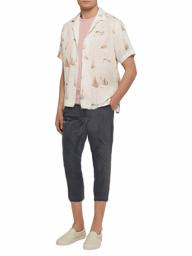} & &
\includegraphics[width=0.16\linewidth]{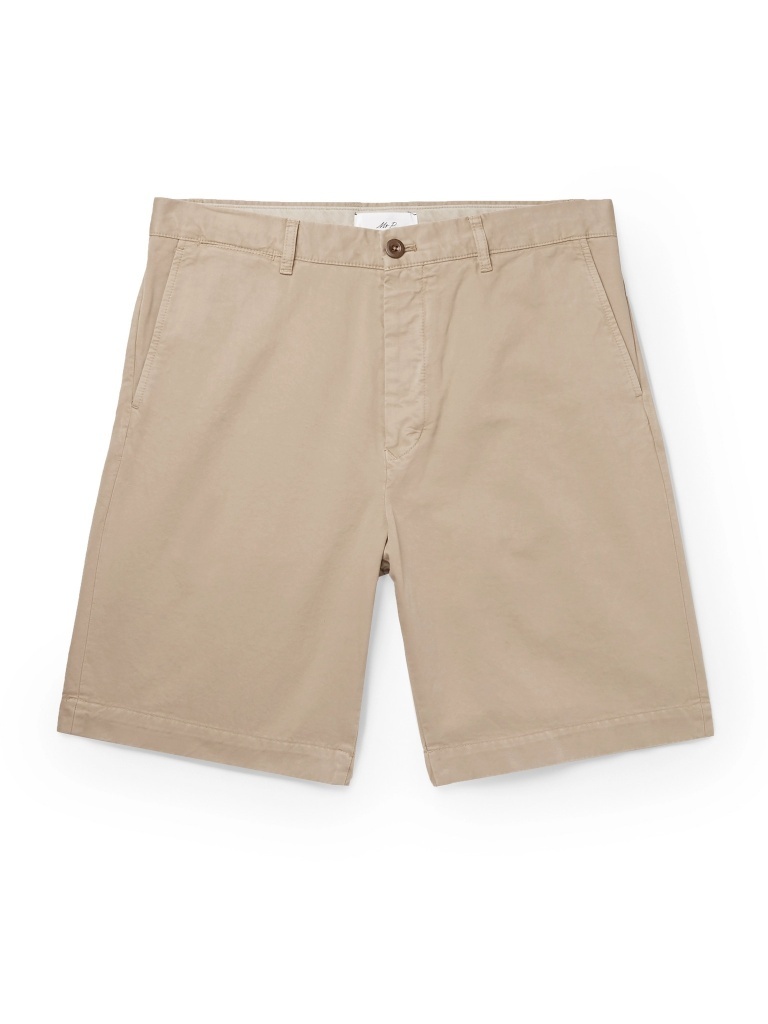} &
\includegraphics[width=0.16\linewidth]{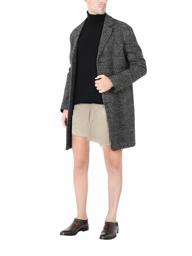} \\
\includegraphics[width=0.16\linewidth]{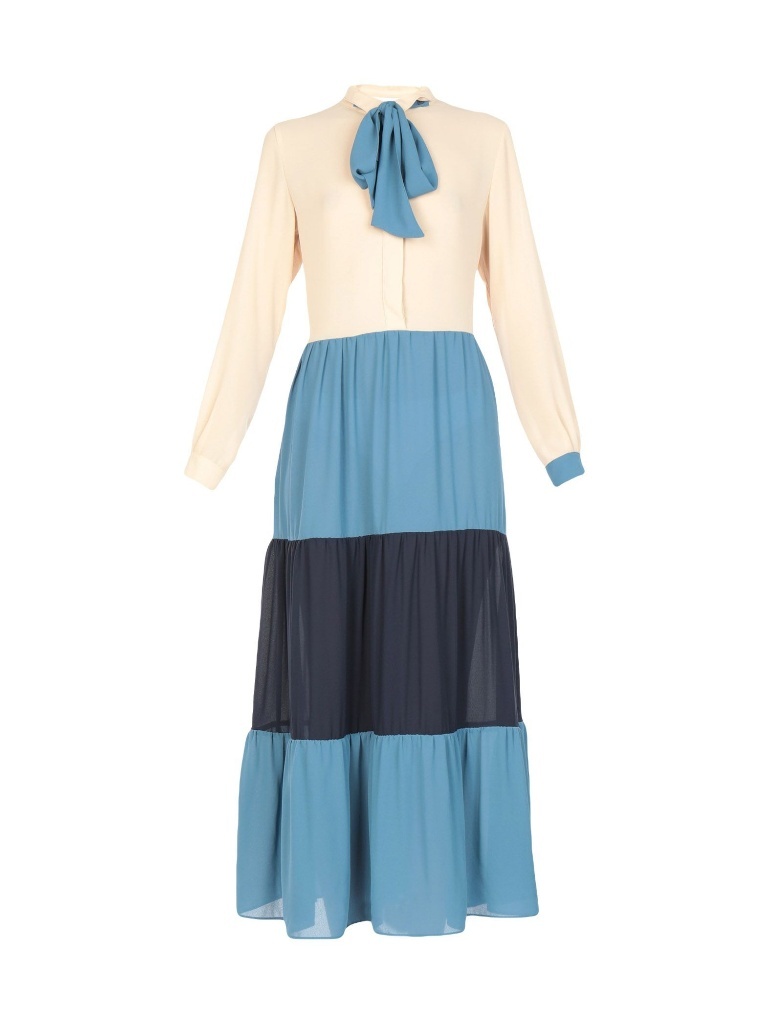} &
\includegraphics[width=0.16\linewidth]{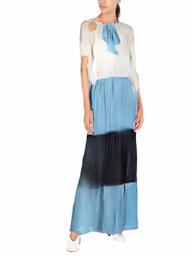} & &
\includegraphics[width=0.16\linewidth]{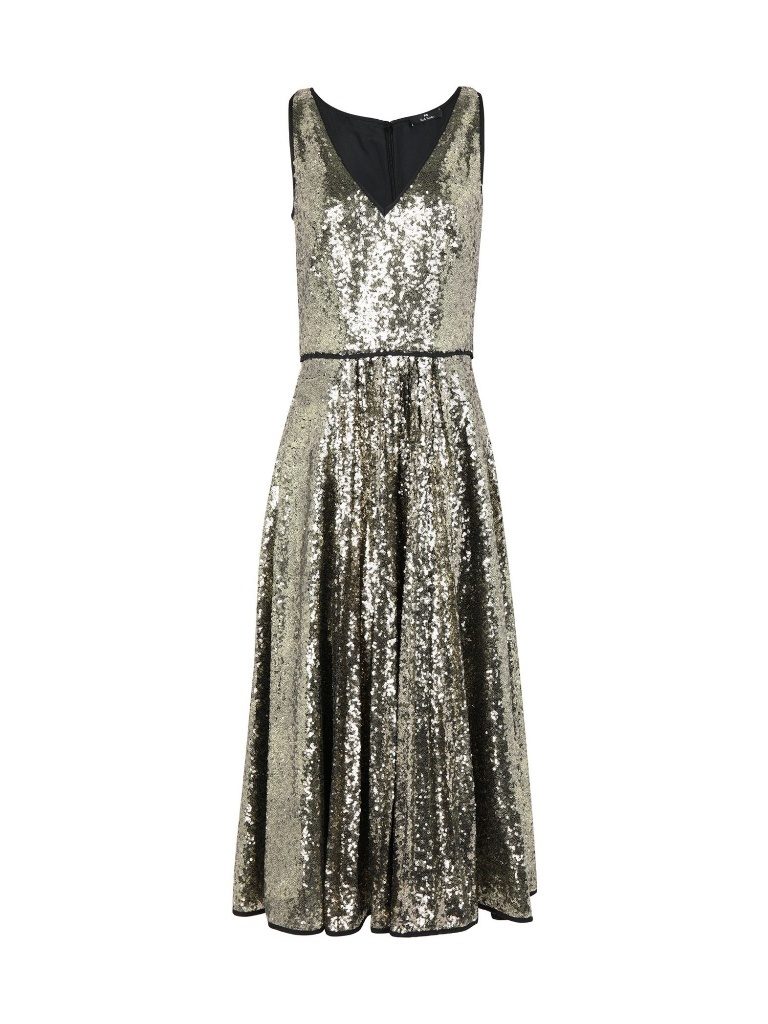} &
\includegraphics[width=0.16\linewidth]{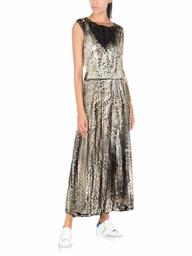} & &
\includegraphics[width=0.16\linewidth]{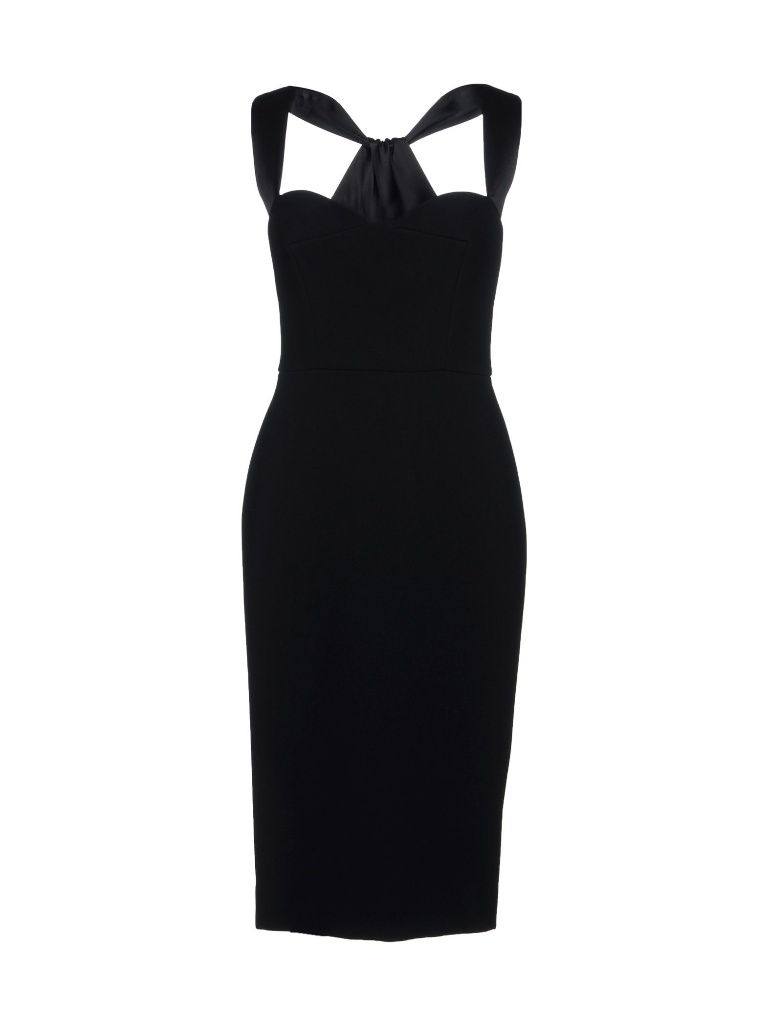} &
\includegraphics[width=0.16\linewidth]{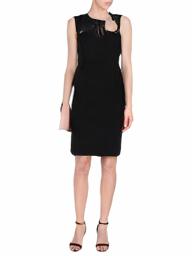} & &
\includegraphics[width=0.16\linewidth]{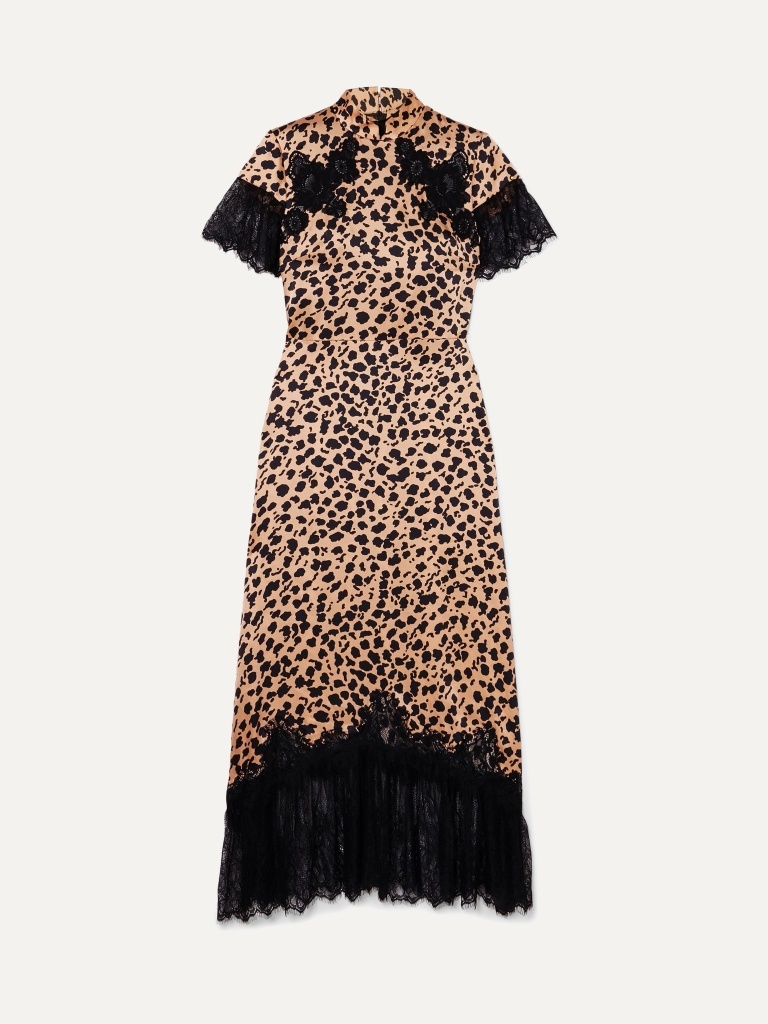} &
\includegraphics[width=0.16\linewidth]{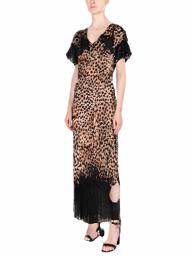} \\
\end{tabular}
}
\caption{Failure cases on the Dress Code test set.}
\label{fig:failure_cases}
\end{figure*}

\begin{figure*}[t]
\centering
\scriptsize
\setlength{\tabcolsep}{.2em}
\resizebox{\linewidth}{!}{
\begin{tabular}{ccc c ccc c ccc} 
& \textbf{Ours} & \textbf{Ours} & & & \textbf{Ours} & \textbf{Ours} & & & \textbf{Ours} & \textbf{Ours}  \\ 
& \textbf{(Patch)} & \textbf{(PSAD)} & & & \textbf{(Patch)} & \textbf{(PSAD)} & & & \textbf{(Patch)} & \textbf{(PSAD)} \\ 
\addlinespace[0.08cm]
\includegraphics[width=0.16\linewidth]{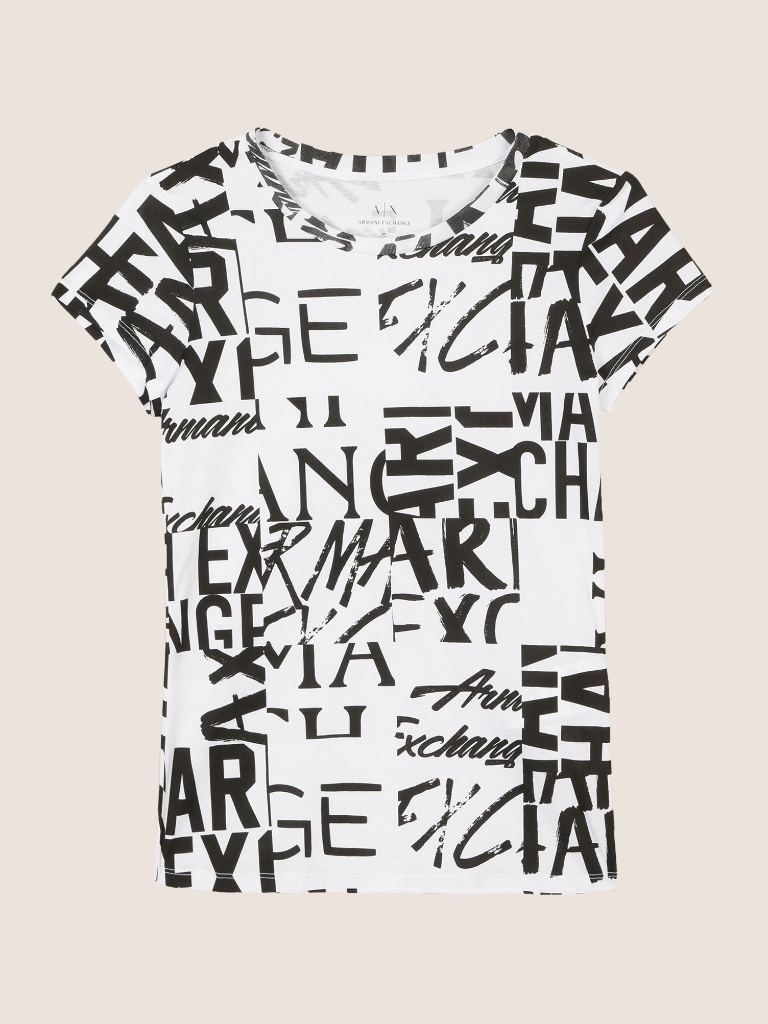} &
\includegraphics[width=0.16\linewidth]{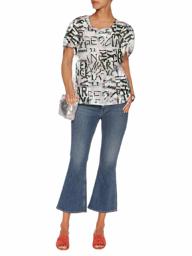} & 
\includegraphics[width=0.16\linewidth]{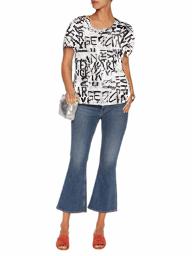} & &
\includegraphics[width=0.16\linewidth]{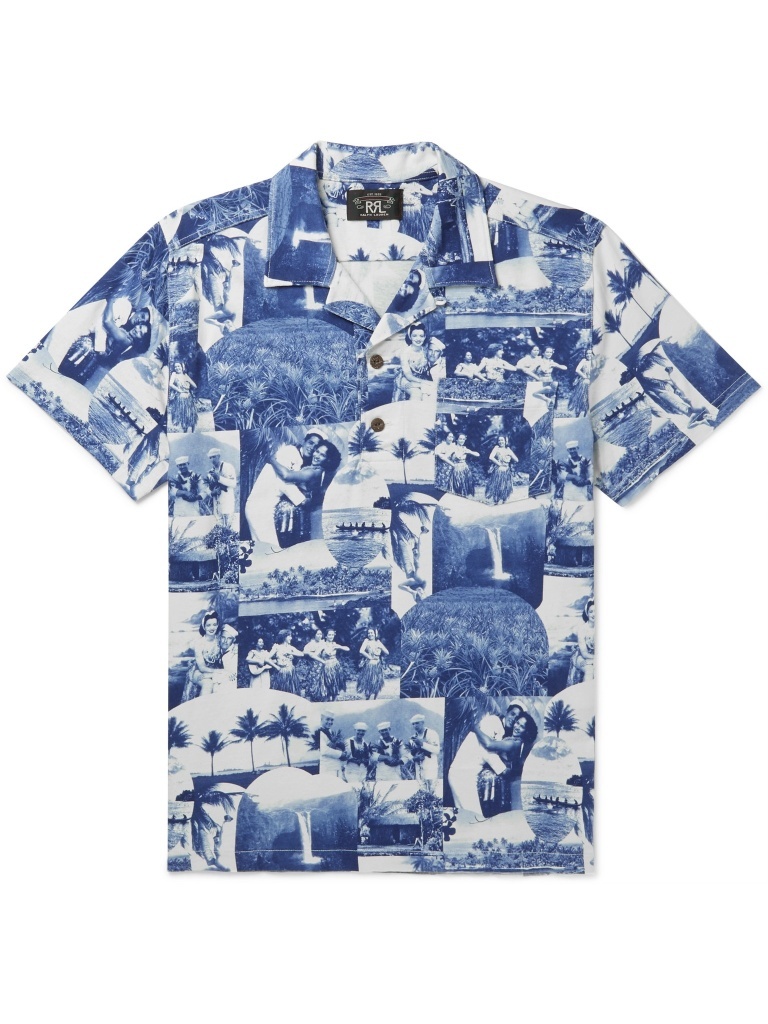} &
\includegraphics[width=0.16\linewidth]{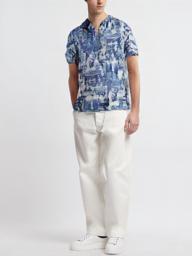} & 
\includegraphics[width=0.16\linewidth]{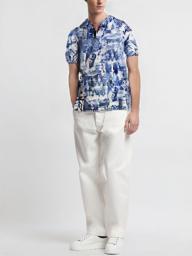} & &
\includegraphics[width=0.16\linewidth]{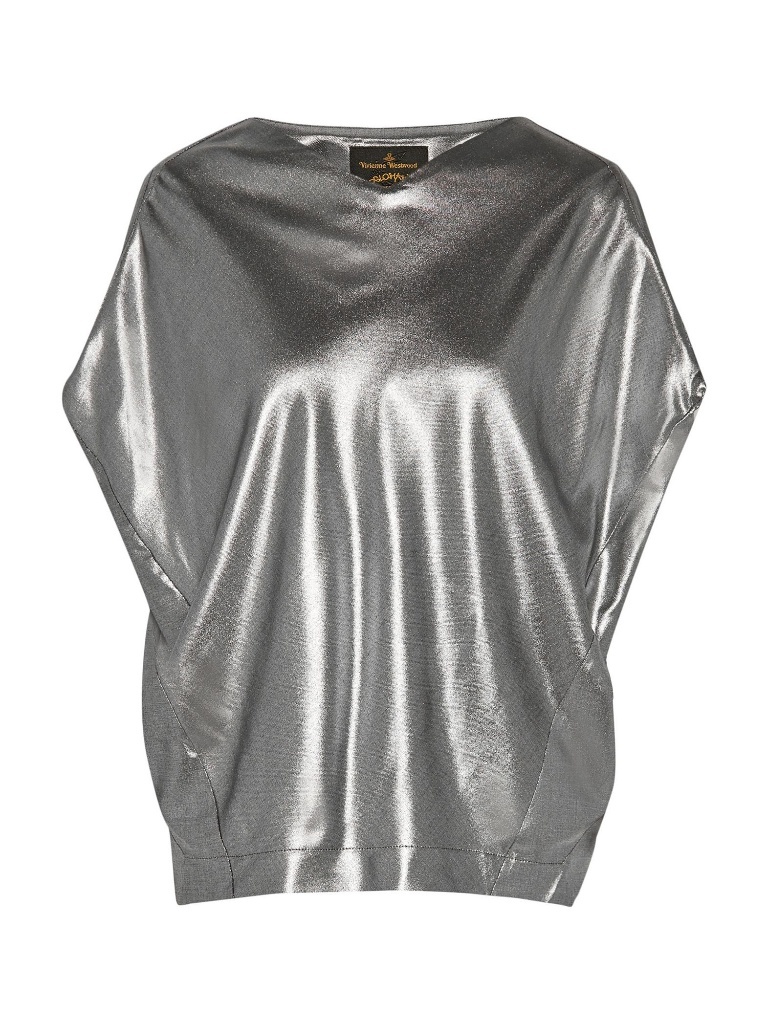} &
\includegraphics[width=0.16\linewidth]{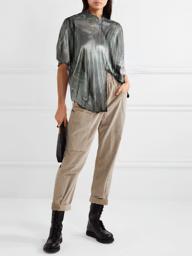} & 
\includegraphics[width=0.16\linewidth]{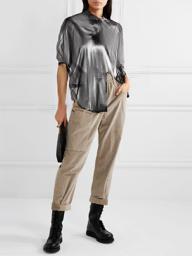} \\
\includegraphics[width=0.16\linewidth]{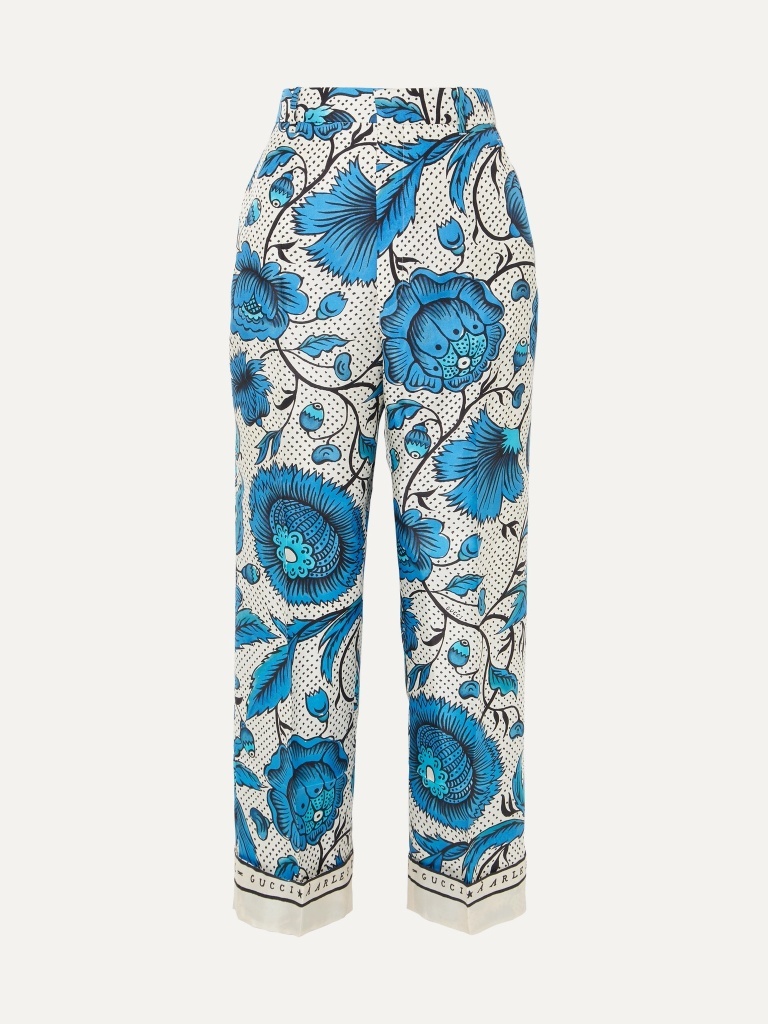} &
\includegraphics[width=0.16\linewidth]{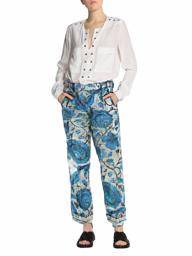} & 
\includegraphics[width=0.16\linewidth]{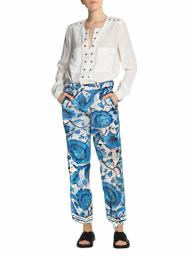} & &
\includegraphics[width=0.16\linewidth]{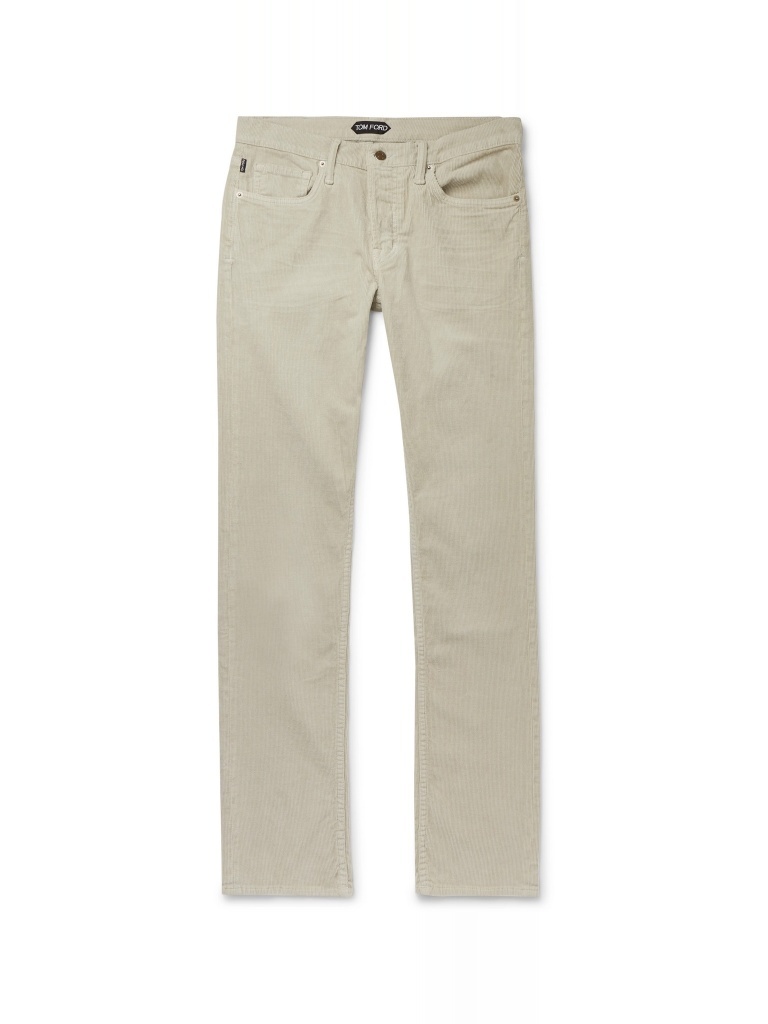} &
\includegraphics[width=0.16\linewidth]{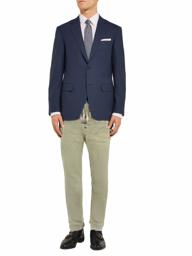} & 
\includegraphics[width=0.16\linewidth]{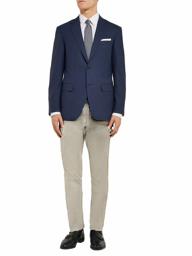} & &
\includegraphics[width=0.16\linewidth]{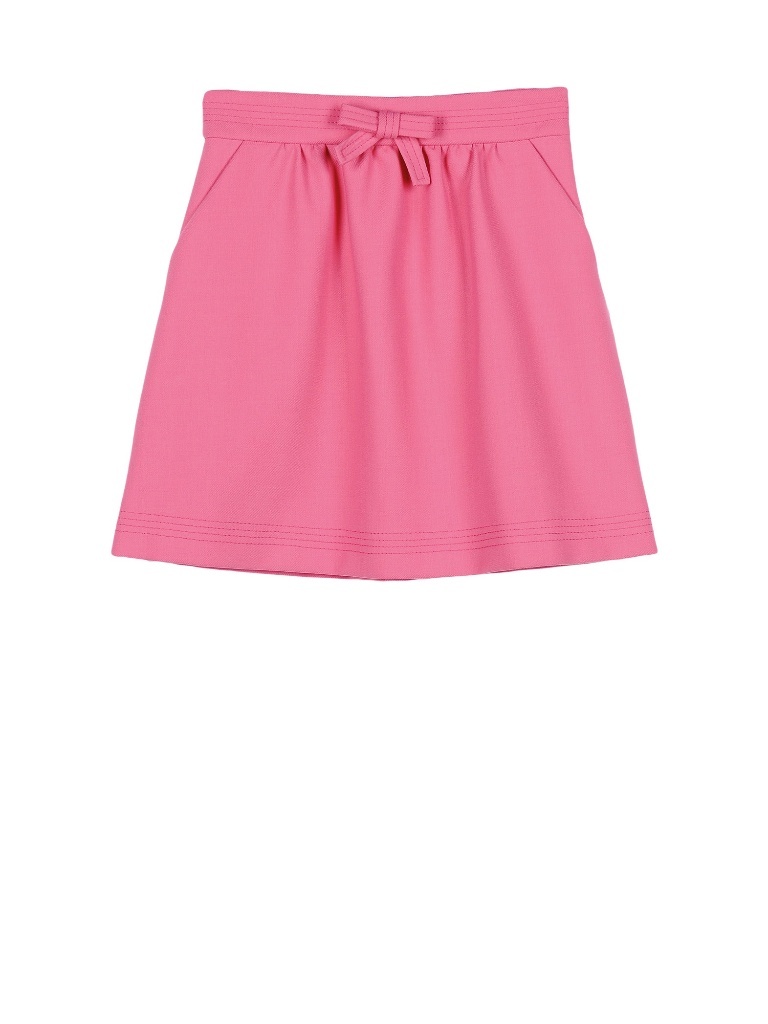} &
\includegraphics[width=0.16\linewidth]{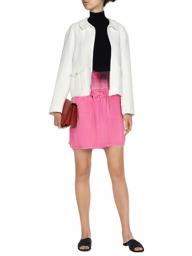} & 
\includegraphics[width=0.16\linewidth]{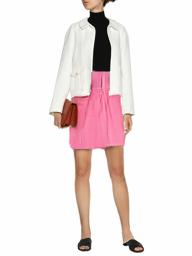} \\
\includegraphics[width=0.16\linewidth]{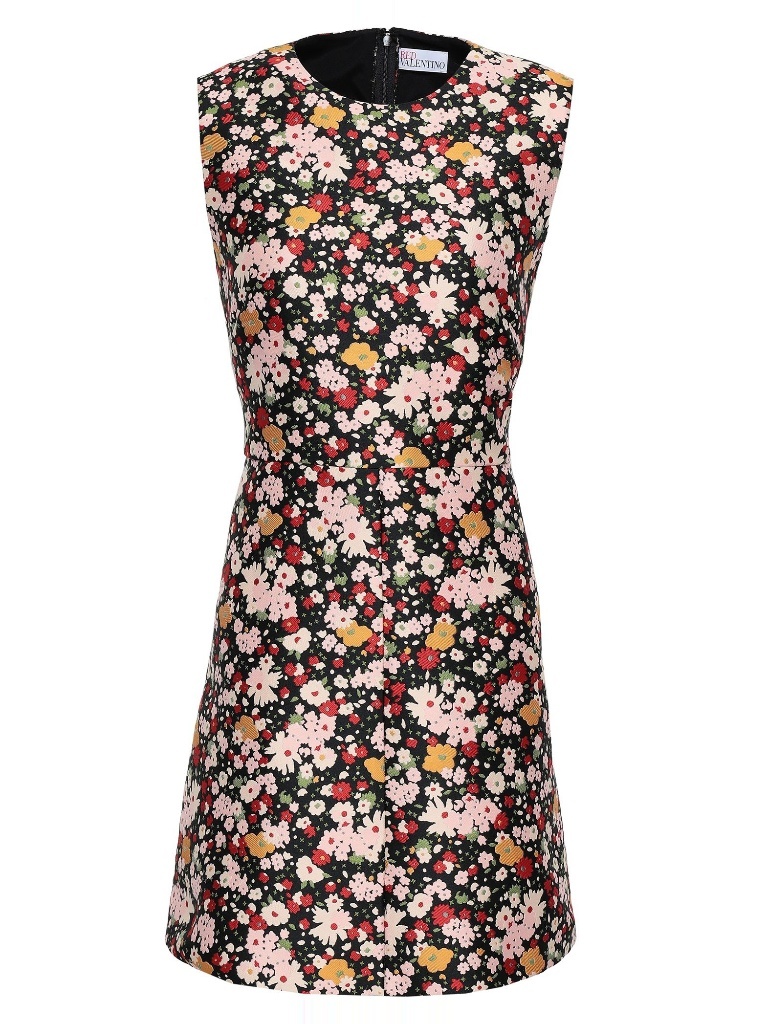} &
\includegraphics[width=0.16\linewidth]{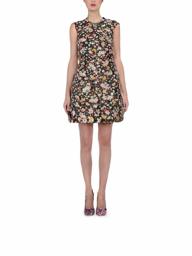} & 
\includegraphics[width=0.16\linewidth]{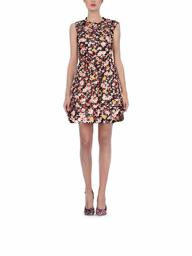} & &
\includegraphics[width=0.16\linewidth]{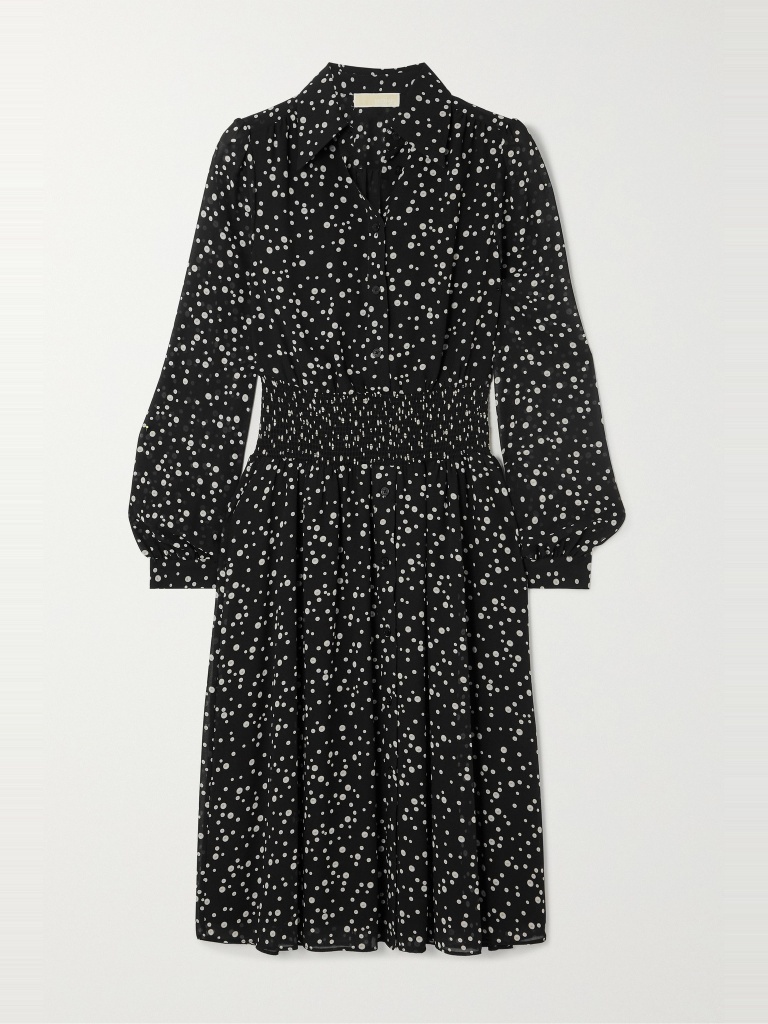} &
\includegraphics[width=0.16\linewidth]{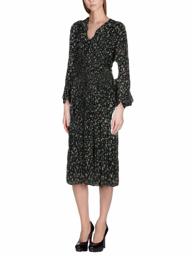} & 
\includegraphics[width=0.16\linewidth]{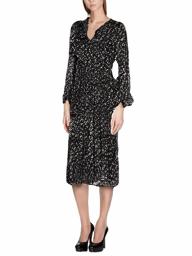} & &
\includegraphics[width=0.16\linewidth]{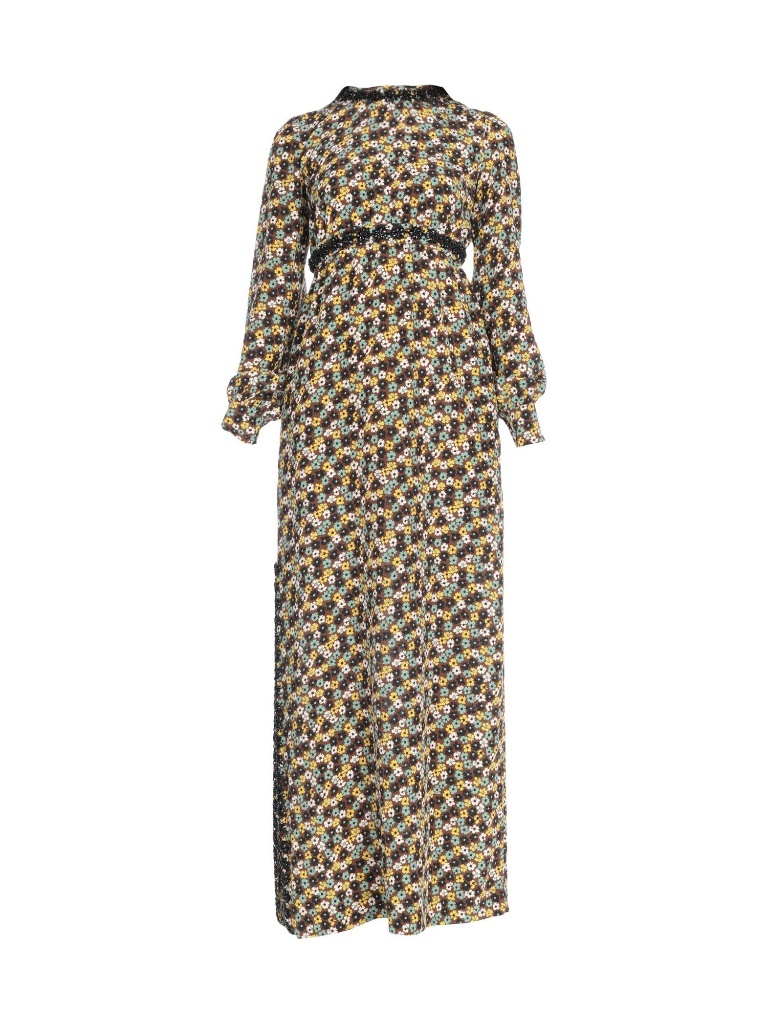} &
\includegraphics[width=0.16\linewidth]{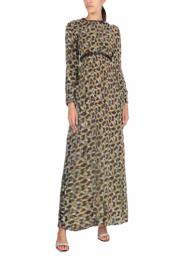} & 
\includegraphics[width=0.16\linewidth]{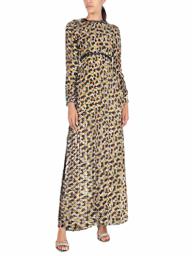} \\
\end{tabular}
}
\caption{Qualitative comparison between PSAD and the Patch-based baseline.}
\label{fig:comparison_disc}
\end{figure*}

\begin{figure*}[t]
\centering
\footnotesize
\setlength{\tabcolsep}{.2em}
\resizebox{\linewidth}{!}{
\begin{tabular}{cccccc}
\includegraphics[width=0.18\linewidth]{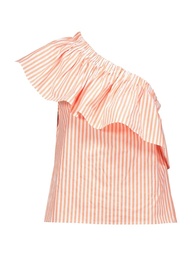} &
\includegraphics[width=0.18\linewidth]{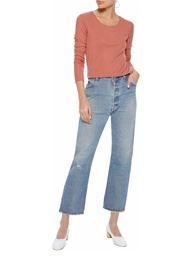} & 
\includegraphics[width=0.18\linewidth]{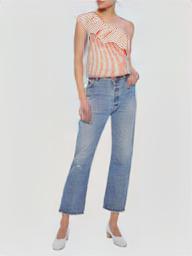} &
\includegraphics[width=0.18\linewidth]{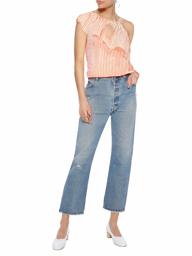} &
\includegraphics[width=0.18\linewidth]{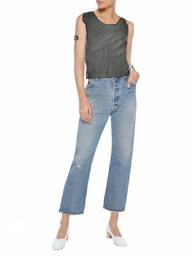} & 
\includegraphics[width=0.18\linewidth]{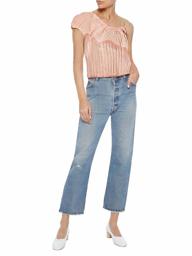}\\
\includegraphics[width=0.18\linewidth]{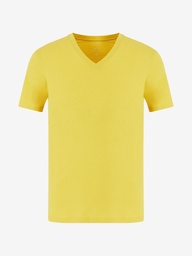} &
\includegraphics[width=0.18\linewidth]{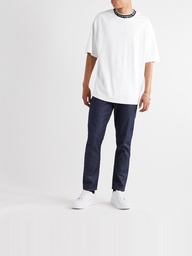} & 
\includegraphics[width=0.18\linewidth]{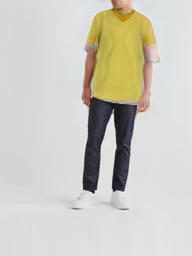} &
\includegraphics[width=0.18\linewidth]{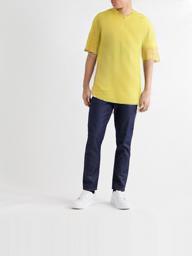} &
\includegraphics[width=0.18\linewidth]{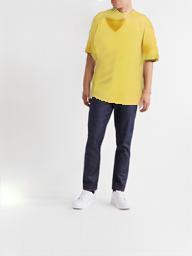} & 
\includegraphics[width=0.18\linewidth]{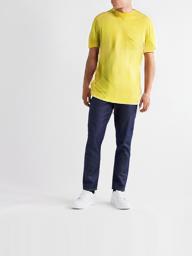}\\
\includegraphics[width=0.18\linewidth]{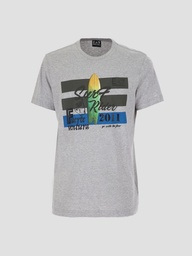} &
\includegraphics[width=0.18\linewidth]{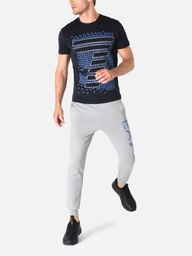} & 
\includegraphics[width=0.18\linewidth]{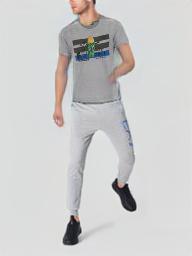} &
\includegraphics[width=0.18\linewidth]{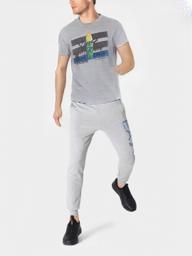} &
\includegraphics[width=0.18\linewidth]{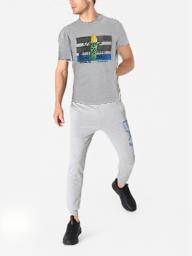} & 
\includegraphics[width=0.18\linewidth]{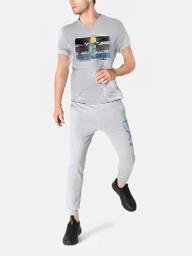}\\
\includegraphics[width=0.18\linewidth]{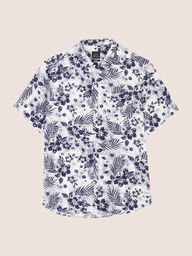} &
\includegraphics[width=0.18\linewidth]{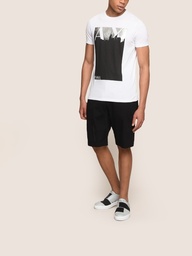} & 
\includegraphics[width=0.18\linewidth]{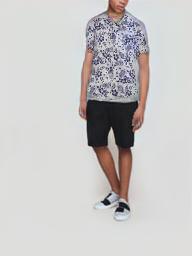} &
\includegraphics[width=0.18\linewidth]{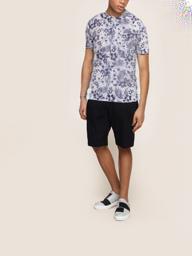} &
\includegraphics[width=0.18\linewidth]{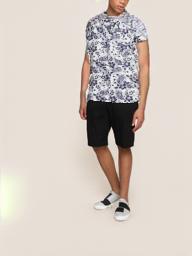} & 
\includegraphics[width=0.18\linewidth]{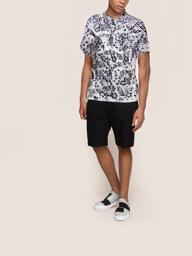}\\
\includegraphics[width=0.18\linewidth]{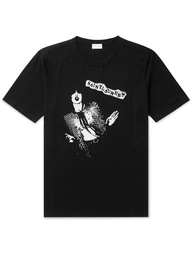} &
\includegraphics[width=0.18\linewidth]{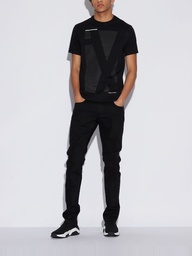} & 
\includegraphics[width=0.18\linewidth]{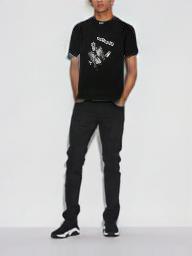} &
\includegraphics[width=0.18\linewidth]{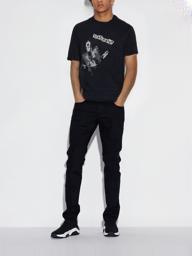} &
\includegraphics[width=0.18\linewidth]{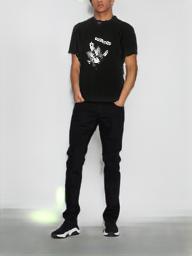} & 
\includegraphics[width=0.18\linewidth]{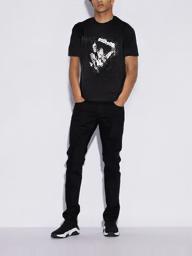}\\
\includegraphics[width=0.18\linewidth]{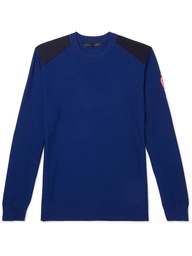} &
\includegraphics[width=0.18\linewidth]{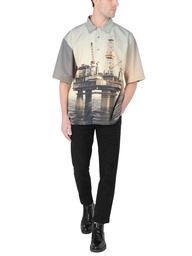} & 
\includegraphics[width=0.18\linewidth]{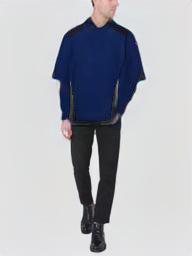} &
\includegraphics[width=0.18\linewidth]{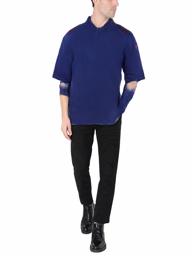} &
\includegraphics[width=0.18\linewidth]{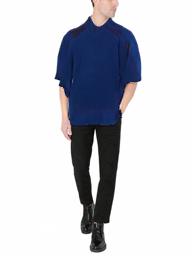} & 
\includegraphics[width=0.18\linewidth]{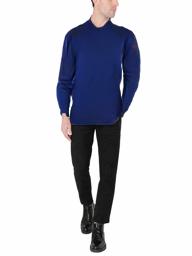}\\
\end{tabular}
}
\caption{Sample try-on results using upper-body clothes and reference models from the Dress Code test set.}
\label{fig:upper_body}
\end{figure*}

\begin{figure*}[t]
\centering
\footnotesize
\setlength{\tabcolsep}{.2em}
\resizebox{\linewidth}{!}{
\begin{tabular}{cccccc}
\includegraphics[width=0.18\linewidth]{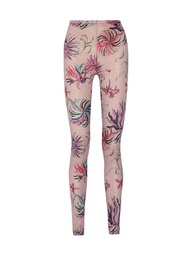} &
\includegraphics[width=0.18\linewidth]{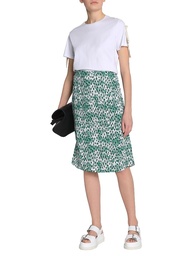} & 
\includegraphics[width=0.18\linewidth]{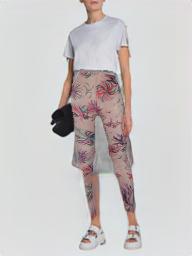} &
\includegraphics[width=0.18\linewidth]{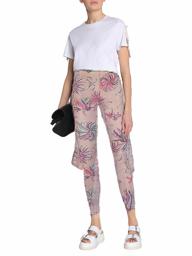} &
\includegraphics[width=0.18\linewidth]{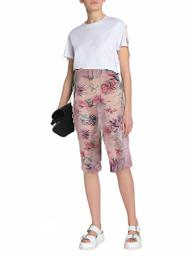} & 
\includegraphics[width=0.18\linewidth]{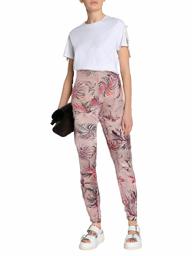}\\
\includegraphics[width=0.18\linewidth]{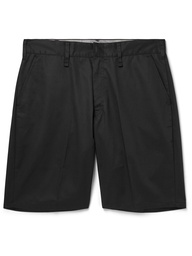} &
\includegraphics[width=0.18\linewidth]{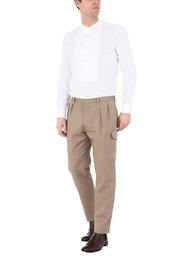} & 
\includegraphics[width=0.18\linewidth]{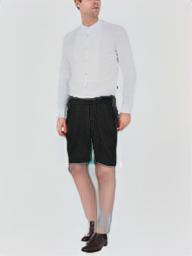} &
\includegraphics[width=0.18\linewidth]{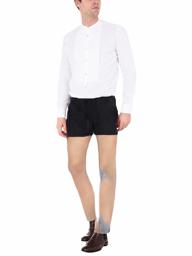} &
\includegraphics[width=0.18\linewidth]{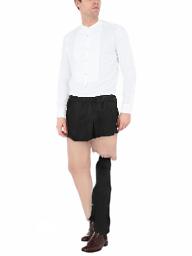} & 
\includegraphics[width=0.18\linewidth]{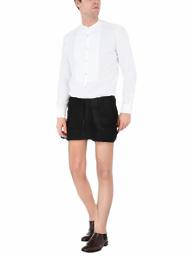}\\
\includegraphics[width=0.18\linewidth]{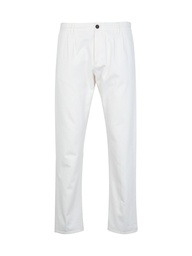} &
\includegraphics[width=0.18\linewidth]{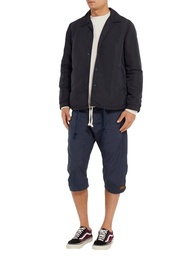} & 
\includegraphics[width=0.18\linewidth]{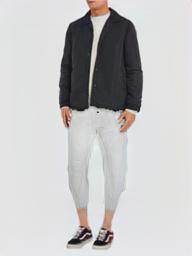} &
\includegraphics[width=0.18\linewidth]{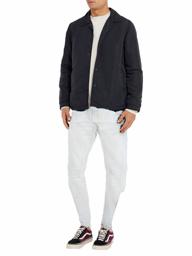} &
\includegraphics[width=0.18\linewidth]{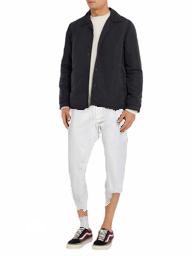} & 
\includegraphics[width=0.18\linewidth]{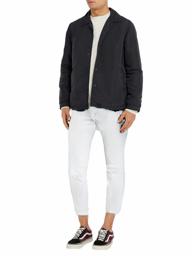}\\
\includegraphics[width=0.18\linewidth]{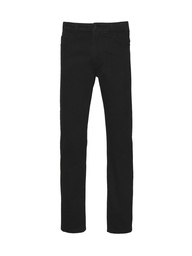} &
\includegraphics[width=0.18\linewidth]{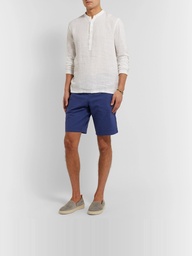} & 
\includegraphics[width=0.18\linewidth]{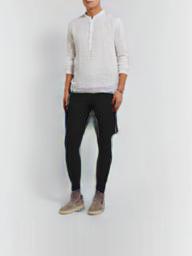} &
\includegraphics[width=0.18\linewidth]{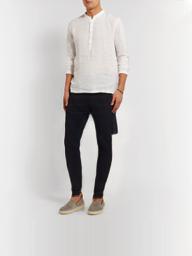} &
\includegraphics[width=0.18\linewidth]{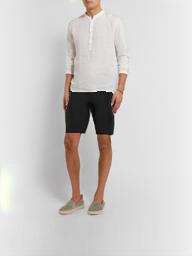} & 
\includegraphics[width=0.18\linewidth]{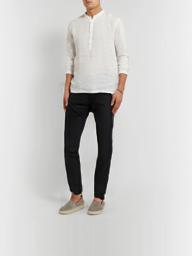}\\
\includegraphics[width=0.18\linewidth]{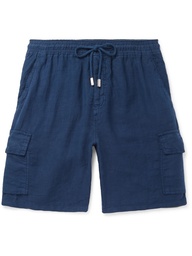} &
\includegraphics[width=0.18\linewidth]{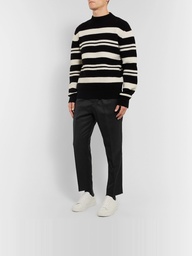} & 
\includegraphics[width=0.18\linewidth]{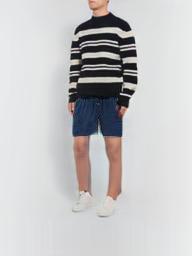} &
\includegraphics[width=0.18\linewidth]{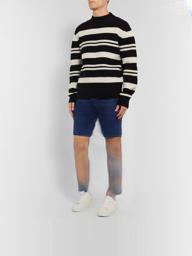} &
\includegraphics[width=0.18\linewidth]{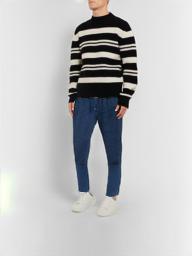} & 
\includegraphics[width=0.18\linewidth]{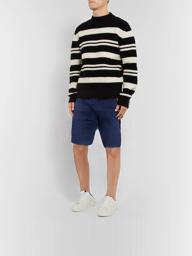}\\
\includegraphics[width=0.18\linewidth]{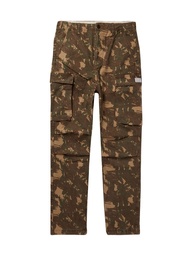} &
\includegraphics[width=0.18\linewidth]{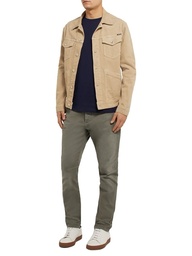} & 
\includegraphics[width=0.18\linewidth]{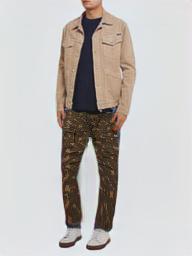} &
\includegraphics[width=0.18\linewidth]{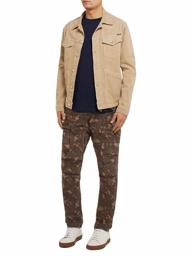} &
\includegraphics[width=0.18\linewidth]{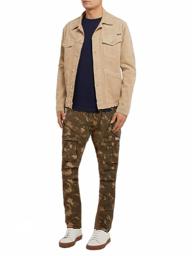} & 
\includegraphics[width=0.18\linewidth]{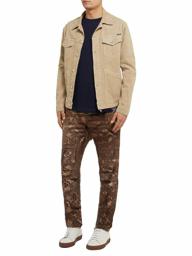}\\
\end{tabular}
}
\caption{Sample try-on results using lower-body clothes and reference models from the Dress Code test set.}
\label{fig:lower_body}
\end{figure*}

\begin{figure*}[t]
\centering
\footnotesize
\setlength{\tabcolsep}{.2em}
\resizebox{\linewidth}{!}{
\begin{tabular}{cccccc}
\includegraphics[width=0.18\linewidth]{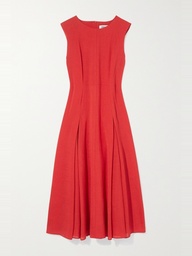} &
\includegraphics[width=0.18\linewidth]{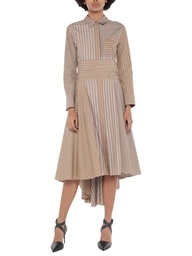} & 
\includegraphics[width=0.18\linewidth]{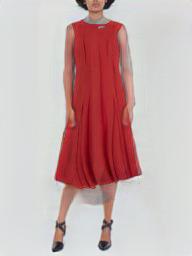} &
\includegraphics[width=0.18\linewidth]{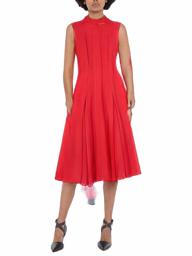} &
\includegraphics[width=0.18\linewidth]{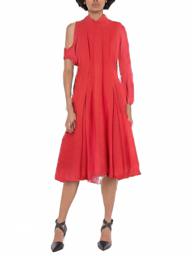} & 
\includegraphics[width=0.18\linewidth]{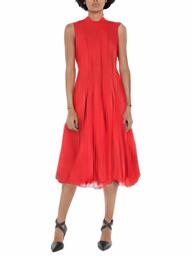}\\
\includegraphics[width=0.18\linewidth]{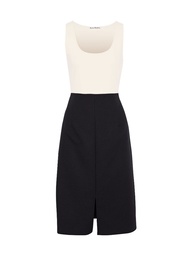} &
\includegraphics[width=0.18\linewidth]{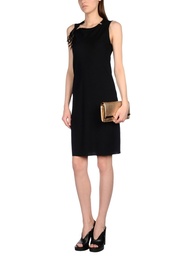} & 
\includegraphics[width=0.18\linewidth]{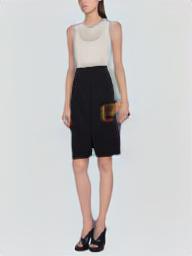} &
\includegraphics[width=0.18\linewidth]{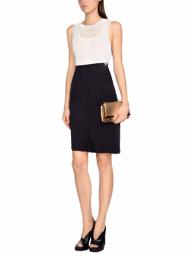} &
\includegraphics[width=0.18\linewidth]{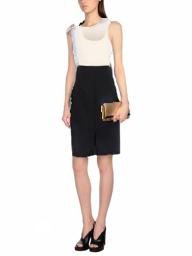} & 
\includegraphics[width=0.18\linewidth]{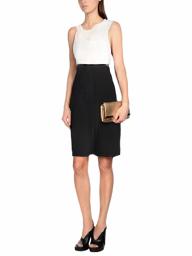}\\
\includegraphics[width=0.18\linewidth]{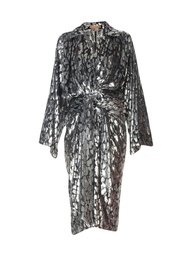} &
\includegraphics[width=0.18\linewidth]{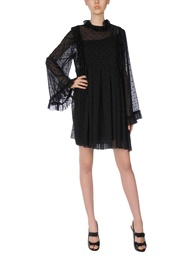} & 
\includegraphics[width=0.18\linewidth]{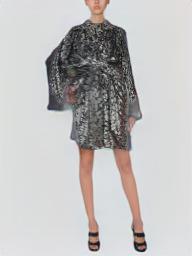} &
\includegraphics[width=0.18\linewidth]{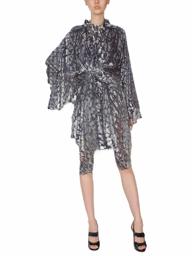} &
\includegraphics[width=0.18\linewidth]{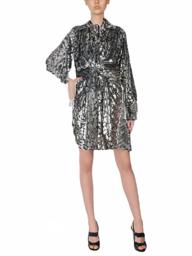} & 
\includegraphics[width=0.18\linewidth]{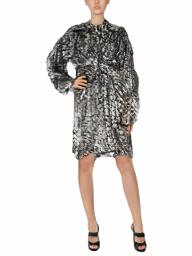}\\
\includegraphics[width=0.18\linewidth]{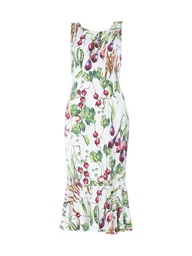} &
\includegraphics[width=0.18\linewidth]{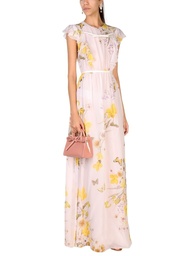} & 
\includegraphics[width=0.18\linewidth]{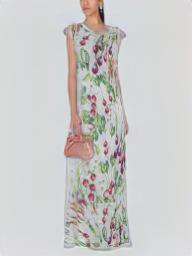} &
\includegraphics[width=0.18\linewidth]{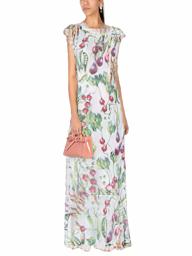} &
\includegraphics[width=0.18\linewidth]{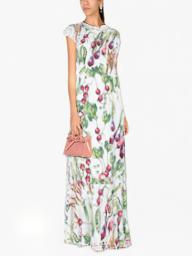} & 
\includegraphics[width=0.18\linewidth]{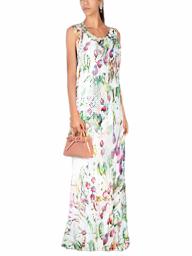}\\
\includegraphics[width=0.18\linewidth]{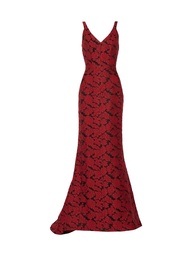} &
\includegraphics[width=0.18\linewidth]{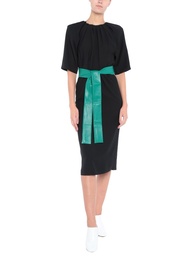} & 
\includegraphics[width=0.18\linewidth]{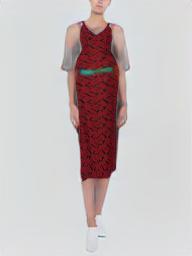} &
\includegraphics[width=0.18\linewidth]{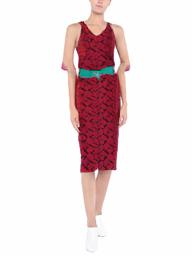} &
\includegraphics[width=0.18\linewidth]{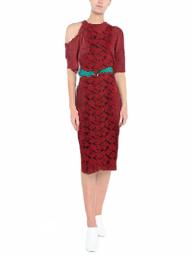} & 
\includegraphics[width=0.18\linewidth]{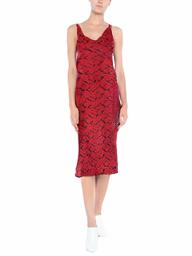}\\
\includegraphics[width=0.18\linewidth]{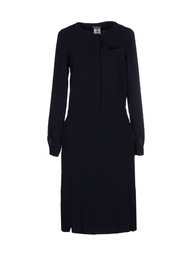} &
\includegraphics[width=0.18\linewidth]{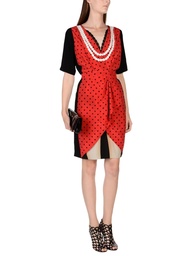} & 
\includegraphics[width=0.18\linewidth]{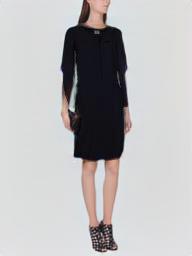} &
\includegraphics[width=0.18\linewidth]{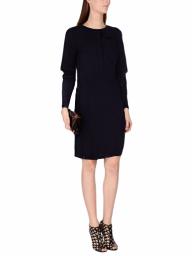} &
\includegraphics[width=0.18\linewidth]{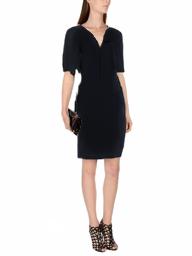} & 
\includegraphics[width=0.18\linewidth]{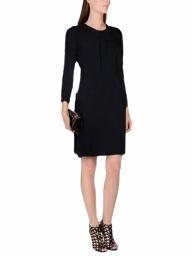}\\
\end{tabular}
}
\caption{Sample try-on results using dresses and reference models from the Dress Code test set.}
\label{fig:dresses}
\end{figure*}

\tinytit{Low-Resolution Results} In Fig.~\ref{fig:failure_cases}, we show some failure cases, while some additional qualitative comparisons between PSAD and the corresponding Patch-based baseline are shown in Fig.~\ref{fig:comparison_disc}. 
In Fig.~\ref{fig:upper_body},~\ref{fig:lower_body}, and~\ref{fig:dresses}, we report further try-on results on sample image pairs respectively extracted from upper-body clothes, lower-body clothes, and dresses by comparing our model with previously proposed try-on architectures re-trained on our newly collected dataset.

\begin{figure*}[t]
\centering
\resizebox{\linewidth}{!}{
\begin{tabular}{c}
\includegraphics[width=0.92\linewidth]{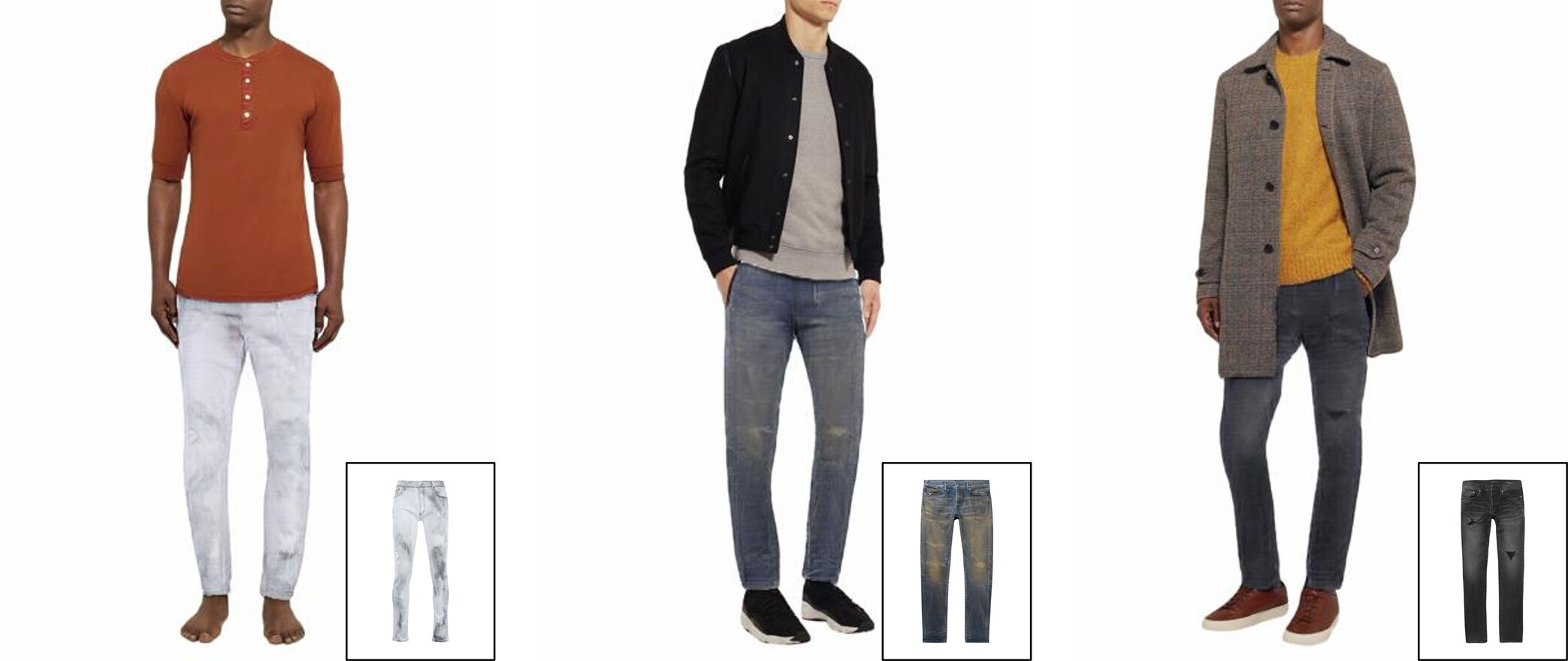} \\
\includegraphics[width=0.92\linewidth]{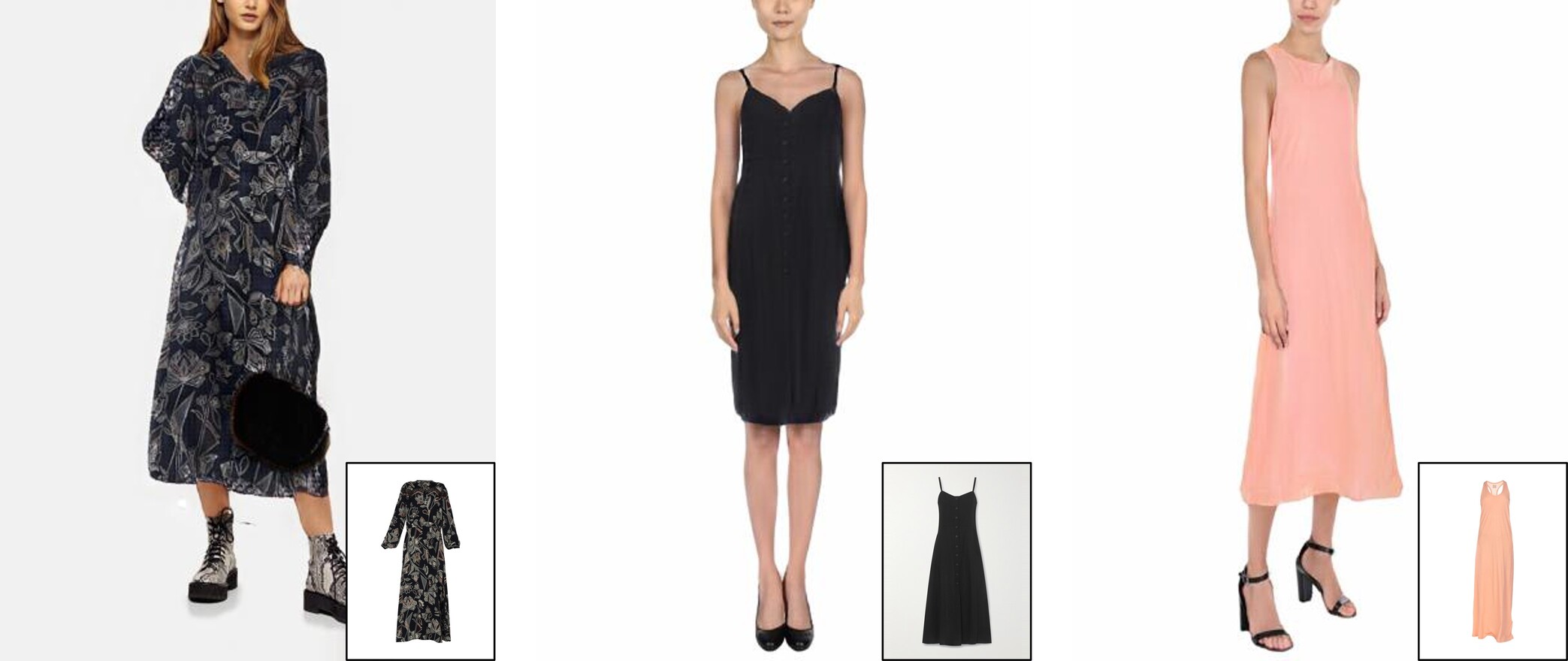} \\
\includegraphics[width=0.92\linewidth]{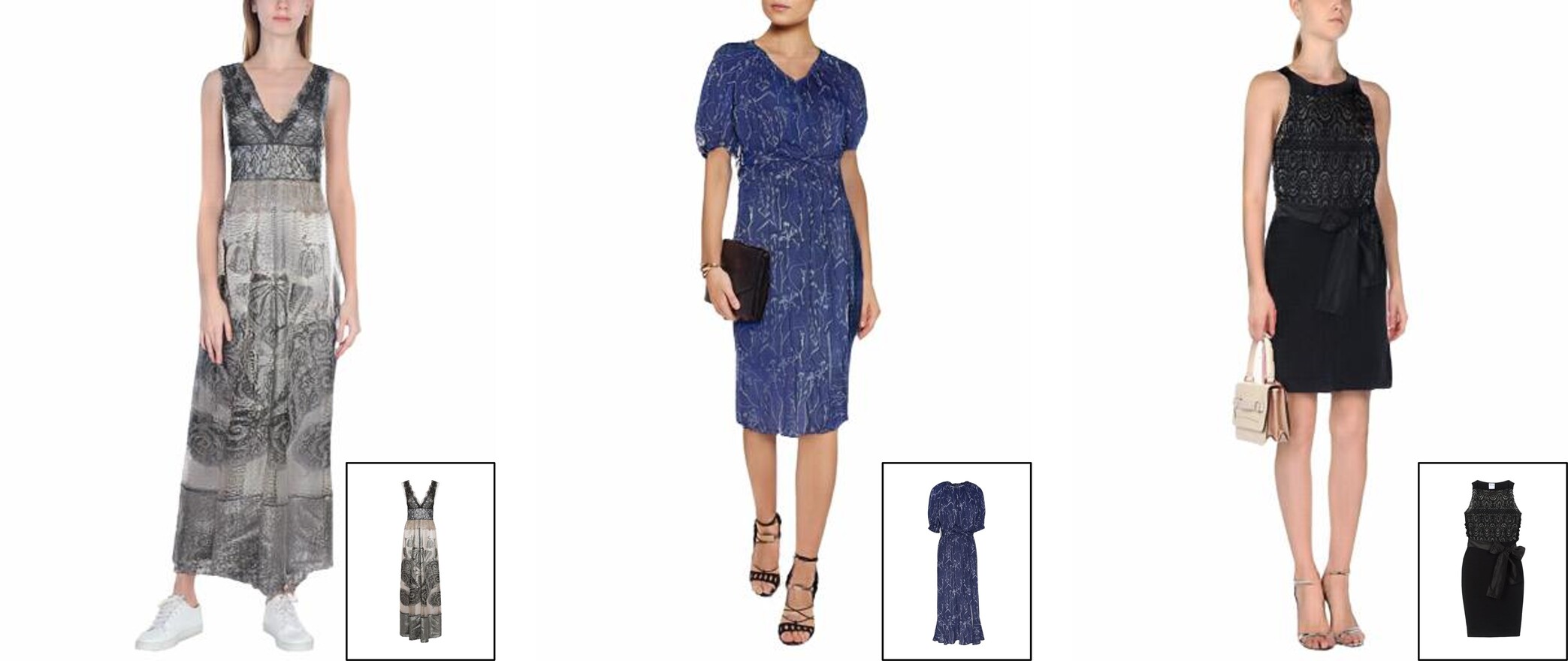} \\
\end{tabular}
}
\caption{Sample high-resolution results on the Dress Code test set.}
\label{fig:hd_res}
\end{figure*}

\tit{High-Resolution Results}
Table~\ref{tab:hd_try-on_res} shows the complete try-on performances when generating high-resolution images while, in Fig.~\ref{fig:hd_res}, we report some qualitative results.

\begin{figure*}[t]
\centering
\scriptsize
\setlength{\tabcolsep}{.2em}
\resizebox{\linewidth}{!}{
\begin{tabular}{cccccc}
& & \textbf{CP-VTON$^\dagger$} & \textbf{WUTON} & \textbf{ACGPN} & \textbf{Ours} \\
& & \cite{wang2018toward} & \cite{issenhuth2019end} & \cite{yang2020towards} & \textbf{(PSAD)} \\
\addlinespace[0.08cm]
\addlinespace[0.08cm]
\includegraphics[width=0.18\linewidth]{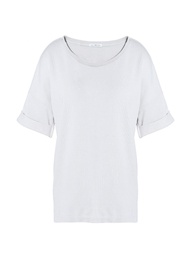} &
\includegraphics[width=0.18\linewidth]{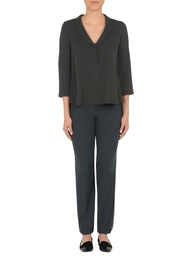} & 
\includegraphics[width=0.18\linewidth]{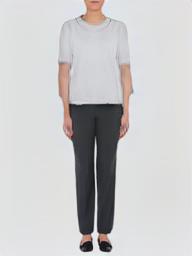} &
\includegraphics[width=0.18\linewidth]{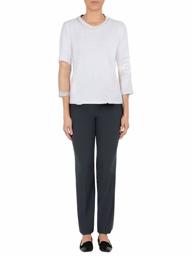} &
\includegraphics[width=0.18\linewidth]{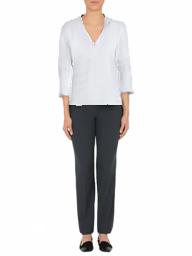} & 
\includegraphics[width=0.18\linewidth]{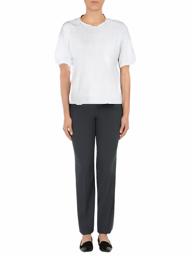}\\
\includegraphics[width=0.18\linewidth]{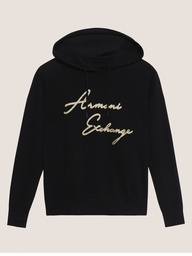} &
\includegraphics[width=0.18\linewidth]{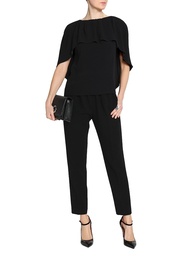} & 
\includegraphics[width=0.18\linewidth]{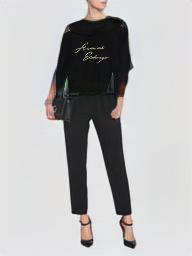} &
\includegraphics[width=0.18\linewidth]{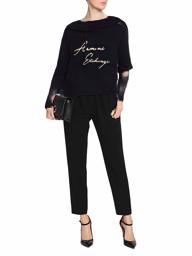} &
\includegraphics[width=0.18\linewidth]{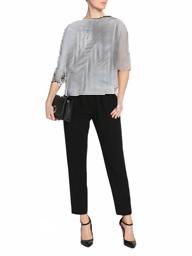} & 
\includegraphics[width=0.18\linewidth]{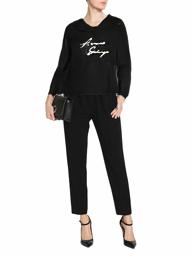}\\
\includegraphics[width=0.18\linewidth]{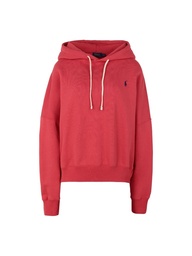} &
\includegraphics[width=0.18\linewidth]{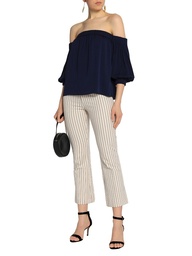} & 
\includegraphics[width=0.18\linewidth]{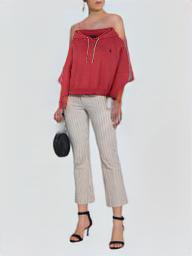} &
\includegraphics[width=0.18\linewidth]{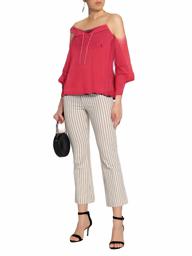} &
\includegraphics[width=0.18\linewidth]{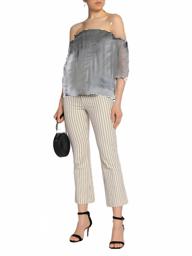} & 
\includegraphics[width=0.18\linewidth]{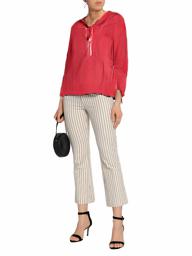}\\
\includegraphics[width=0.18\linewidth]{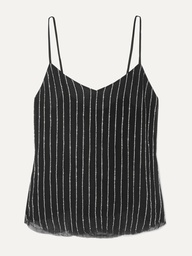} &
\includegraphics[width=0.18\linewidth]{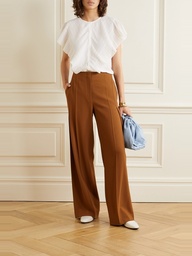} & 
\includegraphics[width=0.18\linewidth]{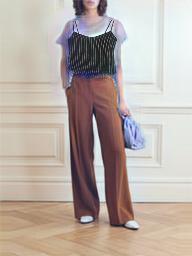} &
\includegraphics[width=0.18\linewidth]{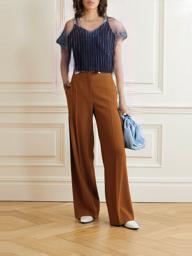} &
\includegraphics[width=0.18\linewidth]{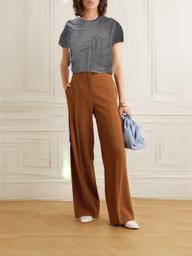} & 
\includegraphics[width=0.18\linewidth]{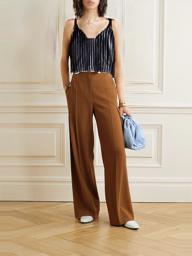}\\
\includegraphics[width=0.18\linewidth]{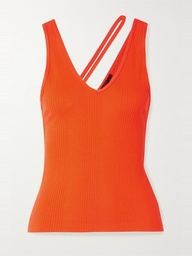} &
\includegraphics[width=0.18\linewidth]{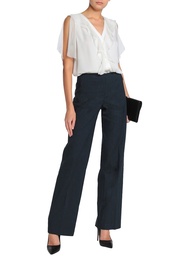} & 
\includegraphics[width=0.18\linewidth]{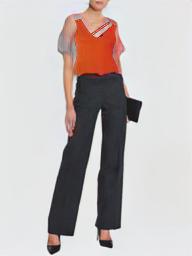} &
\includegraphics[width=0.18\linewidth]{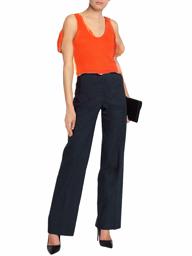} &
\includegraphics[width=0.18\linewidth]{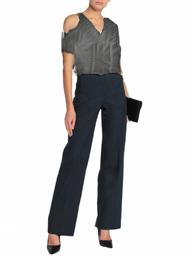} & 
\includegraphics[width=0.18\linewidth]{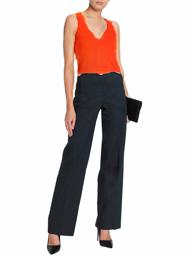}\\
\includegraphics[width=0.18\linewidth]{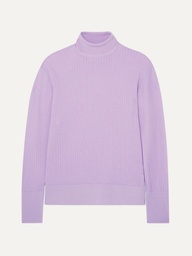} &
\includegraphics[width=0.18\linewidth]{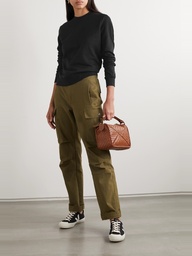} & 
\includegraphics[width=0.18\linewidth]{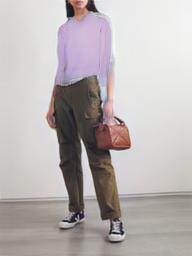} &
\includegraphics[width=0.18\linewidth]{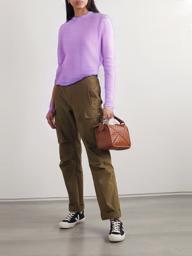} &
\includegraphics[width=0.18\linewidth]{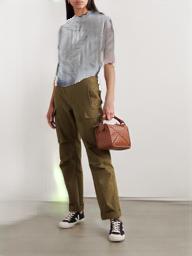} & 
\includegraphics[width=0.18\linewidth]{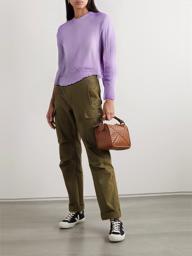}\\
\end{tabular}
}
\end{figure*}

\begin{figure*}[t]
\centering
\scriptsize
\setlength{\tabcolsep}{.2em}
\resizebox{\linewidth}{!}{
\begin{tabular}{cccccc}
& & \textbf{CP-VTON$^\dagger$} & \textbf{WUTON} & \textbf{ACGPN} & \textbf{Ours} \\
& & \cite{wang2018toward} & \cite{issenhuth2019end} & \cite{yang2020towards} & \textbf{(PSAD)} \\
\addlinespace[0.08cm]
\includegraphics[width=0.18\linewidth]{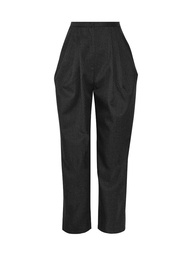} &
\includegraphics[width=0.18\linewidth]{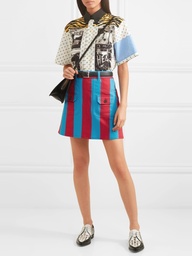} & 
\includegraphics[width=0.18\linewidth]{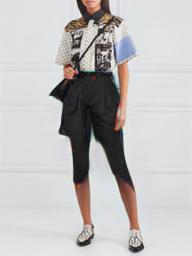} &
\includegraphics[width=0.18\linewidth]{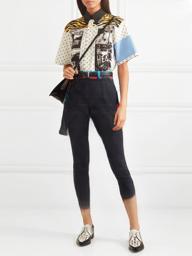} &
\includegraphics[width=0.18\linewidth]{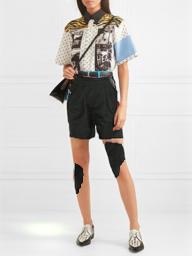} & 
\includegraphics[width=0.18\linewidth]{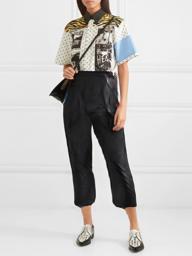}\\
\includegraphics[width=0.18\linewidth]{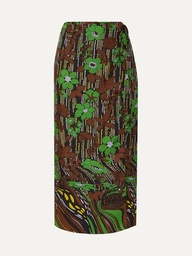} &
\includegraphics[width=0.18\linewidth]{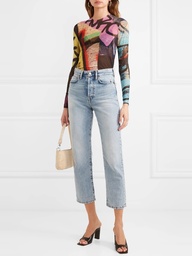} & 
\includegraphics[width=0.18\linewidth]{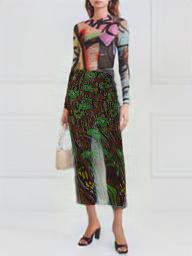} &
\includegraphics[width=0.18\linewidth]{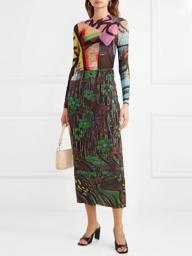} &
\includegraphics[width=0.18\linewidth]{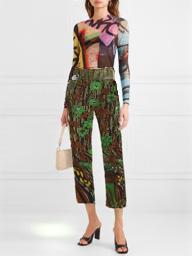} & 
\includegraphics[width=0.18\linewidth]{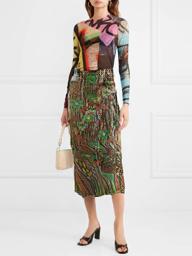}\\
\includegraphics[width=0.18\linewidth]{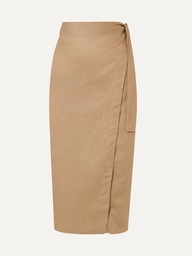} &
\includegraphics[width=0.18\linewidth]{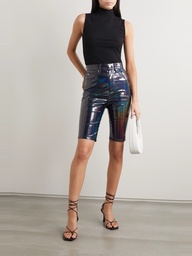} & 
\includegraphics[width=0.18\linewidth]{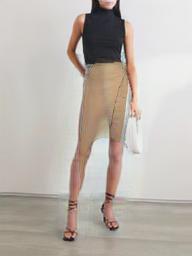} &
\includegraphics[width=0.18\linewidth]{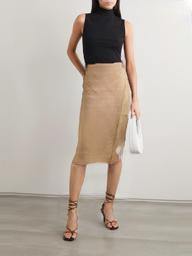} &
\includegraphics[width=0.18\linewidth]{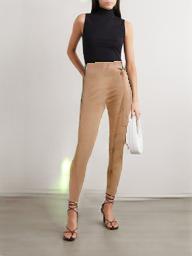} & 
\includegraphics[width=0.18\linewidth]{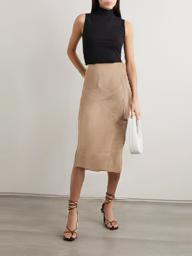}\\
\includegraphics[width=0.18\linewidth]{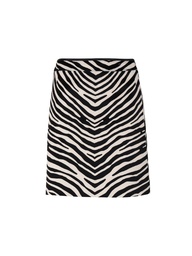} &
\includegraphics[width=0.18\linewidth]{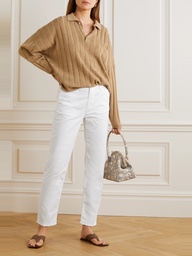} & 
\includegraphics[width=0.18\linewidth]{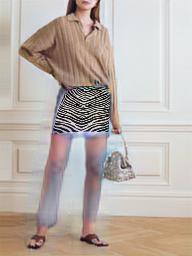} &
\includegraphics[width=0.18\linewidth]{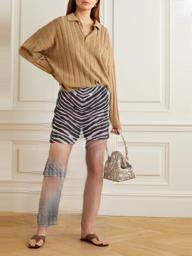} &
\includegraphics[width=0.18\linewidth]{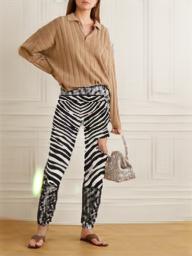} & 
\includegraphics[width=0.18\linewidth]{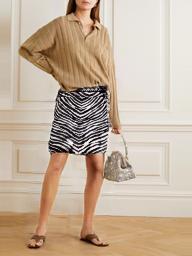}\\
\includegraphics[width=0.18\linewidth]{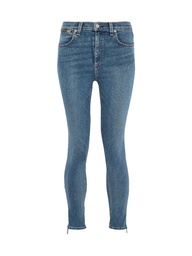} &
\includegraphics[width=0.18\linewidth]{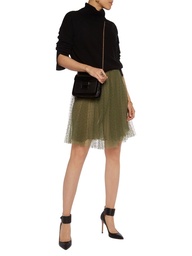} & 
\includegraphics[width=0.18\linewidth]{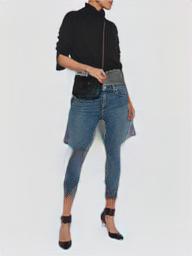} &
\includegraphics[width=0.18\linewidth]{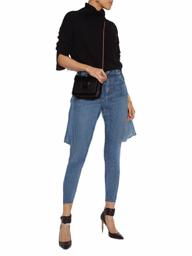} &
\includegraphics[width=0.18\linewidth]{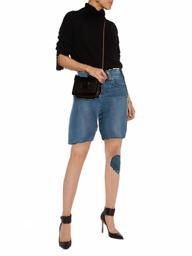} & 
\includegraphics[width=0.18\linewidth]{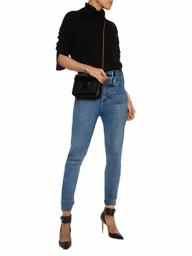}\\
\includegraphics[width=0.18\linewidth]{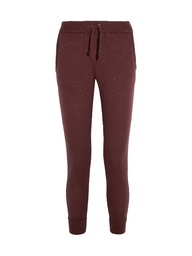} &
\includegraphics[width=0.18\linewidth]{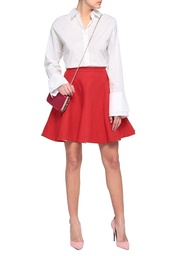} & 
\includegraphics[width=0.18\linewidth]{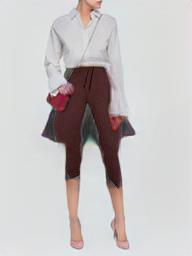} &
\includegraphics[width=0.18\linewidth]{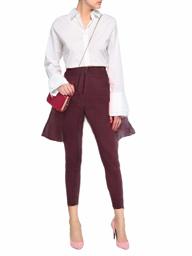} &
\includegraphics[width=0.18\linewidth]{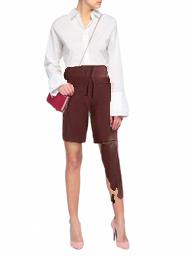} & 
\includegraphics[width=0.18\linewidth]{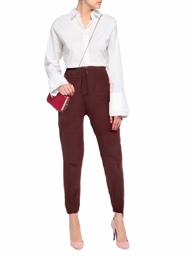}\\
\end{tabular}
}
\end{figure*}

\begin{figure*}[t]
\centering
\scriptsize
\setlength{\tabcolsep}{.2em}
\resizebox{\linewidth}{!}{
\begin{tabular}{cccccc}
& & \textbf{CP-VTON$^\dagger$} & \textbf{WUTON} & \textbf{ACGPN} & \textbf{Ours} \\
& & \cite{wang2018toward} & \cite{issenhuth2019end} & \cite{yang2020towards} & \textbf{(PSAD)} \\
\addlinespace[0.08cm]
\includegraphics[width=0.18\linewidth]{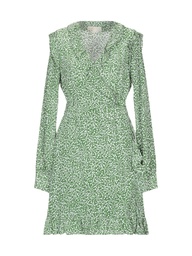} &
\includegraphics[width=0.18\linewidth]{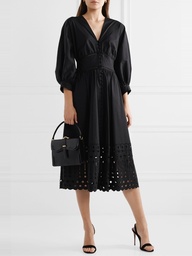} & 
\includegraphics[width=0.18\linewidth]{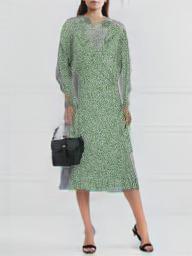} &
\includegraphics[width=0.18\linewidth]{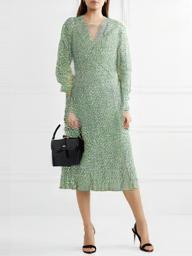} &
\includegraphics[width=0.18\linewidth]{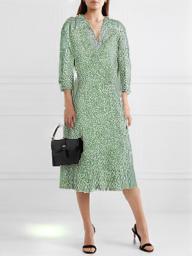} & 
\includegraphics[width=0.18\linewidth]{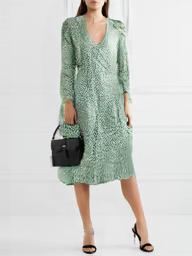}\\
\includegraphics[width=0.18\linewidth]{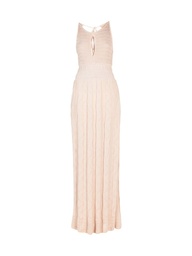} &
\includegraphics[width=0.18\linewidth]{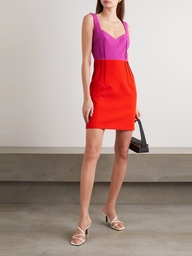} & 
\includegraphics[width=0.18\linewidth]{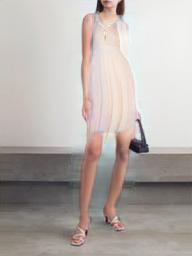} &
\includegraphics[width=0.18\linewidth]{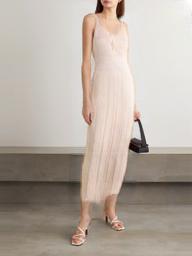} &
\includegraphics[width=0.18\linewidth]{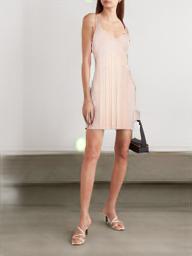} & 
\includegraphics[width=0.18\linewidth]{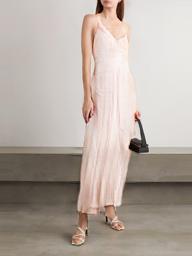}\\
\includegraphics[width=0.18\linewidth]{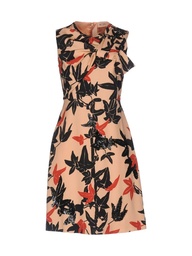} &
\includegraphics[width=0.18\linewidth]{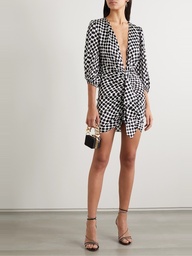} & 
\includegraphics[width=0.18\linewidth]{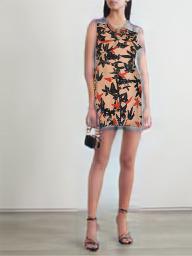} &
\includegraphics[width=0.18\linewidth]{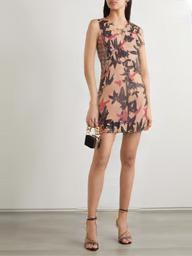} &
\includegraphics[width=0.18\linewidth]{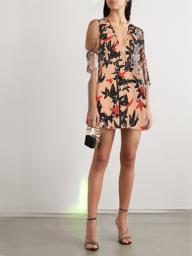} & 
\includegraphics[width=0.18\linewidth]{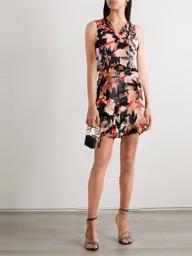}\\
\includegraphics[width=0.18\linewidth]{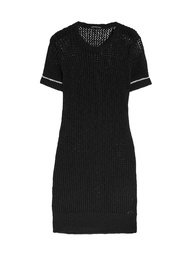} &
\includegraphics[width=0.18\linewidth]{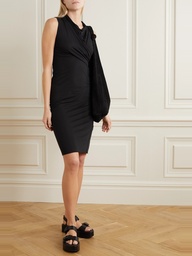} & 
\includegraphics[width=0.18\linewidth]{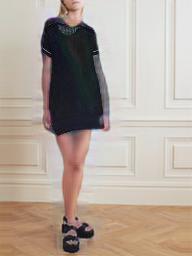} &
\includegraphics[width=0.18\linewidth]{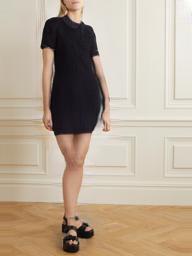} &
\includegraphics[width=0.18\linewidth]{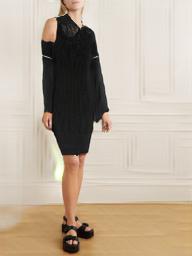} & 
\includegraphics[width=0.18\linewidth]{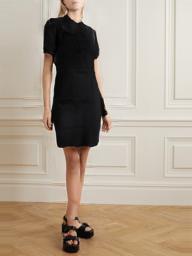}\\
\includegraphics[width=0.18\linewidth]{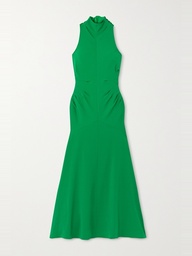} &
\includegraphics[width=0.18\linewidth]{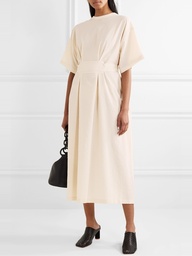} & 
\includegraphics[width=0.18\linewidth]{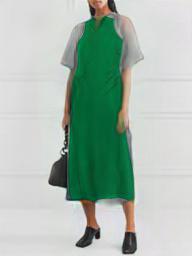} &
\includegraphics[width=0.18\linewidth]{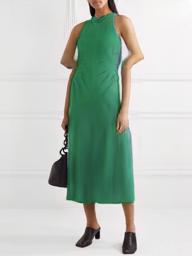} &
\includegraphics[width=0.18\linewidth]{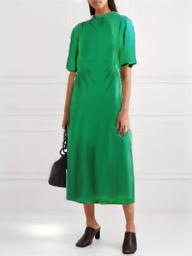} & 
\includegraphics[width=0.18\linewidth]{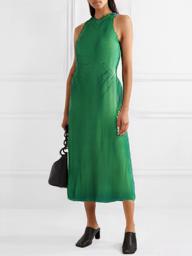}\\
\includegraphics[width=0.18\linewidth]{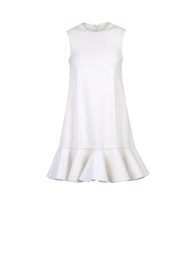} &
\includegraphics[width=0.18\linewidth]{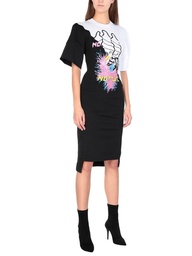} & 
\includegraphics[width=0.18\linewidth]{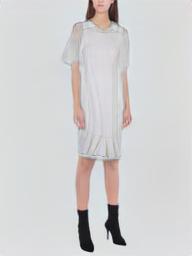} &
\includegraphics[width=0.18\linewidth]{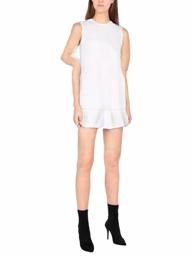} &
\includegraphics[width=0.18\linewidth]{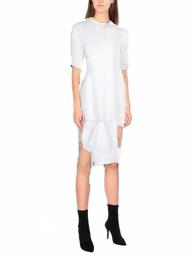} & 
\includegraphics[width=0.18\linewidth]{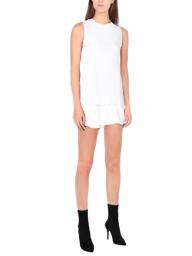}\\
\end{tabular}
}
\end{figure*}

\begin{figure*}[t]
\centering
\resizebox{\linewidth}{!}{
\begin{tabular}{c}
\includegraphics[width=0.92\linewidth]{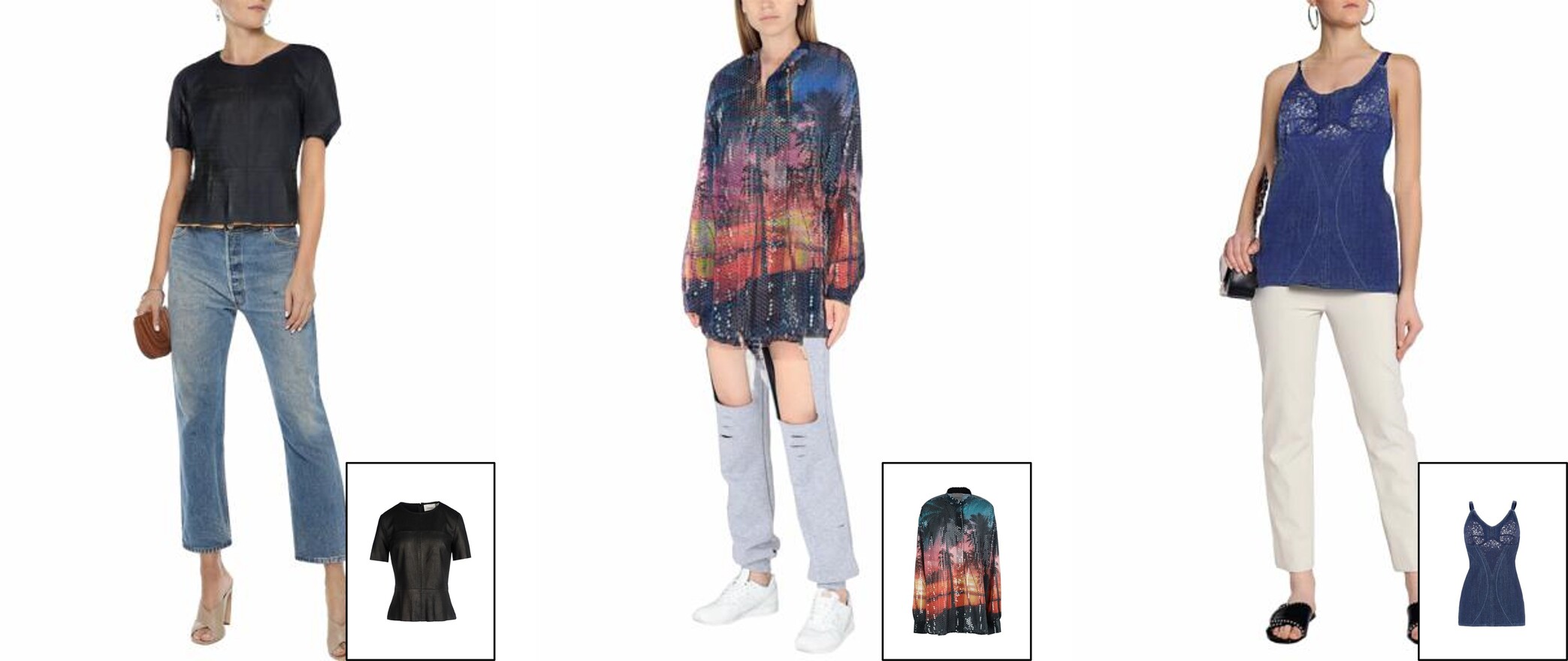} \\
\includegraphics[width=0.92\linewidth]{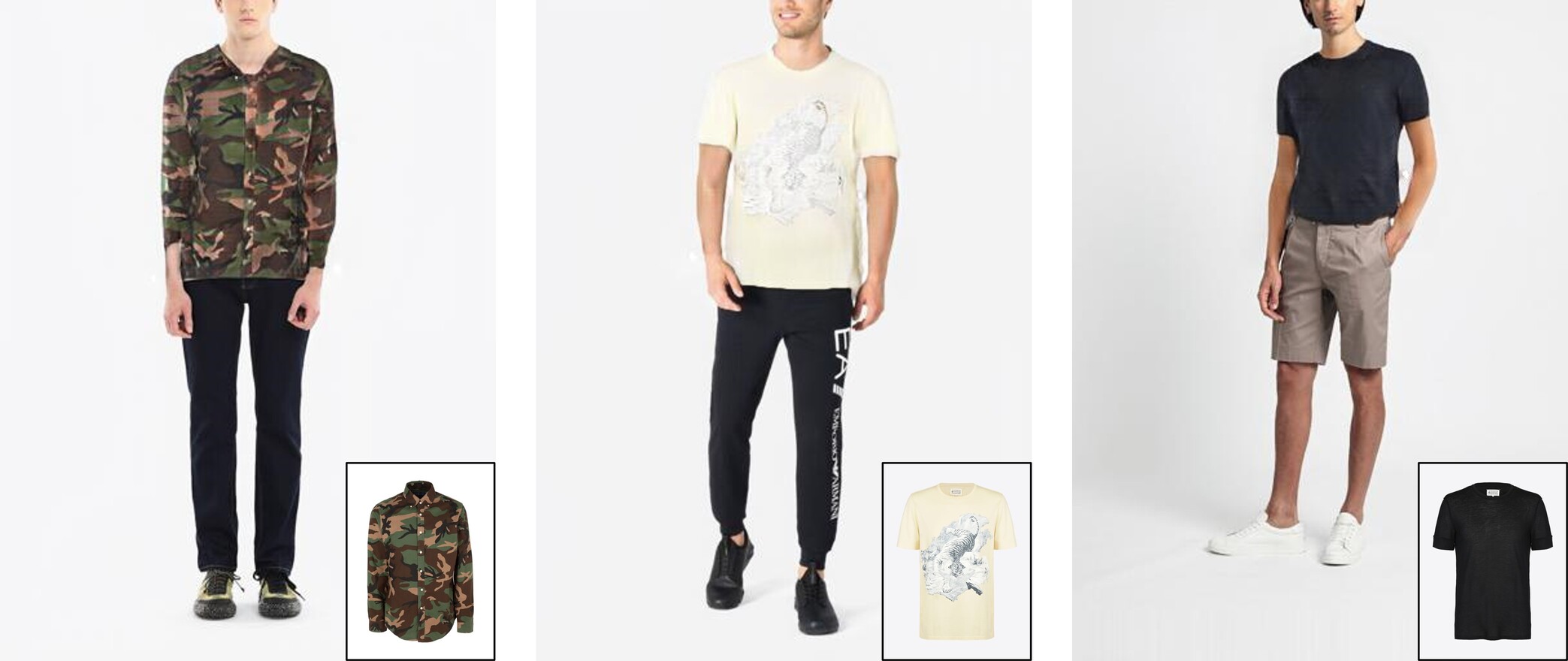} \\
\includegraphics[width=0.92\linewidth]{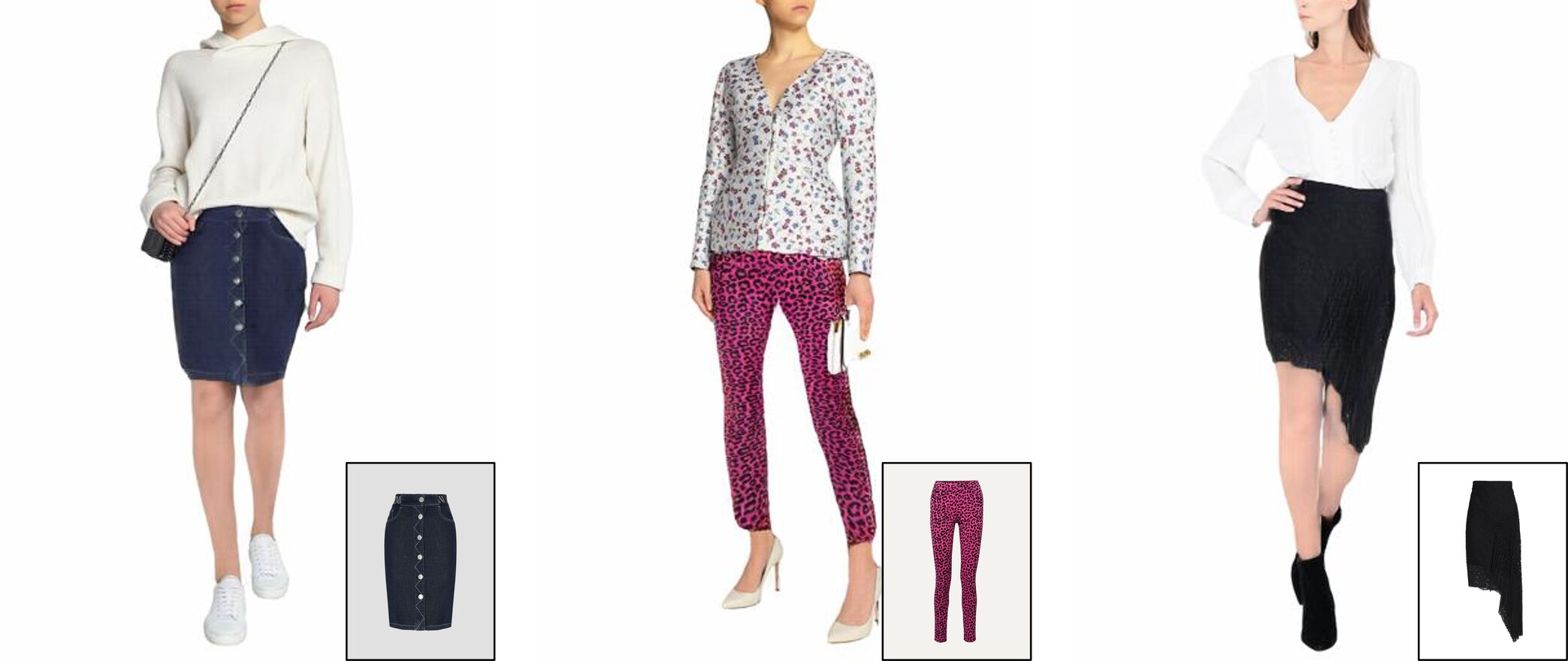} \\
\end{tabular}
}
\end{figure*}

\end{document}